\documentclass{article}

\usepackage{microtype}
\usepackage{graphicx}
\usepackage{subfigure}
\usepackage{booktabs} 

\usepackage{hyperref}

\usepackage[accepted]{icml2021}

\usepackage{url}

\def\psfancypar#1#2{\begingroup\def\par{\endgraf\endgroup\lineskiplimit=0pt}
               \setbox2=\hbox{\large\sc #2}
               \newdimen\tmpht \tmpht \ht2 \advance\tmpht by \baselineskip
               \font\hhuge=Times-Bold at \tmpht
               \setbox1=\hbox{{\hhuge #1}}
               \count7=\tmpht \count8=\ht1
               \divide\count8 by 1000 \divide\count7 by \count8 
               \tmpht=.001\tmpht\multiply\tmpht by \count7 
               \font\hhuge=Times-Bold at \tmpht
               \setbox1=\hbox{{\hhuge #1}}
               \noindent
                \hangindent1.05\wd1
               \hangafter=-2 {\hskip-\hangindent
               \lower1\ht1\hbox{\raise1.0\ht2\copy1}%
                \kern-0\wd1}\copy2\lineskiplimit=-1000pt}

\newcommand{\E}{\mbox{{\rm E}}}

\def\boxit#1{\vbox{\hrule\hbox{\vrule\kern3pt
        \vbox{\kern3pt#1\kern3pt}\kern3pt\vrule}\hrule}}

\def\reals{ { {\rm  I \kern-0.15em R }  } }
\def\complex{ {\,{{\rm C} \kern-0.50em \raise0.20ex {  |}}\, }}

\def\Rbf{{\bf R}}

\def\Ac{{\cal A}}

\def\Dc{{\cal D}}

\def\Hc{{\cal H}}

\def\Sc{{\cal S}}

\def\be{\vskip .3cm \begin{equation}}
\def\ee{\end{equation} \vskip .4cm \noindent}
\newcommand{\R}{\mbox{$\hat {\bf R}_{N}$}}

\def\Rxx{\Rbf_{\ssstyle X\kern-.1em X}}

\let\ssstyle=\scriptscriptstyle

\def\Kout{\setbox1=\hbox{\Huge\bf K}\hbox to
1.05\wd1{\hspace{.05\wd1}
}}
\def\Sout{\setbox1=\hbox{\Huge\bf S}\hbox to 1.05\wd1{\hspace{.05\wd1}
}}

  \ifx\LabelFigloaded\MYundefined\relax
  \else
   \fi

  \def\LabelFigloaded{\relax}

  \chardef\LabelFigCatAt\the\catcode`\@
  \catcode`\@=11

 \let\LabelFigwlog@ld\wlog
 \def\wlog#1{\relax}

 \ifx\\\MYundefined@
    \let\\\relax
 \fi

  \def\ms@g{\immediate\write16}

 \def\N@wif{\csname newif\endcsname }
 \def\Temp@ {\N@wif\ifIN@}
 \ifx\INN@\MYundefined@
    \else \let\Temp@\relax
 \fi
 \Temp@

  \def\IN@{\expandafter\INN@\expandafter}
  \long\def\INN@0#1@#2@{\long\def\NI@##1#1##2##3\ENDNI@
    {\ifx\m@rker##2\IN@false\else\IN@true\fi}%
     \expandafter\NI@#2@@#1\m@rker\ENDNI@}
  \def\m@rker{\m@@rker}
 
  \newtoks\Initialtoks@  \newtoks\Terminaltoks@
  \def\SPLIT@{\expandafter\SPLITT@\expandafter}
  \def\SPLITT@0#1@#2@{\def\TTILPS@##1#1##2@{%
     \Initialtoks@{##1}\Terminaltoks@{##2}}\expandafter\TTILPS@#2@}

 \def\Shifted@@#1#2#3{\setbox0=\hbox{#3}%
   \raise -\dp0\vbox {\kern-#2%
       \hbox {\kern#1\unhbox0\kern-#1}%
           \kern#2}}

 \newcount\gridcount
 \newbox\auxGridbox@ \newbox\hGridbox@ \newbox\vGridbox@
 \newbox\Labelbox@ \newbox\auxLabelbox@
 \newbox\Coordinatebox@
 \newtoks\Labeltoks@
 \newdimen\Wdd@ \newdimen\Htt@
 \newdimen\Wddd@ \newdimen\Httt@
 
 \def\Wr@{\immediate\write16}

 \newdimen\GL@wd
 \def\GridLineWidth#1{\GL@wd=#1}

 \def\gobble#1{}
 \def\EdgeErr@{\Wr@{}%
      \Wr@{\string\Edges\space argument
      1, 10, 100 or 1000 please\string!}%
      }

 \newcount\Edgect@

 \def\Sweepup#1\endSweepup{}

 \def\SetEdges@{%
    \edef\Zr@@s{\expandafter\gobble\number\Edgect@\empty}%
        \count255=0\Zr@@s\relax
        \ifnum\count255=\z@\else\EdgeErr@\show\tailtest\fi
        \count255=1\Zr@@s\relax
        \ifnum\count255=\Edgect@\relax\else\EdgeErr@\show\leadtest\fi
    \EdgGl@b\edef\Zr@s{\expandafter\gobble\Zr@@s\empty}
    \ifnum\Edgect@>\@ne\relax\EdgGl@b\let\L@Dc\empty
        \else\EdgGl@b\edef\L@Dc{\string.}\fi
    \ifnum\Edgect@>\@ne\relax
        \EdgGl@b\edef\Edgescale@##1{\divide##1 by \Edgect@}%
        \else\EdgGl@b\edef\Edgescale@##1{}\fi
    }

 \def\Edges#1{\Edgect@=#1\relax
     \let\EdgGl@b\global \SetEdges@}

 \Edges{1}

 \def\hhrule{\hrule height \GL@wd\vskip-.\GL@wd}

 \def\hRule@{%
   \advance\gridcount -2%
   \vfil\hhrule\vfil
   \llap{\smash{\raise -2.5pt
     \hbox{\L@Dc\number\gridcount\Zr@s\kern2pt}}}%
   \hhrule
   }

\def\vvrule{\vrule width \GL@wd \kern-\GL@wd}

 \def\vRule@{\advance\gridcount 2%
   \hfil\vvrule\hfil
   \setbox\auxGridbox@=\vbox to 0pt
      {\vskip \Htt@\vskip 2pt
        \hbox to 0pt{\hss\L@Dc\number\gridcount\Zr@s\hss}\vss}%
      \wd\auxGridbox@=0pt \box\auxGridbox@
   \vvrule
   }

 \def\PlaceGrid@@{\gridcount=10 
  \setbox\hGridbox@=\hbox{%
        \hbox{%
             \hskip-.4pt\vrule
             \vbox to \Htt@{%
               \offinterlineskip\parindent=\z@\relax
               \hbox to \Wdd@{\hfil}
               \hRule@\hRule@\hRule@\hRule@
               \vfil\hhrule\vfil}%
             \vrule\hskip-.4pt}
    }%
  \gridcount=0%
  \setbox\vGridbox@=\hbox{%
      \vbox{\offinterlineskip\parindent=0pt\hsize=0pt
         \vskip-.4pt\hrule%
         \hbox to \Wdd@{%
                 \vtop to \Htt@{\vfil}%
                 \vRule@\vRule@\vRule@\vRule@
                 \hfil\vvrule\hfil}%
         \hrule\vskip-.4pt}}%
  \wd\hGridbox@=0pt\ht\hGridbox@=0pt
  \wd\vGridbox@=0pt\ht\vGridbox@=0pt
  \hbox{\box\hGridbox@\box\vGridbox@}%
  }

 \def\LabelsGlobal{\def\LabGl@b{\global}}
 \def\LabelsLocal{\def\LabGl@b{}}
 \LabelsGlobal 

 \def\SetLabels#1\endSetLabels{%
   \LabGl@b\Labeltoks@={#1()\\}%
   }

 \LabGl@b\Labeltoks@={()\\}

 \def\ShowGrid{\LabGl@b\let\PlaceGrid@\PlaceGrid@@}
 \def\HideGrid{\LabGl@b\let\PlaceGrid@\relax}
 \def\Grids{\ShowGrid\LabGl@b\let\GridSwitch@\ShowGrid}
 \def\noGrids{\HideGrid\LabGl@b\let\GridSwitch@\HideGrid}

 \noGrids

 \def\bAdjust@@{%
     \setbox\auxLabelbox@=\hbox{\raise \dp\auxLabelbox@
            \box\auxLabelbox@}}
 \def\bAdjust@{\let\vAdjust@\bAdjust@@}

 \def\eAdjust@@{\dimen0=-.5\ht\auxLabelbox@
     \advance\dimen0 by .5\dp\auxLabelbox@
     \setbox\auxLabelbox@=
            \hbox{\raise\dimen0\box\auxLabelbox@}}
 \def\eAdjust@{\let\vAdjust@\eAdjust@@}

 \def\tAdjust@@{%
     \setbox\auxLabelbox@=\hbox{\raise-\ht\auxLabelbox@
            \box\auxLabelbox@}}
 \def\tAdjust@{\let\vAdjust@\tAdjust@@}

 \let\vAdjust@\relax

 \def\lAdjust@{\let\hAdjust@\rlap}
 \def\rAdjust@{\let\hAdjust@\llap}

 \let\hAdjust@\relax\let\vAdjust@\relax

 \def\FetchLabel@#1(#2)#3\\{%
     \IN@0#2@@\ifIN@
        \setbox0=\hbox{\ignorespaces#1#3\unskip}%
        \ifdim\wd0>0pt
           \ms@g{}%
           \ms@g{ !!! Bad label(s)? !!!}%
           \message{ #1(#2)#3}%
        \fi
        \def\LabelMole@##1\endFetchLabel@{%
            \IN@0()\\@##1@%
            \ifIN@\def\Temp@{\FetchLabel@##1\endFetchLabel@}%
            \else\def\Temp@{}%
            \fi
            \Temp@
           }%
     \else
       \ignorespaces#1\unskip
       \setbox\auxLabelbox@=%
         \hbox to 0pt{\hss\ignorespaces\hAdjust@
          {\ignorespaces#3\unskip}\hss}%
       \vAdjust@
       \let\hAdjust@\relax\let\vAdjust@\relax
       \AugmentLabelBox@@{#2}%
       \ht\Labelbox@=0pt\dp\Labelbox@=0pt
       \let\LabelMole@\FetchLabel@%
     \fi\LabelMole@}

 \newtoks\XYSep@ 
 \def\SetXYSeparator#1{%
     \IN@0#1@@\ifIN@\XYSep@{*}%
     \else
     \XYSep@{#1}%
     \fi
     }

 \SetXYSeparator*

 \def\AugmentLabelBox@@#1{%
     \IN@0\the\XYSep@ @#1@\ifIN@
       \SPLIT@0\the\XYSep@ @#1@%
       \setbox\Labelbox@=\hbox to 0pt{%
         \unhbox\Labelbox@
         \Shifted@@{\the\Initialtoks@\Wddd@}%
         {\the\Terminaltoks@\Httt@}%
         {\box\auxLabelbox@}}%
     \else
         \ms@g{}%
         \ms@g{ !!! Bad insertion point. !!!}%
         \message{ (#1\ this point was rejected.)}%
     \fi
    }

 \def\FetchOption@#1[#2]#3\endFetchOption@{%
    \def\temp{#1}
    \ifx\temp\empty
       \Edgect@=#2\relax
       \let\EdgGl@b\relax
       \SetEdges@
       \Cleaner@#3%
    \fi}

 \def\Cleaner@#1[@]{\Labeltoks@{#1}}
     
 \def\PlaceLabels@@{\mathsurround=0pt
     \def\Cr@{\\}%
     \let\L\lAdjust@\let\R\rAdjust@
     \let\B\bAdjust@\let\E\eAdjust@\let\T\tAdjust@
     \expandafter\FetchOption@\the\Labeltoks@[@]\endFetchOption@
     \Wddd@=\Wdd@ \Edgescale@\Wddd@ 
     \Httt@=\Htt@ \Edgescale@\Httt@
     \expandafter\FetchLabel@\the\Labeltoks@\endFetchLabel@
     \box\Labelbox@
     }%

 \let \PlaceLabels@\PlaceLabels@@

 \def\AffixLabels#1{\setbox\Coordinatebox@=\hbox{#1}%
      \Wdd@=\wd\Coordinatebox@ \Htt@=\ht\Coordinatebox@
      \advance\Htt@ \dp\Coordinatebox@
      \hbox{\copy\Coordinatebox@\kern-\Wdd@ 
           \Shifted@@{0pt}{-\dp\Coordinatebox@}%
           {\PlaceLabels@\PlaceGrid@}%
           \kern\Wdd@}%
      \GridSwitch@ 
      \LabGl@b\Labeltoks@{()\\}%
      }
 
   \let\wlog\LabelFigwlog@ld   
   \catcode`\@=\LabelFigCatAt  

                                By

              Raymond S\'eroul <A18645@FRCCSC21.BITNET>
                                and 
              Laurent Siebenmann <lcs@topo.math.u-psud.fr>
    
              VERSIONS: July 1991, Oct 1991, Jan 1992, July 1992

INTRODUCTION

      This labelling package is intended for TeX users who
rely on non-TeX sources for for their graphics inserts.  It
provides means for adding TeX labels to such inserts with a
minimum of fuss. 

       For most labels, TeX users have in the past found it
reasonably convenient to rely on non-TeX sources. Typical
occasions when an inescapable need for TeX labels seemed to
arise are

 (a) when the graphics program lacks certain exotic or complex
mathematical symbols

 (b) when the very highest typographical quality is wanted for the
labels

 (c) when labels included with the graphics fail to print, 
 and you cannot figure out why (cf. boxedeps.doc).  The labels
 provided by labelfig.tex are 100

       Since this package first appeared, many users, who in the
past scarcely dreamed of using TeX labels, have come to use
nothing but.  So it is now appropriate to add

Intoxication Warning:  TeX labels may be addictive and expensive. 

     If you have a fast preview you may disagree, and even find
that this package provides an agreeable paste-up environment; see
extra applications at end.

     Note to publishers: It is possible and convenient to ultimately
export the TeX labels produced by labelfig.tex to become an integral
part of the EPS file. This is often desired by a publisher who typically
uses an "upmarket" graphics or page layout program, with which the
staff is skilled in perfecting figures.  See Appendix I for
a recipe.

     The authors are grateful to Patrick Ion of Math Reviews for
helpful comments and encouragement.

BASIC INSTRUCTIONS

    After reading in the macro file using

preview or proof your figure with a coordinate grid printed on
top, by typing the following:

    \ShowGrid  
    \AffixLabels{<the graphics insertion>}

Here <the graphics insertion> is what you would type to insert
the graphics object alone without the grid.  This must provide
for the space around it. For example <the graphics insertion>
might well be \BoxedEPSF{MyFigure scaled 700} using the
boxedeps.tex macro package (from same source); this provides a
TeX box containing the encapsulated PostScript insert specified by
the file MyFigure. \AffixLabels{...} provides the grid (supposing
\ShowGrid is present) and later, once you have specified labels
using the grid, it will "tack on" the labels.

     The grid is a sort of (usually elongated) checkerboard of
ten rows and ten columns and its (internal) partitions are by
default numbered  .1, ... ,.9  both horizontally (X-coordinate
running left to right) and vertically (Y-coordinate running bottom
to top).  Thus the points enclosed by the grid correspond to the
points of the unit square in the cartesian "X-Y" plane, the lower
left corner corresponding to the origin (0,0).  By extrapolation,
the full page corresponds to a larger rectangle in the plane.

     These coordinates serve to position labels as follows.
Before the \AffixLabels{...} command type label specifications:

  \SetLabels
   (<X-coordinate>*<Y-coordinate>) <first label> \\
   .
   .
   .
   (<X-coordinate>*<Y-coordinate>)  <last label> \\
  \endSetLabels

Each row specifies one label and is terminated by \\.  In each
row, the position indicator comes first; it is written as a
standard cartesian point except that the X- and Y- coordinates
are separated by * rather than a comma because TeX allows a
comma as decimal point. There are no dimension units to specify
as the unit is the grid itself.

     By default, this cartesian point specifies where the middle
of the baseline of the label will be located.  However if you precede
the point by \L [or \R] the left [or right] edge of the baseline will
be located there. Similarly you may also precede the point by \T, \E,
or \B to vertically align the top equator or bottom of the label box
at the specified point.  This gives nine standard positions of
the label with respect to the insertion point --- corresponding to
the eight principle points of the compas and the center

                     \L\T     \T      \R\T

                     \L\E     \E      \R\E

                     \L\B     \B      \R\B

But this neglects the default "baseline" level of TeX,
giving potentially three more positions

                     \L    <no tag>   \R

For text, the baseline level is often the preferred. Its relation to
the others is variable. It will often coincide with the bottom level,
as happens for "X".  But it is often distinct, as for "g", in which
case you have in all 12 distinct positions rather than 9.

     It is convenient to think of this specification of label
position as attaching the label by a thumb-tack to the coordinate
grid. There are up to twelve positions of the thumb-tack on the
label, while the position of the thumb-tack on the coordinate grid is
arbitrary.  Normally, one choses the position of the thumb-tack on
the label to be the one that is the closest to the item being
labeled.  There are good reasons for this "rule of thumb":

   (a)  It facilitates correct positioning at first try.

   (b)  If the scale of the figure must be altered after labels
have been affixed, the labels have a good chance of remaining well
positioned.

   (c)  The visible grid need not extend beyond the "bounding box"
for the figure, because the best preferred position is always
(at least almost) within the bounding box .

The second reason is particularly important. Indeed it often
happens that scale has to be altered after labelling begins, in
order to either provide space for the labels, or to adjust
proportions between the labels and the figure.  (The size of labels
is unaffected by scaling.)

     Here is an artificial but self-contained test which uses
TeX rules to make a graphics object.

TEST

    Do not skip this!

 \def\FrameIt#1{\hbox{\vrule$\vcenter {\hrule\kern3pt%
             \hbox {\kern3pt #1\kern3pt}%
               \kern3pt\hrule}$\relax\vrule}}

 \def\Caption#1#2{\FrameIt{%
       \vtop {\hsize=#1\relax \parindent=0pt
         \leftskip=0pt \rightskip=0pt plus15pt
         \parfillskip=0pt
         \lineskip=1pt\baselineskip=0pt
         #2}}}

 \def\FirstQuadrant{\hbox to 100pt{\vrule\vbox to 100pt{%
        \hbox to 100pt{\hfil}\vfil\hrule}\hss}}

  \SetLabels
    \R(.5*.2) $\zeta\,\cdot$\\
    (.9*-.10) $\xi$\\
    \R(-.03*.9) $\eta$\\
    \T(.5*.9) \Caption{70pt}{%
          \it The norm of
          $g(\xi+i\eta)$ is indicated on
          contours of this invisible surface.}\\
  \endSetLabels

  \AffixLabels{\FirstQuadrant}

  \end

  Note that the coordinates to use for labels are indicated on the
edges of the grid (when visible) corresponding to the conventional
x- and y- axes of the Cartesian plane. By default the grid is
1-by-1. However, by the command \Edges{100}, you can change this
to 100-by-100 and many users find this alternative most
convenient. Place the command \Edges{...} in your style file (or
header) since its effect is is global. Other possible edge values
are 10 and 1000.

  If you use the command \Edges{...} at all, do so with care.  For
if you accidentally delete an \Edges{...} command your labels will
abruptly be badly misplaced and may logically but mysteriously
generate "dimension too big" errors under TeX and "off page" errors
under your driver.  

  You can dictate the edgescale for an individual figure by giving
the scale in brackets immediately after \SetLabels.  Thus, to
import into an article using say \Edge{100} a figure labelled using
another edgescale, say the original 1-by-1 default, you can use
\SetLabels[1]...\endSetLabels.

GETTING IT DOWN PAT

     Complicated labeling deserves the same respect as
complicated mathematics.  Do not expect it to come out perfect the
first time!  What is needed in either case is a mechanism to
repeatedly typeset troublesome pieces.

     One mechanism is always available.  One does complicated
labelling in a separate "test" file involving just the figure being
labelled;  a texpert will know how to \dump TeX's current state as
a temporary format that restarts rapidly at each retry.  Usually,
one then pastes the completed labelled figure back into the main
TeX file, but, of course, one can also \input it as an auxiliary
file.

     If you do not have a TeXpert at handy, here is a first
approximation to an efficient setup. By deletions reduce a copy
of your article to just a few lines before and after the figure.
Now label the figure, and finally, copy and paste the labelled
figure to the original article. Then copy the next figure to label
into this testbed and repeat. The TeXpert can improve the  speed
at which TeX starts up, by compiling a format specifically for
your article; just one caution: best NOT include in the format
ephemeral details of setup like \Set<mydriver>ArtSpecials (from
boxedeps.tex because this reads  figure dimensions which you may
change during your work session.

     An improved mechanism to repeatedly typeset troublesome
pieces is now available on the Macintosh; it is called LinoTeX;
see the same ftp sources.  It could be set up on many types
of computer.

     Before using labelfig.tex to attach labels to a graphics
object inserted using boxedeps.tex or BoxedArt.tex, make it a
firm rule to carefully adjust the bounding box using the trimming
commands of these packages, and also at least tentatively scale
and position the object. Beware of changing the grid inadvertently
after the labels have been positioned.  For example, correcting
the bounding box of a PostScript graphics object can foul up the
labels by changing the coordinate grid to which the labels are
attached. This is particularly true for the trimming  commands of
boxedeps.tex and BoxedArt.tex. However, as noted already, change
of scale is much less disruptive, and modest adjustments should be
well tolerated.

     Sometimes the labels protrude so far from the bounding box
of a figure that the figure has to be repositioned.  Best do this
by ad hoc spacing, say using \hglue and \vglue; altering the
bounding box would create a vicious circle.

     Remember that you are responsible for preventing labels
from overlapping. You are responsible for all label typography
including size and style. A label is really just about anything
that can be put in a TeX box. Note that spaces at the beginning
and end of labels will normally be suppressed; if you really want
them you must protect them with TeX braces.

     This package temporarily sets the \mathsurround parameter
of TeX to zero  while the labels are being affixed. This is done
because nonzero \mathsurround space would influence the position
of left and right aligned labels; then, when a texpert or printer
modifies mathsurround, diagram labeling might be disastrously
altered. There is a small price to pay involving labels that are
formatted as caption boxes including mathematics: you  may want or
need to specify an explicit mathsurround space within the caption
box; it will not influence anything outside.

     Those hostile to the use of * as separator between
the X and Y coordinates of label insertion points, are free to
impose another using \SetXYSeparator{<the new separator>}.  
Americans may prefer "," to "*" since they never use a 
comma as a decimal point; on the other hand, * may be more visible.

APPENDIX (I)  MERGING labelfig.tex LABELS INTO AN EPSF GRAPHICS OBJECT.

     As promised in the introduction, here is a recipe useful for
publishers. It works at least on Macintosh and at least for vectorized
graphics and Adobe type1 fonts.  (There is surely a similar recipe for
PCs under MSWindows.)

 (a)  Use boxedeps.tex utility to integrate the figure given by the eps
file, "x.eps" say, with a visible frame around it.  See
\ShowDisplacementBoxes command in boxedeps.tex.  To get precise results
automatically it is important to use the \Trim... commands of
boxedeps.tex making the "DisplacementBox" neatly fit the figure.

 (b)  Use the TeX printer driver and LaserWriter (versions >= 8.1.1) to
export to an EPSF the DVI page containing the integrated, labelled
figure. You now have an EPS file  "xx.eps"  that contains too much, and at
the wrong scale, and at wrong position.

 (c)  Convert the EPSF to an Adode Illustrator format EPSF using
the shareware utility called epsConvert by Sam Weiss
1993-- (currently $25).

 (d)  In Illustrator (or a compatible program), group the labels and the
"DisplacementBox"; copy them to the clipboard and paste them into "x.ps".
This step requires that all the label fonts be "visible to the Macintosh.

 (e)  Translate and scale the pasted group consisting of the labels plus
the "DisplacementBox" so as to make the "DisplacementBox" the bounding
box of (labelless) figure represented by "x.eps".  At this point the
labels will be correctly placed on the figure "x.eps".

 (f)  Ungroup and delete the "DisplacementBox".  The result is the
desired single EPS file, "x+.eps" say, It contains the original figure
plus its labels.  

     Using grouping and ungrouping appropriately in "x+.eps", a
publisher's staff can very efficiently improve label positions etc.

APPENDIX II)  SOME EXOTIC APPLICATIONS

     The grid of labelfig.tex is analogous to a light-table in
classical page makeup with wax or latex glue.  In principle, you
can use it to compose any page from its indivisible parts.  This
even has some of the artisanal charm of classical paste-up
provided you have a fast screen preview to make the process
"interactive".

     In practice labelfig.tex is a tool for nonstandard jobs.
Here are a few going beyond the labelling already discussed.

(I)  GRAPHICS INTEGRATION.

     This is accomplished by treating the imported graphics
objects as labels.  The underlying graphics object is then
typically an empty  \vbox to <dimension>{\vfill} in a TeX
\midinsert...\endinsert construction.  A label line
might be of the form

   (.1*.1) \\

The exact form of the special command varies from driver to
driver.  However, in the case of encapsulated PostScript graphics
(EPSF norm), by relying on boxedeps.tex, one can have the
following standard syntax (independant of driver  (see
boxedeps.doc for details.
  
  (.1*.1) \BoxedEPSF{MyFigure scaled <scale in mils>}\\

This may be slow since it requires TeX to read the PostScript
file to read bounding box using many complex macros.  So you
may want to try

  (.1*.1) \EPSFSpecial{MyFigure}{<scale in mils>}\\

which is fast and driver independant, but it squashes the
bounding box, normally to its lower left corner.

     Similarly for graphics of the Macintosh PICT norm ---
using BoxedArt.tex (same sources) in place of boxedeps.tex.

     This approach to integration is to be recommended when
one is assembling a composite graphics object.

 (II)  COMMUTATIVE DIAGRAM ENHANCEMENT

     Commutative diagrams or arrays of mathematical objects
connected by arrows of various sorts are common in mathematics.
The mathematical objects require the use of TeX.  Recently TeX
acquired a good collection of arrows of all slopes --- that of
LamSTeX --- plus pwerful macros to build the diagrams.

     However, even the LamSTeX collection is often
inadequate; it lacks for example double shafted arrows, dotted
arrows and curved arrows. Fortunately it is possible to produce
such arrows on an individual basis using sophisticated graphics
programs such as Illustrator and AldusFreehand (both serving
the EPSF norm) or using Metafont (with its public domain norm).
Since the creation of each new arrow is a work of love, you
probably want to limit the number of arrows by using LamSTeX
for most arrows. The 40K commutative diagram module of LamSTeX
has been adapted to work with AmSTeX and a copy may be posted
with LabelFig and related files. Unfortunately no one has yet
offered a version that works with Plain TeX or LaTeX.

       Suffice it here to say that when the exotic arrow has
been somehow imported into TeX, labelfig.tex treats it as a
label that one affixes to the commutative diagram.  Two other
steps will be treated in separate notes, namely the matter of
extracting the dimension specifications for the arrow and the
construction of the arrow --- for these steps are far from
unique and often depend intimately on your computer environment. 
Notes for the Macintosh-Textures-Illustrator combination are
found in the file ExoticArrows.doc.

 (III) NESTING 

Ingenuity pays off in exploiting labelfig.tex. One can
mix graphics and typography quite freely.  labelfig.tex is good
for freeform or overlapping arrangements, while boxedeps.tex (or
BoxedArt.tex) is best for regimented non-overlapping
arrangements --- and the two can be combined.

     The default behavior of labelfig.tex is not ideal 
for nesting objects, because to prevent trouble for beginners
the register for labels is globally cleared when \AffixLabels
concludes.  But there are switches available

      \LabelsGlobal      \LabelsLocal

which change this.  To understand this, extend the above test 
by something like:

 \LabelsLocal

 \SetLabels
    (.5*.5) AAA\\
 \endSetLabels

 {
 \SetLabels
    (.5*.5) ZZZ\\
 \endSetLabels
   \AffixLabels{\FirstQuadrant}
 }

   \AffixLabels{\FirstQuadrant}

     There are however potential pitfalls.  Neither
labelfig.tex nor boxedeps.tex has been tested under extreme
conditions. Problems may occur if their procedures are
indiscriminately nested. For boxedeps.tex (not labelfig.tex)
there is a precise cause for worry, namely many of its
variables are "global", which means that TeX braces will not
provide the protection one might expect.

COMMAND SUMMARY FOR labelfig.tex

  Here [...] means optional (one or zero)
       [...]* means any number of such constructs

  \SetLabels
    [[<P>](<X><Sep><Y>) <label> \\]*
  \endSetLabels
  \ShowGrid  
  \AffixLabels{<the figure>}

   --- <P> is tack position, one of eleven or empty
              order irrelevant

                   \L\T      \T      \R\T

                   \L\E      \E      \R\E

                     \L               \R

                   \L\B      \B      \R\B

   --- (<X><Sep><Y>) insertion point;
  <Sep> is separator, = * by default;
  \SetXYSeparator{<Sep>} changes it.
   <X> and <Y> are real numbers

  --- <label> a label to attach 

  --- <the figure> the figure to label 

  \GlobalLabels (default)     
  \LocalLabels  setting for nested constructs.

 \Grids makes ALL grids appear; \HideGrid then makes just next disappear.
 \noGrids returns to default.  The commands are always global.

 \GridLineWidth{<dimension>} adjusts width of grid lines. Default is very
small, to give "hairline" effect. If your grid lines are missing try
setting \GridLineWidth{1pt}.

 \Edges#1 globally changes the edge size of all grids to the numerical 
value #1, which must be 1, 10, 100, or 1000.  The default is 1.

VERSION HISTORY.
 --- Jan 1993: \Edges#1 and [??] option after \SetLabels
 --- July 1992: \Grids, \noGrids, \HideGrid;
       Gridlines become hairlines; \GridLineWidth{<dimension>}.
 --- Oct 1991, Jan 1992: \SetXYSeparator{<Sep>},  \LabelsGlobal,
       \LabelsLocal.
 --- July 1991: first release

Address for bugs and other feedback:

        Raymond S\'eroul
        IREM and Lab. de Typographie Informatise
        Univ. Rene Descartes
        Strasbourg

    Tel 33-88-41-63-45
    Email:  A18645@FRCCSC21.BITNET

        Laurent Siebenmann
        Mathematique, Bat. 425,
        Univ de Paris-Sud,
        91405-Orsay,
        France

    Tel 33-1-6941-7949; 
    Email: lcs@topo.math.u-psud.fr  

\usepackage{subfigure}
\usepackage{amsmath,amssymb,amsfonts,latexsym,verbatim,color,epsfig,psfrag}
\usepackage{dsfont}
\usepackage{times,multirow,multicol, array}
\usepackage{algorithm,algorithmic}
\usepackage{setspace}
\usepackage{float}
\usepackage{caption}
\usepackage{adjustbox}

\newtheorem{lemma}{Lemma}
\newtheorem{theorem}{Theorem}
\newenvironment{customthm}[1]
  {\renewcommand\thetheorem{#1}\theorem}
  {\endtheorem}
\newenvironment{customlem}[1]
  {\renewcommand\thelemma{#1}\lemma}
  {\endlemma}
\newcolumntype{L}[1]{>{\raggedright\let\newline\\\arraybackslash\hspace{0pt}}m{#1}}
\newcolumntype{C}[1]{>{\centering\let\newline\\\arraybackslash\hspace{0pt}}m{#1}}
\newcolumntype{R}[1]{>{\raggedleft\let\newline\\\arraybackslash\hspace{0pt}}m{#1}}
\def\tcr{\textcolor{red}}

\newcommand {\Ebb}{{\mathbb{E}}}

\icmltitlerunning{Diversity Actor-Critic: Sample-Aware Entropy Regularization for Sample-Efficient Exploration}

\begin{document}

\twocolumn[
\icmltitle{Diversity Actor-Critic: Sample-Aware Entropy Regularization for Sample-Efficient Exploration}




\begin{icmlauthorlist}
\icmlauthor{Seungyul Han}{kaist}
\icmlauthor{Youngchul Sung}{kaist}
\end{icmlauthorlist}

\icmlaffiliation{kaist}{Department of Electrical Engineering, Korea Advanced Institute of Science and Technology, Daejeon, South Korea}
\icmlcorrespondingauthor{Youngchul Sung}{ycsung@kaist.ac.kr}

\icmlkeywords{Machine Learning, ICML}

\vskip 0.3in
]



\printAffiliationsAndNotice{}  

\begin{abstract}
In this paper, sample-aware policy entropy regularization is proposed to enhance the conventional policy entropy regularization for better exploration.   
Exploiting the  sample distribution obtainable from the replay buffer, the proposed sample-aware entropy regularization   maximizes the entropy of the weighted sum of the policy action distribution and the sample action distribution from the replay buffer for sample-efficient exploration. A practical algorithm named diversity actor-critic (DAC) is developed by applying policy iteration to the objective function with the proposed sample-aware entropy regularization. 
Numerical results show that DAC significantly outperforms existing recent algorithms for  reinforcement learning. 
\end{abstract}

\section{Introduction}
\label{sec:intro}

Reinforcement learning (RL) aims to maximize the expected return under Markov decision process (MDP) \cite{sutton1998reinforcement}. When the given task is complex, e.g., the environment has high action-dimensions or  sparse rewards, it is  important to  explore state-action pairs well  for high performance  \citep{agre1996computational}.
For better exploration, recent RL considers various methods: maximizing the policy entropy to take actions more uniformly \citep{ziebart2008maximum,fox2015taming,haarnoja2017rein},
maximizing diversity gain that yields intrinsic reward to explore rare states by counting the number of visiting states \citep{strehl2008analysis,lopes2012exploration}, maximizing information gain \citep{houthooft2016vime,hong2018diversity}, maximizing model prediction error \citep{achiam2017surprise,pathak2017curiosity}.
In particular, based on policy iteration for soft Q-learning, \citet{haarnoja2018soft} extended maximum entropy RL and proposed an  off-policy actor-critic algorithm,
soft actor-critic (SAC), which has competitive performance for challenging continuous control tasks.

In this paper, we consider the problem of policy entropy regularization  in off-policy learning and propose a generalized approach to policy entropy regularization for sample-efficient exploration. In off-policy learning, we store samples in the replay buffer and reuse old samples to update the current policy \citep{mnih2015human}. Thus, the sample buffer has information about the old samples. 
However, the simple policy entropy regularization tries to  maximize the entropy of the current policy irrespective of the distribution of the previous samples in the replay buffer.  
In order to exploit the sample information in the replay buffer and enhance performance, we propose sample-aware entropy regularization, which tries to maximize the entropy of the weighted sum of the current policy action distribution and the sample action distribution from the replay buffer. 
We develop a practical and efficient algorithm for return maximization based on the proposed sample-aware entropy regularization without explicitly computing the replay-buffer sample distribution, and demonstrate that the proposed algorithm yields significant enhancement in exploration and final performance on various difficult environments such as tasks with sparse reward or high action dimensions.

\section{Related Works}
\label{sec:related}

{\bf Entropy regularization:} Entropy regularized RL maximizes the sum of the expected return and the policy action entropy. It encourages the agent to visit the action space uniformly for each given state, and  can provide more accurate model prediction  \citep{ziebart2010modeling}. Entropy regularization is considered in various domains: inverse reinforcement learning \citep{ziebart2008maximum}, stochastic optimal control problems \citep{todorov2008general,toussaint2009robot,rawlik2013stochastic}, and off-policy reinforcement learning \citep{fox2015taming,haarnoja2017rein}. 
\citet{nachum2017bridging} showed that there exists a connection between value-based and policy-based RL under entropy regularization.  \citet{o2016combining} proposed an algorithm combining value-based and policy-based RL, and \citet{schulman2017equivalence} proved that they are equivalent. \citet{hazan2019provably} maximized the entropy of state distribution induced by the current policy by using state mixture distribution for better pure exploration.

{\bf Diversity gain:} Diversity gain is used to provide a guidance for exploration to the agent. To achieve diversity gain, many intrinsically-motivated approaches and intrinsic reward design methods have been considered, e.g.,
intrinsic reward based on curiosity \citep{chentanez2005intrinsically, baldassarre2013intrinsically}, model prediction error \citep{achiam2017surprise,pathak2017curiosity,burda2018exploration}, divergence/information gain \citep{houthooft2016vime,hong2018diversity}, counting  \citep{strehl2008analysis,lopes2012exploration,tang2017exploration,martin2017count}, and unification of them \citep{bellemare2016unifying}. \citet{eysenbach2018diversity} explicitly maximized diversity based on mutual information.

{\bf Off-policy learning:} Off-policy learning  reuses samples generated from behaviour policies for  policy update \citep{sutton1998reinforcement,degris2012off}, so it is sample-efficient  compared to on-policy learning.  In order to reuse old samples, a replay buffer that stores trajectories generated by previous policies is used for Q-learning \citep{mnih2015human,lillicrap2015continuous,fujimoto2018addressing,haarnoja2018soft}. To further enhance both stability and sample efficiency, several methods were considered, e.g., combining on-policy and off-policy   \citep{wang2016sample,gu2016q,gu2017interpolated}, and generalization from on-policy to off-policy \citep{nachum2017trust,han2019dimension}.

\section{Background}
\label{sec:background}

\textbf{Setup:} ~We assume  a basic RL setup composed of an environment and an agent. The environment  follows an infinite horizon Markov decision process $(\mathcal{S},\mathcal{A},P,\gamma,r)$, where $\mathcal{S}$ is the state space, $\mathcal{A}$ is the action space, $P$ is the transition probability, $\gamma$ is the discount factor, and $r : \mathcal{S} \times \mathcal{A} \rightarrow \mathbb{R}$ is the reward function. In this paper, we consider  continuous state and action spaces. The agent has  policy distribution $\pi \in \Pi : \mathcal{S} \times \mathcal{A}\rightarrow [0,\infty)$ which selects an action $a_t$ for  given state $s_t$ at time step $t$, where $\Pi$ is the policy space.  Then, the agent receives reward $r_t := r(s_t,a_t)$ from the environment and the state changes to $s_{t+1}$. Standard RL aims to maximize the discounted return $\mathbb{E}_{s_0\sim p_0, \tau_0\sim \pi}[\sum_{t=0}^{\infty} \gamma^t r_t]$, where $\tau_t=(s_t,a_t,s_{t+1},a_{t+1}\cdots)$ is an episode trajectory.

\textbf{Soft Actor-Critic:} ~Soft actor-critic (SAC) includes a policy entropy regularization term in the objective function  with the aim of performing more diverse actions for each given state and visiting states with higher entropy for better exploration \citep{haarnoja2018soft}. The entropy-augmented policy objective function of SAC is given by 

\vspace{-1em}
\begin{equation}\label{eq:sacobj}
J_{SAC}(\pi) = \mathbb{E}_{\tau_0 \sim \pi}\left[\sum_{t=0}^{\infty}\gamma^t(r_t + \beta \mathcal{H}(\pi(\cdot|s_t)))\right],
\end{equation}
where $\mathcal{H}$ is the entropy function and $\beta \in (0,\infty)$ is the entropy coefficient.
SAC is a practical off-policy actor-critic algorithm based on soft policy iteration (SPI) that alternates soft policy evaluation to estimate the true soft $Q$-function and soft policy improvement to find the optimal policy that maximizes \eqref{eq:sacobj}.  
SPI theoretically guarantees convergence to the optimal policy that maximizes \eqref{eq:sacobj} for finite MDPs.

\section{The Diversity Actor-Critic Algorithm}
\label{sec:dac}

\subsection{Motivation of Sample-Aware Entropy}
\label{subsec:motivation}

In order to guarantee the convergence of Q-learning, there is a key assumption: \emph{Each state-action pair must be visited infinitely often} \citep{watkins1992q}. 
Without proper exploration, policy can converge to local optima and  task performance can be degraded severely \citep{plappert2017parameter}.  Therefore, exploration for visiting diverse state-action pairs is important for RL.  There has been extensive research for better exploration in RL. One important line of recent methods is to use intrinsic reward based on prediction model \citep{chentanez2005intrinsically, baldassarre2013intrinsically, achiam2017surprise,pathak2017curiosity,burda2018exploration}. In this approach, we have a prediction model for a target value or distribution, and the  prediction model is learned with samples. Then, the prediction error is used as the intrinsic reward added to the actual reward from the environment, and the discounted sum of the actual and intrinsic rewards is maximized. The fundamental rationale  behind this approach is that the  prediction model is well learned for the frequently-observed state-action pairs in the sample history and hence the  prediction error is small. On the other hand, for unobserved or less-observed  state-action pairs in the sample history, the  prediction model training is not enough and the prediction error is large. In this way, un- or less-explored state-action pairs are favored.

Another successful method for exploration is policy entropy regularization  with the representative method shown in \eqref{eq:sacobj}. In \eqref{eq:sacobj}, one can view the policy action entropy $\Hc(\pi(\cdot|s_t))$ as an intrinsic reward added to the actual reward $r_t$. 
This method relies on the fact that the entropy attains maximum when the distribution is uniform \citep{Cover:book}. Thus, maximizing the discounted sum of the actual reward $r_t$ and the policy action entropy as in \eqref{eq:sacobj} yields a policy that tries not only to maximize the actual reward but also to visit states with high action entropy and to take more uniform actions for better exploration. Furthermore, in this method the weighting factor $\beta$ in \eqref{eq:sacobj} can be learned adaptively based on the Lagrangian method to maintain a certain level of entropy \cite{haarnoja2018soft2}.  However, on the contrary to the  prediction model-based method,  the entropy regularization  method  does not exploit the previously-observed sample information to construct the intrinsic reward at current time $t$ since the intrinsic reward $\Hc(\pi(\cdot|s_t))$ depends only on the policy $\pi(\cdot|s_t)$, and $\pi(\cdot|s_t)$ for given $s_t$ does not directly capture the sample distribution information from the replay buffer.

In this paper, we consider the maximum entropy framework in off-policy learning  and extend this framework by devising an efficient way to exploit the sample information in the replay buffer so as to harness the merits of the two aforementioned approaches: taking more uniform actions and promoting un- or less-performed actions in the past.

\subsection{Proposed Policy Objective Function}

In order to use the previous sample information in entropy-based exploration, we first define the mixture  distribution
\begin{equation}
q_{mix}^{\pi,\alpha}(\cdot|s):=\alpha \pi(\cdot|s) + (1-\alpha)q(\cdot|s),
\end{equation}
where $\alpha\in [0,1]$ is the weighting factor, $\pi(\cdot|s)$ is the policy (action) distribution, and $q(\cdot|s)$ is the sample action distribution of the replay buffer $\mathcal{D}$ which stores previous samples.
Then, we propose maximizing  the following objective function  

\vspace{-2em}
\begin{equation}\label{eq:objpi}
J(\pi) =  \mathbb{E}_{\tau_0\sim \pi}\left[\sum_{t=0}^\infty \gamma^t (r_t + \beta \mathcal{H}(q_{mix}^{\pi,\alpha}(\cdot|s_t)))\right],
\end{equation}
where we refer to $\Hc(q_{mix}^{\pi,\alpha}(\cdot|s_t))$ as  the {\em sample-aware entropy}. Note that maximizing the sample-aware entropy enhances sample-efficient exploration because in this case the learning guides the policy  to choose actions so that the mixture distribution $q_{mix}^{\pi,\alpha}(\cdot|s_t)$ becomes uniform. That is, $\pi(\cdot|s_t)$ will choose actions rare in the replay buffer (i.e., the density $q(\cdot|s_t)$ is low) with high probability  and choose actions stored many times in the replay buffer (i.e., the density $q(\cdot|s_t)$ is high) with low probability so as  to make the mixture distribution uniform.  Indeed, we can decompose the sample-aware entropy $\mathcal{H}(q_{mix}^{\pi,\alpha})$ for given $s_t$  as

\vspace{-1.5em}
{ \footnotesize
\begin{align}
&\mathcal{H}(q_{mix}^{\pi,\alpha}) =\hspace{-0.1em} -\hspace{-0.1em}\int_{a\in\mathcal{A}}\hspace{-0.1em}(\alpha\pi \hspace{-0.1em}+\hspace{-0.1em} (1\hspace{-0.1em}-\hspace{-0.1em}\alpha)q) \log (\underbrace{\alpha\pi \hspace{-0.1em}+\hspace{-0.1em} (1\hspace{-0.1em}-\hspace{-0.1em}\alpha)q}_{=q_{mix}^{\pi,\alpha}})\label{eq:HmixDecomp2}\\
&=\int \alpha \pi \log \frac{\alpha \pi}{\alpha\pi + (1-\alpha)q} +\int (1-\alpha)q \log \frac{(1-\alpha)q}{\alpha\pi + (1-\alpha)q}\nonumber\\
&~~~~-\int \alpha \pi \log (\alpha \pi) - \int (1-\alpha)q \log ((1-\alpha)q)
\label{eq:HmixDecomp1}\\
&=D_{JS}^\alpha (\pi || q)+\alpha \mathcal{H}(\pi)+ (1-\alpha)\mathcal{H}(q)+  \mbox{constant}, \label{eq:HmixDecomp}
\end{align}}where $D_{JS}^\alpha (\pi||q){:=}\alpha\int \pi \log \frac{\pi}{\alpha \pi + (1-\alpha) q} + (1-\alpha)\int q \log \frac{q}{\alpha \pi + (1-\alpha) q}$ is the $\alpha$-skew Jensen-Shannon (JS)-symmetrization of KL divergence \citep{nielsen2019jensen}.  $D_{JS}^\alpha$ reduces to the standard JS divergence for $\alpha=\frac{1}{2}$ and to zero for $\alpha=0$ or $1$. When $\alpha=1$, $\mathcal{H}(q_{mix}^{\pi,\alpha})$ reduces to the simple entropy and the problem reduces to \eqref{eq:sacobj}. When $\alpha\in (0, 1)$, on the other hand, all the first three terms in the right-hand side (RHS) of \eqref{eq:HmixDecomp} remain. Thus, the added regularized term in \eqref{eq:objpi} will guide the policy to have more uniform actions due to $\Hc(\pi)$ and simultaneously to promote  actions away from $q$ due to $D_{JS}^\alpha(\pi||q)$. Thus, the proposed policy objective function \eqref{eq:objpi} has the desired properties. Note that the sample-aware entropy is included as reward  not as an external regularization term added to the discounted return, and this targets optimization for high total sample-aware entropy of the entire trajectory.  (An analytic  toy example showing the efficiency of the sample-aware entropy regularization is provided in Appendix \ref{sec:toy}.) The main challenge to realize  policy design with \eqref{eq:objpi} is how to compute the sample distribution $q$, which is necessary to compute the objective function. Explicit computation of 
 $q$ requires a method such as discretization and counting for continuous state and action spaces. This should be done for each environment and can be a difficult and tedious job for high dimensional environments. Even if such empirical $q$ is obtained by discretization and counting, generalization of $q$ to arbitrary state-action pairs is typically required to actually implement an algorithm  based on function approximation and this makes the problem difficult further.  In the remainder of this paper, circumventing this difficulty, we develop a practical and efficient algorithm to realize \eqref{eq:objpi} without explicitly  computing $q$.

\subsection{Algorithm Construction}
\label{subsec:pisem}

Our algorithm construction for the objective function \eqref{eq:objpi} is based on {\em diverse policy iteration}, which is a modification of the soft policy iteration of \citet{haarnoja2018soft}. Diverse policy iteration is composed of diverse policy evaluation and diverse policy improvement. Note that the sample action distribution $q$ is updated as iteration goes on. However, it changes very slowly since the buffer size is much larger than the time steps of one iteration. Hence, for the purpose of algorithm derivation,  we regard the action distribution $q$ as a fixed distribution in this section. 

\setcounter{equation}{12}

\begin{figure*}[!t]
\vspace{-1.3em}
\begin{align}
&J_{\pi_{old}}(\pi(\cdot|s_t)):=\beta\{\mathbb{E}_{a_t\sim\pi}\left[Q^{\pi_{old}}(s_t,a_t) + \alpha(\log R^{\pi,\alpha}(s_t,a_t) - \log\alpha\pi(a_t|s_t))\right]\nonumber\\
&~~\hspace{11em}+(1-\alpha)\mathbb{E}_{a_t\sim q}\left[\log R^{\pi,\alpha}(s_t,a_t)-\log\alpha\pi(a_t|s_t)\right]\}.\label{eq:pracobj}
\end{align}
\vspace{-2em}
\begin{align}
&\tilde{J}_{\pi_{old}}(\pi(\cdot|s_t)):=\beta\mathbb{E}_{a_t\sim\pi}[Q^{\pi_{old}}(s_t,a_t)+\alpha(\log R^{\pi_{old},\alpha}(s_t,a_t)-\log\pi(a_t|s_t))],  \label{eq:finalObj}
\end{align}
\vspace{-2.1em}
\end{figure*}

\setcounter{equation}{7}

As in typical policy iteration, for diverse policy iteration, we first define 
 the true diverse $Q$-function $Q^\pi$ as 
 
\vspace{-1em}
{\small
\begin{align}
Q^\pi(s_t,a_t)&:= \frac{1}{\beta} r_t  \nonumber\\
&\hspace{-3em}+ \mathbb{E}_{\tau_{t+1}\sim \pi} \biggl[  \sum_{l=t+1}^\infty \gamma^{l-t-1} \biggl(\frac{1}{\beta}r_l+ \mathcal{H}(q_{mix}^{\pi,\alpha}(\cdot|s_l))\biggr)\biggr],  \label{eq:TrueQpiFunc}
\end{align}
}by including the term $\mathcal{H}(q_{mix}^{\pi,\alpha}(\cdot|s_l))$. Since $Q^\pi(s_t,a_t)$ includes  $\mathcal{H}(q_{mix}^{\pi,\alpha}(\cdot|s_l))$, it  seemingly requires computation of $q$, as seen in 
\eqref{eq:HmixDecomp2}-\eqref{eq:HmixDecomp}. In order to circumvent this difficulty, we define the following ratio function: 
\begin{equation}  \label{eq:Rphialpha}
R^{\pi,\alpha}(s_t,a_t) = \frac{\alpha\pi(a_t|s_t)}{\alpha \pi(a_t|s_t) + (1-\alpha)q(a_t|s_t)},
\end{equation}
and express the objective and all required loss functions in terms of the ratio function not  $q$. For this, based on \eqref{eq:HmixDecomp2}, we rewrite  $\mathcal{H}(q_{mix}^{\pi,\alpha}(\cdot|s_t))$ as follows:

\vspace{-1em}
{\footnotesize
\begin{align}
&\mathcal{H}(q_{mix}^{\pi,\alpha}) 
=
\alpha {\mathbb{E}}_{a_t \sim \pi(\cdot|s_t)}[\log R^{\pi,\alpha} (s_t,a_t) - \log \alpha \pi(a_t|s_t)] \nonumber\\
&+(1-\alpha)\mathbb{E}_{a_t\sim q(\cdot|s_t)}[\log R^{\pi,\alpha} (s_t,a_t) - \log \alpha \pi(a_t|s_t)]. \label{eq:HmixInRateFunction2}
\end{align}
}The above equation is obtained by exploiting 
 the entropy definition  \eqref{eq:HmixDecomp2}. As seen in \eqref{eq:HmixDecomp2}, $\Hc(q_{mix}^{\pi,\alpha})$ is the sum of $-\alpha \Ebb_{a_t \sim \pi(\cdot|s_t)}\log q_{mix}^{\pi,\alpha}$ and $-(1-\alpha)\Ebb_{a_t \sim q(\cdot|s_t)}\log q_{mix}^{\pi,\alpha}$, but $\log q_{mix}^{\pi,\alpha}$ in both terms can be expressed as $\log \frac{\alpha \pi(a_t|s_t)}{R^{\pi,\alpha}(s_t,a_t)}$ by the definition of the ratio function  \eqref{eq:Rphialpha}. So, we obtain \eqref{eq:HmixInRateFunction2}. Now, the $\Hc(q_{mix}^{\pi,\alpha})$  expression in the RHS of  \eqref{eq:HmixInRateFunction2} contains only the ratio function $R^{\pi,\alpha}$ and the policy $\pi$, and thus fits our purpose. Note that the expectation $\Ebb_{a_t \sim q(\cdot|s_t)}$ will eventually be replaced by  empirical expectation based on the samples in the replay buffer in a practical algorithm. So, it does not cause a problem. Thus, the added term $\Hc(q_{mix}^{\pi,\alpha})$ in \eqref{eq:objpi} and \eqref{eq:TrueQpiFunc} is fully expressed in terms of the ratio function $R^{\pi,\alpha}$ and the policy $\pi$. Now, we present the diverse policy iteration composed of diverse policy evaluation and diverse policy improvement.

\emph{Diverse policy evaluation:} We first define a diverse action value function estimate $Q: \mathcal{S}\times\mathcal{A}\rightarrow\mathbb{R}$. Then, we define a modified Bellman backup operator acting on $Q$  to estimate $Q^\pi$ as
\begin{equation}\label{eq:bellman}
\mathcal{T}^\pi Q(s_t,a_t) := \frac{1}{\beta}r_t + \gamma \mathbb{E}_{s_{t+1}\sim P} [V(s_{t+1})],
\end{equation}
where $V(s_t)$ is the estimated diverse state value function given by the sum of $\Ebb_{a_t \sim \pi}[Q(s_t,a_t)]$ and $\Hc(q_{mix}^{\pi,\alpha})$ for given $s_t$, i.e.,
{\footnotesize
\begin{align}
\hspace{-0.35em}V(s_t)&\hspace{-0.2em}=\hspace{-0.2em} \mathbb{E}_{a_t\sim \pi} [Q(s_t,a_t)\hspace{-0.2em}+\hspace{-0.2em}\alpha\log R^{\pi,\alpha}(s_t,a_t)\hspace{-0.2em}-\hspace{-0.2em}\alpha\log\alpha\pi(a_t|s_t)] \nonumber\\
& + \hspace{-0.2em}(1\hspace{-0.2em}-\hspace{-0.2em}\alpha)\mathbb{E}_{a_t\sim q}[\log R^{\pi,\alpha}(s_t,a_t) \hspace{-0.2em}-\hspace{-0.2em} \log \alpha\pi(a_t|s_t)], \label{eq:Vestst}
\end{align}
}where we used the expression \eqref{eq:HmixInRateFunction2} for $\Hc(q_{mix}^{\pi,\alpha})$. Note that for the diverse policy evaluation, the policy $\pi$ under evaluation is given. Hence, the ratio function $R^{\pi,\alpha}(s_t,a_t)$  is  given  for given $\pi$ by its definition \eqref{eq:Rphialpha}. Hence,  $V(s_t)$ in \eqref{eq:Vestst} is well defined and thus the mapping $\mathcal{T}^\pi Q(s_t,a_t)$ on the current estimate $Q(s_t,a_t)$ in \eqref{eq:bellman} is well defined. By repeating the mapping $\mathcal{T}^\pi$ on $Q$, the resulting  sequence  converges to $Q^\pi$ . Proof is given in Lemma \ref{lem:eval} in Appendix \ref{sec:proofs}.

\setcounter{equation}{14}

\emph{Diverse policy improvement:} Now, consider diverse policy improvement.  Suppose that we are given  $Q^{\pi_{old}}(\cdot,\cdot)$ for the current policy $\pi_{old}$. (In this diverse policy improvement step, we use the notation $\pi_{old}$ for the given current policy to distinguish from the notation $\pi$ as the optimization argument.)  Then, we construct the diverse policy objective function  $J_{\pi_{old}}(\pi(\cdot|s_t))$ as shown in \eqref{eq:pracobj}, where the notation $\pi$ in \eqref{eq:pracobj} represents the optimization  argument.  $J_{\pi_{old}}(\pi)$ is the policy objective function estimated under $Q^{\pi_{old}}$.  If we replace $\pi_{old}$ in $J_{\pi_{old}}(\pi)$  with $\pi$ and view  state $s_t$ as the initial state, then (\ref{eq:pracobj}) reduces to $J(\pi)$ in \eqref{eq:objpi}. (This can be checked with  \eqref{eq:objpi}, \eqref{eq:TrueQpiFunc}, \eqref{eq:HmixInRateFunction2} and (\ref{eq:pracobj}).)  Note that  $\pi$ in the $R^{\pi,\alpha}$ and  $\log (\alpha\pi)$ terms inside the expectations in \eqref{eq:pracobj} is the optimization argument $\pi$. 
We update the policy from $\pi_{old}$ to $\pi_{new}$ as 
\begin{equation}\label{eq:divimprove}
    \pi_{new} = \mathop{\arg\max}_{\pi \in \Pi}~J_{\pi_{old}}(\pi).
\end{equation}
Then, $\pi_{new}$ satisfies  $Q^{\pi_{new}}(s_t,a_t)\geq Q^{\pi_{old}}(s_t,a_t)$, $\forall (s_t,a_t) \in \mathcal{S}\times\mathcal{A}$. Proof is given in Lemma \ref{lem:improve} in Appendix \ref{sec:proofs}.

Then, in a similar way to the proof of the convergence of the soft policy iteration \citep{haarnoja2018soft}, we can prove the convergence of the diverse policy iteration, stated in the following theorem.

\begin{theorem}[Diverse Policy Iteration]\label{thm1}
	By repeating iteration of the diverse policy evaluation applying the Bellman operator \eqref{eq:bellman}  and the diverse policy improvement \eqref{eq:divimprove}, any initial policy  converges to the optimal policy $\pi^*$ s.t. $Q^{\pi^*}(s_t,a_t)\geq Q^{\pi'}(s_t,a_t)$, $\forall~\pi'\in\Pi$, $\forall~ (s_t,a_t)\in\mathcal{S}\times\mathcal{A}$. Furthermore, such $\pi^*$ achieves maximum $J$ in \eqref{eq:objpi}.
\end{theorem}

\emph{Proof.} See Appendix \ref{pfthm1}.

For proof of Theorem \ref{thm1}, we need the assumption of finite MDPs as in the proof of usual policy iteration or SPI. Later, we consider function approximation for the policy and the value functions to implement the diverse policy iteration in continuous state and action spaces, based on the convergence proof in finite MDPs.

Although Theorem \ref{thm1} proves convergence of the diverse policy iteration for finite MDPs and provides a basis for implementation with function approximation for continuous MDPs, actually finding the optimal policy by using Theorem \ref{thm1} is difficult due to  the step \eqref{eq:divimprove} used in Theorem \ref{thm1}.  The reason is as follows.
In order to facilitate  proof of  monotone improvement by  the step \eqref{eq:divimprove},   we set $\pi$ in  the $R^{\pi,\alpha}$ term in \eqref{eq:pracobj}  as  the optimization argument, as seen in Appendix \ref{pfthm1}. Otherwise,  proof of monotone improvement is not tractable.  However, this setup causes a problem in practical implementation. For practical implementation with function approximation,  we will eventually use parameterized estimates for the required functions, as we do shortly. For the policy $\pi$ we will use $\pi_\theta$ with parameter $\theta$. 
Under this situation, let us consider the ratio function again. The ratio function $R^{\pi,\alpha}$ for a given $\pi$ is a mapping from $(\Sc,\Ac)$ to $[0,1)$, as seen in  \eqref{eq:Rphialpha}.   In the case that $\pi$ in  $R^{\pi,\alpha}$ is the optimization argument policy with parameter $\theta$, we need to define a mapping from $(\Sc,\Ac)$ to $[0,1)$ for each of all possible values of $\theta$. That is, the output value of $R^{\pi_\theta,\alpha}(s_t,a_t)$ depends not only on $(s_t,a_t)$ but also on $\theta$. To capture this situation, we need to set the input to the ratio function $R^{\pi_\theta,\alpha}$  as $(s_t,a_t,\theta)$. However, the dimension of the policy (neural network) parameter $\theta$ is huge and thus implementation of $R^{\pi_\theta,\alpha}(s_t,a_t)$  as a function of $(s_t,a_t,\theta)$ is not simple.  To circumvent this difficulty, we  need to modify the policy objective function so that it involves a much simpler form for the ratio function for easy implementation. For this,  instead of using $R^{\pi,\alpha}$ with $\pi$ being the optimization argument, we use the ratio function $R^{\pi_{old},\alpha}$ for the given current policy $\pi_{old}$ so that 
$\pi$ in the $R^{\pi,\alpha}$ term in the policy objective function is not the optimization argument anymore but fixed as the given current policy $\pi_{old}$. With this replacement, we manage to show the following result:

\begin{theorem}\label{thm2}
   Consider the new objective function for policy improvement $\tilde{J}_{\pi_{old}}(\pi(\cdot|s_t))$ in \eqref{eq:finalObj}, where the ratio function inside the expectation in \eqref{eq:finalObj} is the ratio function for the given current policy $\pi_{old}$. Suppose that the policy is parameterized with parameter $\theta$. Then, for parameterized policy $\pi_\theta$, the two objective functions $J_{\pi_{\theta_{old}}}(\pi_\theta(\cdot|s_t))$ in \eqref{eq:pracobj} and $\tilde{J}_{\pi_{\theta_{old}}}(\pi_\theta(\cdot|s_t))$ in  \eqref{eq:finalObj} have the same gradient direction for $\theta$ at $\theta=\theta_{old}$ for all $s_t\in\mathcal{S}$, where $\theta_{old}$ is the parameter of the given current policy $\pi_{old}$.
\end{theorem}

\emph{Proof.} See Appendix \ref{pfthm2}.

Note that maximizing the new policy objective function \eqref{eq:finalObj} is equivalent to minimizing $D_{KL}(\pi(\cdot|s_t)||\exp ( Q^{\pi_{old}}(s_t,\cdot)/\alpha + R^{\pi_{old},\alpha} (s_t,\cdot)) = -\tilde{J}_{\pi_{old}}(\pi(\cdot|s_t))/\beta$, and the improved policy obtained by maximizing \eqref{eq:finalObj} can be expressed as $\pi_{new}(\cdot|s_t)\propto \exp (Q^{\pi_{old}}(s_t,\cdot)/\alpha + R^{\pi_{old},\alpha} (s_t,\cdot))$.    
Note also that the new policy objective function $\tilde{J}_{\pi_{old}}(\pi(\cdot|s_t))$ is a simplified version in two aspects. First, we require the ratio function only for the given current policy. Second, the $\Ebb_{a_t \sim q}$ term in  \eqref{eq:pracobj} disappeared.  Note that $\pi$ in the $\log \pi$ inside the expectation of $\tilde{J}_{\pi_{old}}(\pi(\cdot|s_t))$ in \eqref{eq:finalObj} is still the optimization argument. However, this is not a problem since we  have the parameter $\theta$ for the policy in implementation and this parameter will be updated.
Now, the ratio function in the policy objective function $\tilde{J}_{\pi_{old}}(\pi(\cdot|s_t))$ in \eqref{eq:finalObj} in the diverse policy improvement step is the ratio function for the given current policy. Furthermore,  the ratio function in  \eqref{eq:Vestst} in the diverse policy evaluation step is also for the given current policy. Hence, we need to implement and track only the ratio function for the current policy. Now, by Theorems \ref{thm1} and \ref{thm2}, we can find the optimal policy maximizing  \eqref{eq:objpi} by iterating the diverse policy evaluation \eqref{eq:bellman} and the diverse policy improvement  maximizing $\tilde{J}_{\pi_{old}}(\pi(\cdot|s_t))$ in \eqref{eq:finalObj}.

The final step to complete the proposed diverse policy iteration is learning of the ratio function for the current policy. For this, we define an estimate function $R^\alpha:\mathcal{S}\times\mathcal{A}\rightarrow\mathbb{R}$  for the  true ratio function $R^{\pi,\alpha}$ of the current policy $\pi$ and adopt the learning method proposed in the works of \citet{sugiyama2012density, goodfellow2014generative}. That is, we first define the objective function for $R^\alpha$  as
\begin{align}\label{eq:ratio}
&J_{ratio}(R^\alpha(s_t,\cdot))=\alpha \mathbb{E}_{a_t\sim\pi(\cdot|s_t)}[\log R^\alpha(s_t,a_t)]\nonumber\\
&~~~~~~~\quad+ (1-\alpha) \mathbb{E}_{a_t\sim q(\cdot|s_t)}[\log (1-R^\alpha(s_t,a_t))].
\end{align}
Then, we learn $R^\alpha$ by maximizing the objective  $J_{ratio}(R^\alpha)$. Note that for given $s$, $J_{ratio}(R^\alpha(s,\cdot)) = \int_a  [c_1 \log R^\alpha(s,a) + c_2\log (1-R^\alpha(s,a))]da$,  where $c_1 =\alpha\pi$, and $c_2=(1-\alpha)q$.
The integral is maximized when the integrand for each $a$ is maximized.  The integrand $f(R^\alpha(s,a))=c_1\log R^\alpha(s,a)+c_2 \log (1-R^\alpha(s,a))$ is a concave function of $R^\alpha(s,a)$, and its maximum is achieved when $R^{\alpha}(s,a)= c_1/(c_1+c_2)=  \alpha\pi / (\alpha\pi+(1-\alpha)q)$. Hence, $J_{ratio}(R^\alpha)$ is maximized when $R^{\alpha} (s_t,a_t) = \alpha\pi / (\alpha\pi+(1-\alpha)q)$, which is exactly the desired ration function shown in   \eqref{eq:Rphialpha}. Therefore, the ratio function for the current policy can be estimated by maximizing the objective funtion $J_{ratio}(R^\alpha)$.

\subsection{Diversity Actor Critic Implementation}
\label{subsec:dac}

We use deep neural networks to implement   the policy $\pi$, the ratio function $R^\alpha$, and the diverse value functions $Q$ and $V$, and  their network parameters are  $\theta$, $\eta$, $\phi$, and $\psi$, respectively. Based on $\tilde{J}_{\pi_{old}}(\pi)$ in \eqref{eq:finalObj} and $J_{ratio}(R^\alpha)$ in \eqref{eq:ratio}, we provide the practical objective functions: $\hat{J}_{\pi}(\theta)$ for the parameterized policy $\pi_\theta$, and $\hat{J}_{R^\alpha}(\eta)$ for the parameterized ratio function estimator $R^\alpha_\eta$, given by
\begin{align}
\hat{J}_{\pi}(\theta)&=\mathbb{E}_{s_t\sim\mathcal{D},~a_t\sim\pi_\theta}[Q_\phi(s_t,a_t)+\alpha\log R_\eta^\alpha (s_t,a_t)\nonumber\\
&~~- \alpha\log\pi_\theta(a_t|s_t)],\label{eq:Jpihat}\\
\hat{J}_{R^\alpha}(\eta)&=\mathbb{E}_{s_t\sim\mathcal{D}}[\alpha\mathbb{E}_{a_t\sim\pi_\theta}[\log R_\eta^\alpha(s_t,a_t)]\nonumber\\
&~~+ (1-\alpha)\mathbb{E}_{a_t\sim \mathcal{D}}[\log(1-R_\eta^\alpha(s_t,a_t))]], \label{eq:Jrhat}
\end{align}where $\Dc$ denotes the replay buffer. Based on the Bellman operator $\mathcal{T}^\pi$ in \eqref{eq:bellman}, we  provide the loss functions: $\hat{L}_Q(\phi)$ and $\hat{L}_V(\psi)$ for the parameterized value functions $Q_\phi$ and $V_\psi$,  respectively, given by
\begin{align}
\label{eq:q}
\hat{L}_Q(\phi)&=\mathbb{E}_{(s_t,~a_t)\sim\mathcal{D}}\left[\frac{1}{2}(Q_\phi(s_t,a_t) - \hat{Q}(s_t,a_t))^2\right],\\
\label{eq:v}
\hat{L}_V(\psi)&=\mathbb{E}_{s_t\sim\mathcal{D}}\left[\frac{1}{2}(V_\psi(s_t)-\hat{V}(s_t))^2\right],
\end{align}where the target values $\hat{Q}$ and $\hat{V}$ are defined as
\begin{align}
&\hat{Q}(s_t,a_t) = \frac{1}{\beta}r_t + \gamma \mathbb{E}_{s_{t+1}\sim P}[V_{\bar{\psi}}(s_{t+1})]\label{eq:tarq}\\
&\hat{V}(s_t)=\mathbb{E}_{a_t\sim\pi_\theta}[Q_\phi(s_t,a_t)\nonumber\\
&\hspace{6.5em}+\alpha\log R_\eta^\alpha (s_t,a_t) -\alpha\log \alpha\pi_\theta(a_t|s_t)] \nonumber\\
&\quad+ (1-\alpha)\mathbb{E}_{a_t\sim \mathcal{D}}[\log R_\eta^\alpha(s_t,a_t)-\log \alpha \pi_\theta(a_t|s_t)]\label{eq:tarv}.
\end{align}Here, $\bar{\psi}$ is the network parameter of the target value $V_{\bar{\psi}}$ updated by exponential moving average (EMA) of $\psi$ for stable learning \citep{mnih2015human}. In addition, we use two $Q$-functions $Q_{\phi_i},~i=1,2$ to reduce overestimation bias as proposed in \citep{fujimoto2018addressing} and applied in SAC, and each Q-function is updated to minimize their loss function $\hat{L}_Q(\phi_i)$. For the policy and the value function update, the minimum of two $Q$-functions is used for the policy update.   Combining all up to now, we propose the diversity actor-critic (DAC) algorithm summarized as Algorithm \ref{alg:dac}. Here, note that DAC becomes SAC when $\alpha=1$, and becomes standard off-policy RL without entropy regularization when $\alpha=0$. When $0 < \alpha <1$, we accomplish sample-aware entropy regularization. Detailed implementation of DAC is provided in Appendix \ref{sec:imple}. For DAC, we can  consider the technique of SAC proposed in \citep{haarnoja2018soft2} using $Q$-function only for reducing complexity  or automatic tuning of $\beta$ for balancing the entropy and the return. However, in the case of DAC, both $\alpha$ and $\beta$ affect the entropy term, so both should be tuned simultaneously.

\begin{algorithm}[!t]
	\begin{algorithmic}
		\STATE{Initialize parameter $\theta$, $\eta$, $\psi$, $\bar{\psi}$, $\xi$, $\phi_i,~i=1,2$}
		\FOR{each iteration}
		\STATE{Sample a trajectory $\tau$ of length $N$ by using $\pi_{\theta}$}
		\STATE{Store the trajectory $\tau$ in the buffer $\mathcal{D}$}
		\FOR{each gradient step}
		\STATE{Sample random minibatch of size $M$ from $\mathcal{D}$}
		\STATE{Compute $\hat{J}_{\pi}(\theta)$, $\hat{J}_{R^\alpha}(\eta)$, $\hat{L}_Q(\phi_i)$, $\hat{L}_V(\psi)$ from the minibatch}
		\STATE{$\theta\leftarrow \theta+\delta\nabla_{\theta}\hat{J}_\pi(\theta)$}
		\STATE{$\eta\leftarrow \eta+\delta\nabla_{\eta}\hat{J}_{R^\alpha}(\eta)$}
		\STATE{$\phi_i\leftarrow \phi_i-\delta\nabla_{\phi_i}\hat{L}_Q(\phi_i),~i=1,2$}
		\STATE{$\psi\leftarrow \psi-\delta\nabla_{\psi}\hat{L}_V(\psi)$}
		\STATE{Update $\bar{\psi}$ by EMA from $\psi$}
		\IF{$\alpha$-adpatation is applied}
		\STATE{Compute $\hat{L}_\alpha(\xi)$ from the minibatch}
		\STATE{$\xi\leftarrow \xi-\delta\nabla_{\xi}\hat{L}_\alpha(\xi)$}
		\ENDIF
		\ENDFOR
		\ENDFOR
	\end{algorithmic}
	\caption{Diversity Actor Critic}
	\label{alg:dac}
\end{algorithm}

\section{$\alpha$-Adaptation}
\label{sec:autoalpha}

In the proposed sample-aware entropy regularization, the weighting factor $\alpha$ between the policy and the sample distribution plays an important role in controlling the ratio between the policy distribution $\pi$ and the sample action distribution $q$.
However, it is difficult to find optimal $\alpha$  for each environment. To circumvent this $\alpha$ search,   we  propose an automatic adaptation method for  $\alpha$ based on  max-min principle widely considered in game theory, robust learning, and decision making problems \citep{chinchuluun2008pareto}.
That is, since we do not know optimal $\alpha$, an alternative formulation is that we maximize the return while maximizing the worst-case  sample-aware entropy, i.e., $\min_{\alpha} \mathcal{H}(q_{mix}^{\pi,\alpha})$. Then, the max-min approach can be formulated as follows:
\begin{equation} \label{eq:alphamaxmin}
\small
\max_\pi
\mathbb{E}_{\tau_0\sim\pi}\left[\sum_{t}\gamma^t (r_t + \beta \min_\alpha[\mathcal{H}(q_{mix}^{\pi,\alpha})-\alpha c])\right]
\end{equation}
where $c$ is a control hyperparameter  for $\alpha$ adaptation. Note that we learn $\alpha$ to minimize the sample-aware entropy so that the entropy is maintained above a certain level to explore the state and action spaces well. So, the $\alpha$ learning objective is given by a Lagrangian form. Thus, when the $\alpha$-learning model is parameterized with parameter $\xi$, the  $\alpha$ learning objective is given by 
$\hat{L}_{\alpha}(\xi)=\mathbb{E}_{s_t\sim\mathcal{D}}[\mathcal{H}(q_{mix}^{\pi_\theta,\alpha_\xi}) - \alpha_\xi c]$.
Detailed implementation of $\alpha$-adaptation is given in Appendix \ref{subsec:gradalpha}.

\section{Experiments}
\label{sec:experiments}

\begin{figure}[!t]
\subfigure[Continuous 4-room maze]{\includegraphics[width=0.23\textwidth]{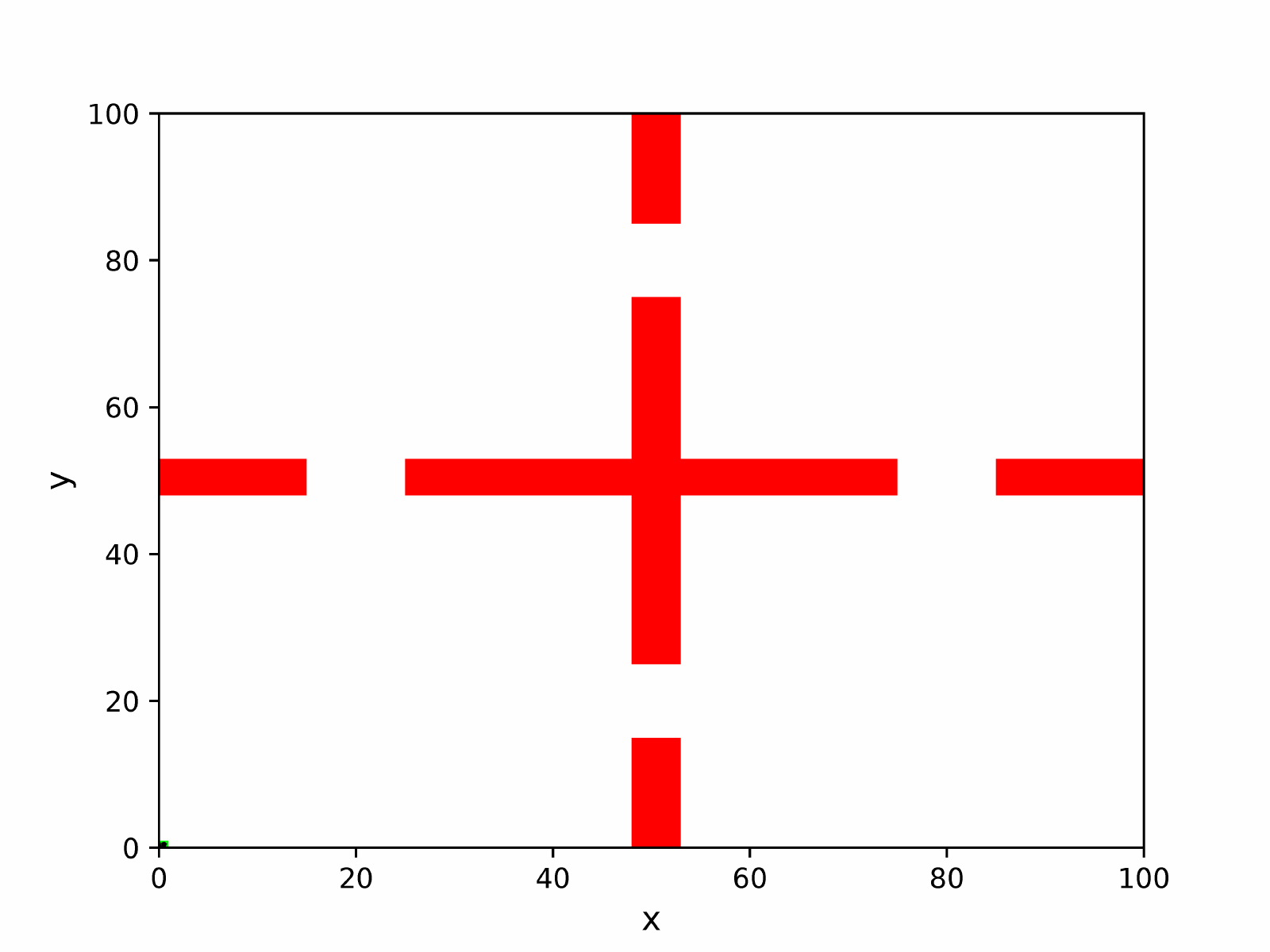}\label{fig:contmaze}}
\vspace{-1em}
\subfigure[Number of state visitation]{\includegraphics[width=0.23\textwidth]{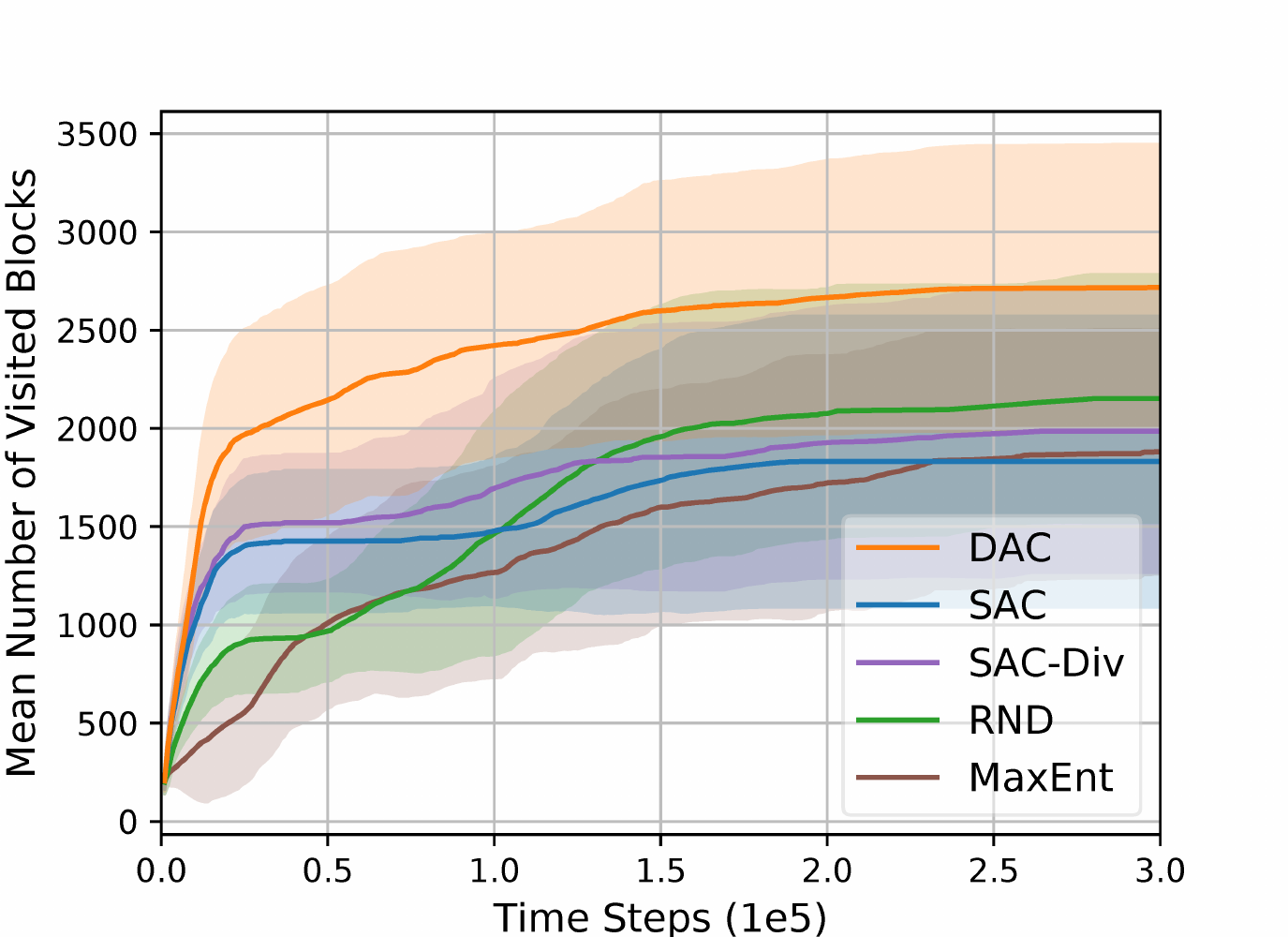}\label{fig:uniquevisit}}
\subfigure[State visit histogram at 5k (left) 50k (middle) 300k (right)  steps]{\includegraphics[width=0.48\textwidth]{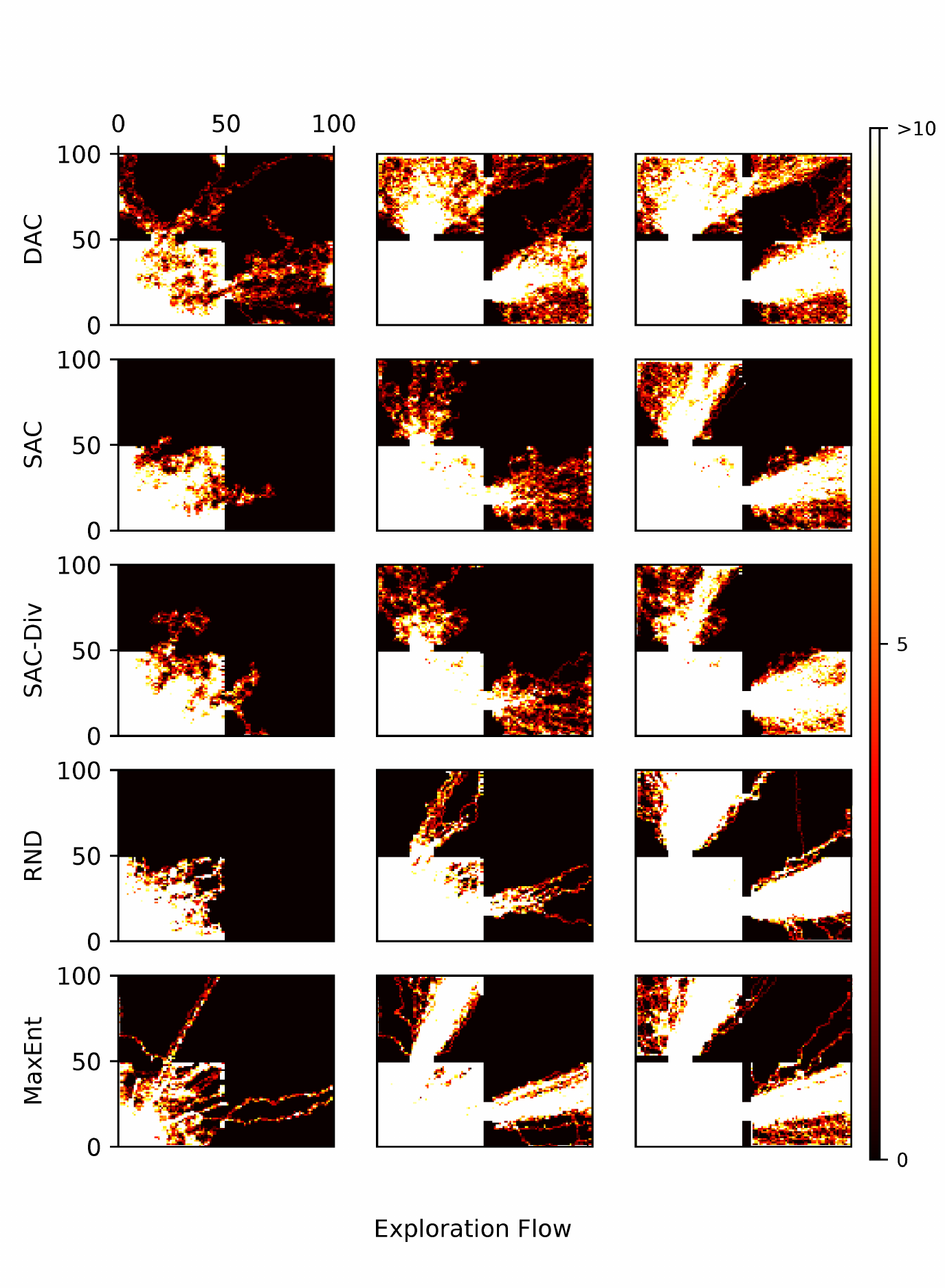}\label{fig:visitstate}}
\caption{Pure exploration task: Continuous 4-room maze}
\vspace{-1.5em}
\end{figure}

\begin{figure*}[!h]
	\centering
	\subfigure[SparseHalfCheetah-v1]{\includegraphics[width=0.24\textwidth]{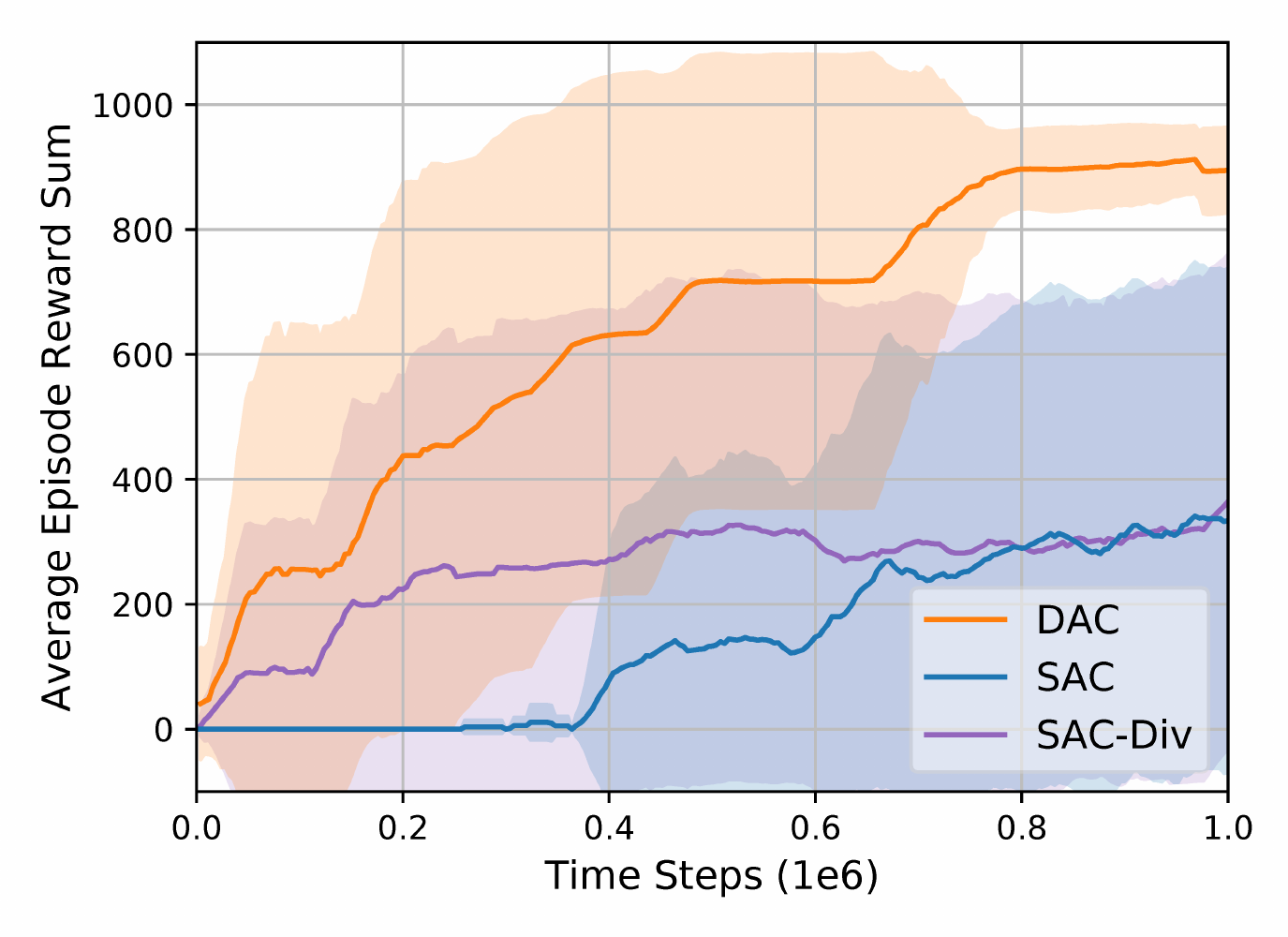}}
	\subfigure[SparseHopper-v1]{\includegraphics[width=0.24\textwidth]{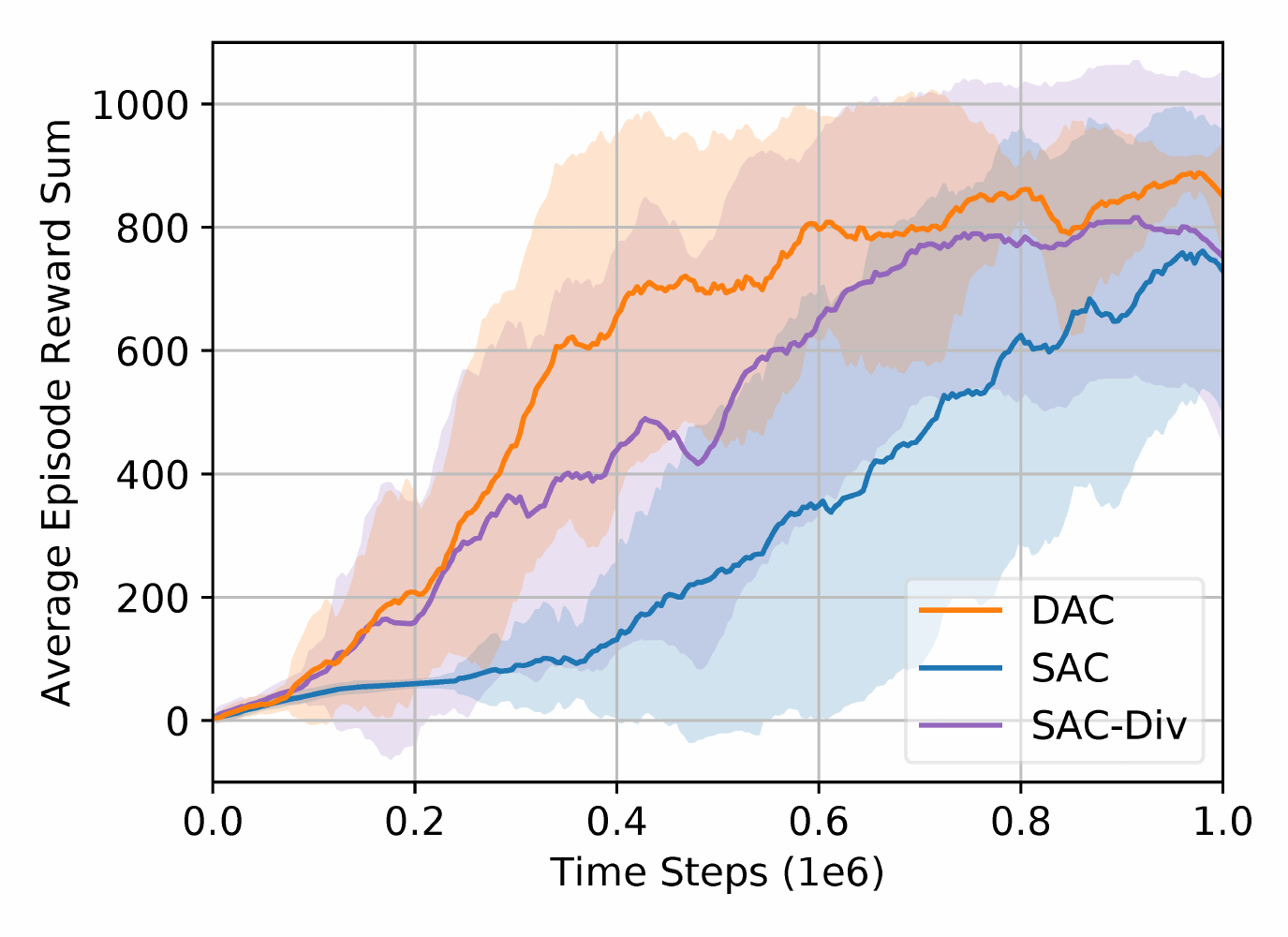}}
	\subfigure[SparseWalker2d-v1]{\includegraphics[width=0.24\textwidth]{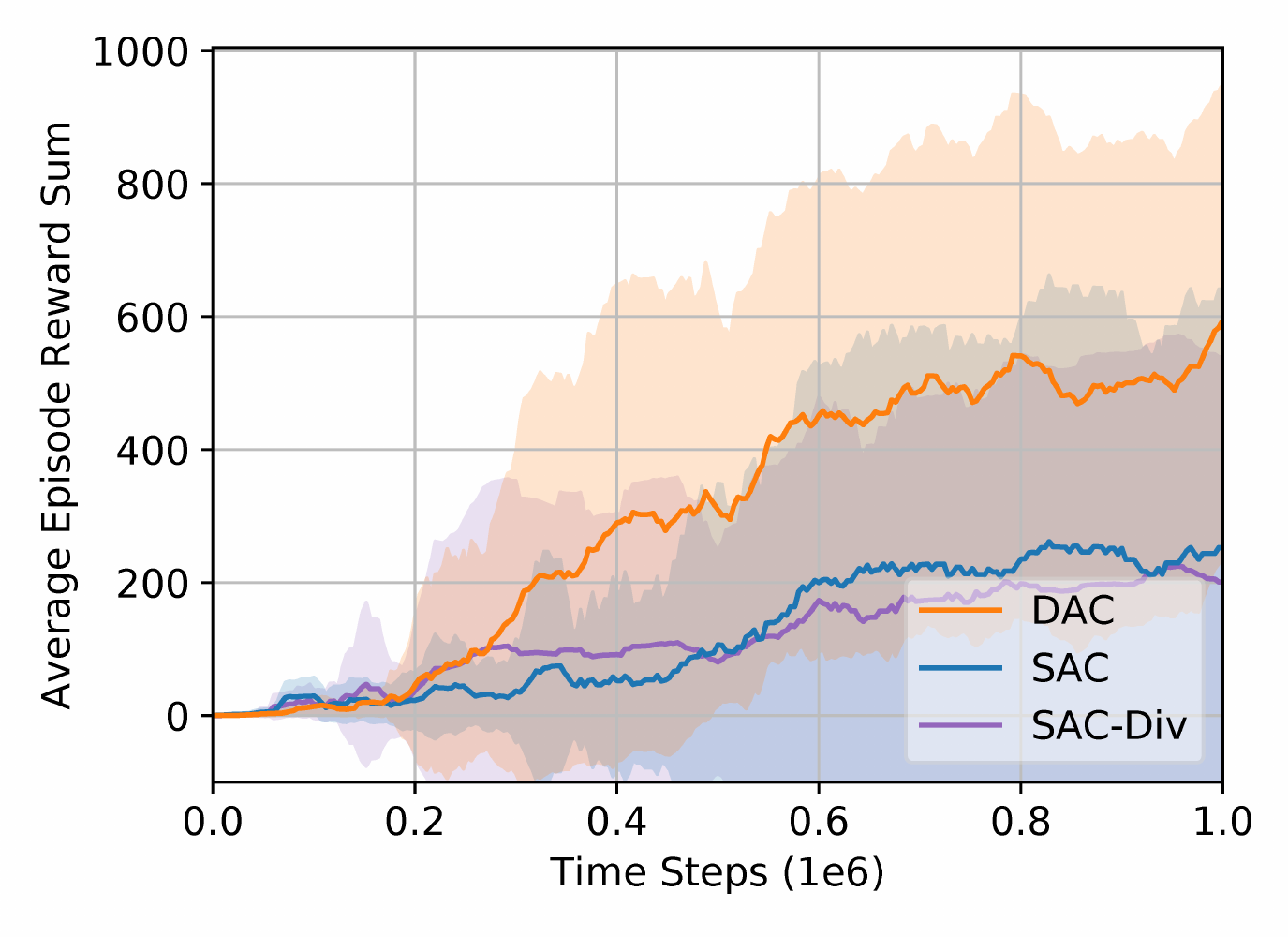}\label{fig:spswalker}}
	\subfigure[SparseAnt-v1]{\includegraphics[width=0.24\textwidth]{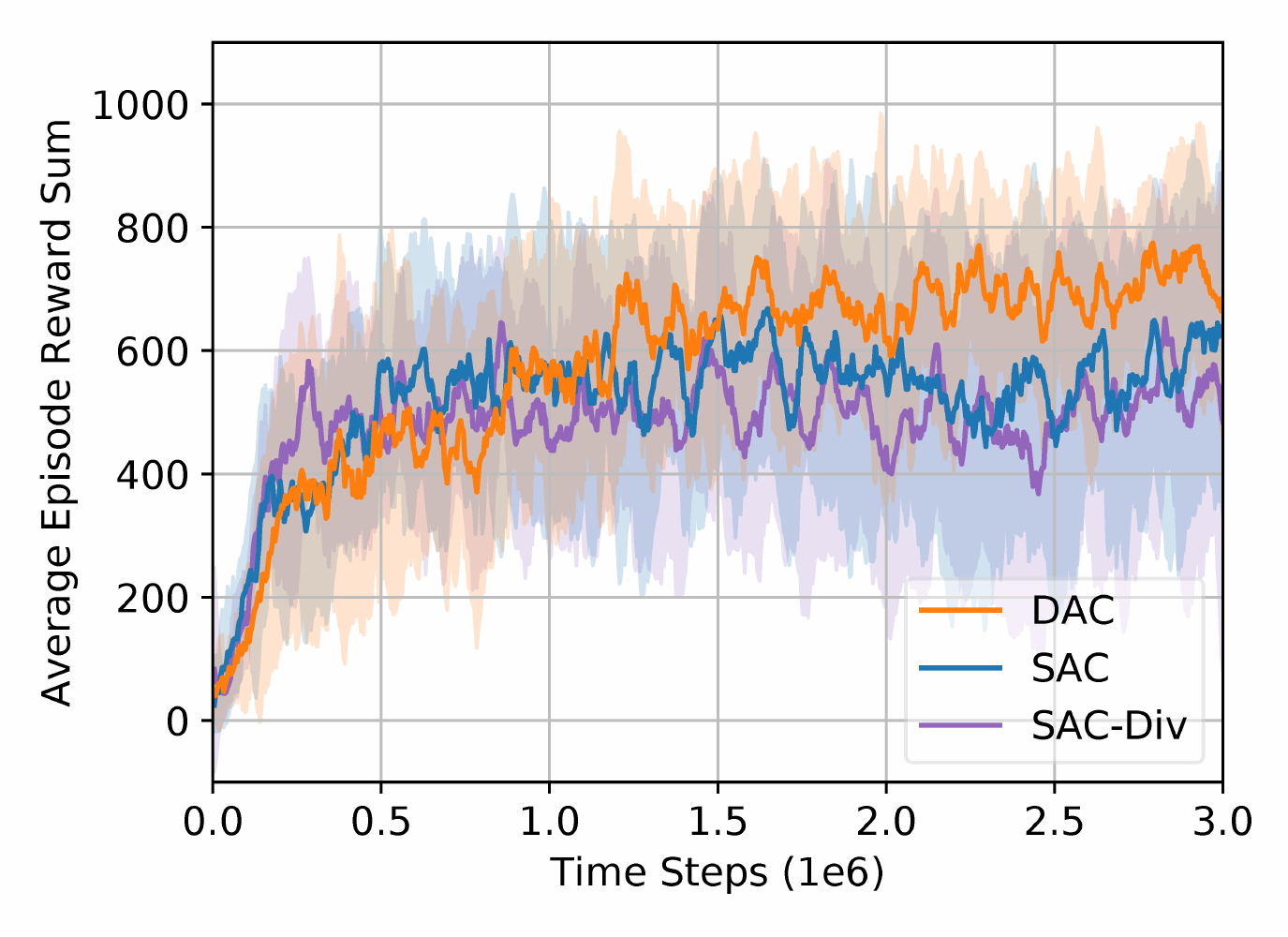}}
	\caption{Performance comparison on Sparse Mujoco tasks}
	\label{fig:compsparsemujoco}
\end{figure*}

In this section, we evaluate the proposed DAC algorithm on various continuous-action control tasks and provide ablation study. We first consider the pure exploration performance and then the performance  on challenging sparse-reward or delayed Mujoco tasks. The source code of DAC based on Python Tensorflow is available at \url{http://github.com/seungyulhan/dac/}.

\subsection{Pure Exploration Performance}
\label{subsec:pureexp}

For comparison baselines, we  first considered the state-of-the-art  entropy regularization methods: SAC and SAC-Div. 
SAC-Div is SAC combined with the exploration method in \citep{hong2018diversity} that diversifies the policy from the buffer distribution by simply maximizing $J_{SAC}(\pi) + \alpha_d D(\pi||q)$ for some divergence $D$, where $J_{SAC}(\pi)$ is given in \eqref{eq:sacobj}. For SAC-Div, we  considered the KL divergence and the adaptive scale $\alpha_d$ with $\delta_d = 0.2$, as suggested in \citep{hong2018diversity}.  The case of JS divergence used for SAC-Div is provided in Appendix \ref{sec:mainablation}.  Note that both  SAC-Div and DAC contain a divergence term in their objective functions (DAC  contains $D_{JS}^\alpha$, as seen in \eqref{eq:HmixDecomp} and \eqref{eq:objpi}), but there is an important difference.  SAC-Div adds a single divergence term on the reward sum $J_{SAC}(\pi)$. So, SAC-Div keeps $\pi$ away from $q$, but does not guide learning of the policy to visit states on which the divergence between $\pi$ and $q$ is large. On the contrary, DAC contains the divergence term $D_{JS}^{\alpha}$ as an intrinsic reward at each time step inside the reward sum of $J(\pi)$, as seen in \eqref{eq:objpi} and \eqref{eq:HmixDecomp}. Hence, DAC not only keeps $\pi$ away from $q$ but also  learns a policy to  visit  states on which the divergence between $\pi$ and $q$ is large,  to have  large $J(\pi)$, so that more actions different from $q$ are possible. This situation is analogous to the situation of SAC in which the entropy is included as an intrinsic reward inside the sum of $J_{SAC}(\pi)$, as seen in \eqref{eq:sacobj}, and hence for large $J_{SAC}(\pi)$, SAC learns a policy to visit states on which the policy action entropy is large. In addition to SAC and SAC-Div, we considered the recent high-performance state-based exploration methods: random network distillation (RND) \citep{burda2018exploration} and MaxEnt(State) \citep{hazan2019provably}.   RND explores rare states by adding an intrinsic reward based on model prediction error, and MaxEnt(State) explores rare states by using a reward functional based on the entropy of state mixture distribution. Detailed simulation setup is provided in Appendix \ref{sec:simsetup}.

In order to see the pure exploration performance of DAC ($\alpha=0.5$ is used), we considered state visitation on a $100\times100$ continuous 4-room maze. The maze environment was designed by modifying a continuous grid map available at \url{https://github.com/huyaoyu/GridMap}, and it is shown in Fig. \ref{fig:contmaze}. State is the $(x,y)$ position of the agent in the maze,  action is $(dx,dy)$ bounded by $[-1,1]\times [-1,1]$, and the agent location after action becomes $(x+dx,y+dy)$. The agent starts from the left lower corner $(0.5,0.5)$ and explores the maze without any reward. Fig. \ref{fig:uniquevisit} shows the  number of  total different state visitations  averaged over $30$ seeds, where the number of state visitations is obtained based on quantized 1 $\times$ 1 squares. Here, the shaded region in the figure represents one standard deviation (1$\sigma$) from the mean. As seen in Fig. \ref{fig:uniquevisit}, DAC visited much more states than the other methods, which shows  the superior exploration performance of DAC.  Fig. \ref{fig:visitstate} shows the corresponding state visit histogram of all seeds with 1 $\times$ 1 square quantization. Here, as the color of a state becomes brighter, the state is visited more times.  It is seen that SAC/SAC-Div rarely visit the  right upper room even at 300k time steps, RND and MaxEnt(State) visit the right upper room more than SAC/SAC-Div, and DAC visits the right upper room far earlier and  more than the other methods. 
\begin{figure}[!b]
	\centering
	\subfigure[Divergence $D_{JS}^\alpha$]{\includegraphics[width=0.23\textwidth]{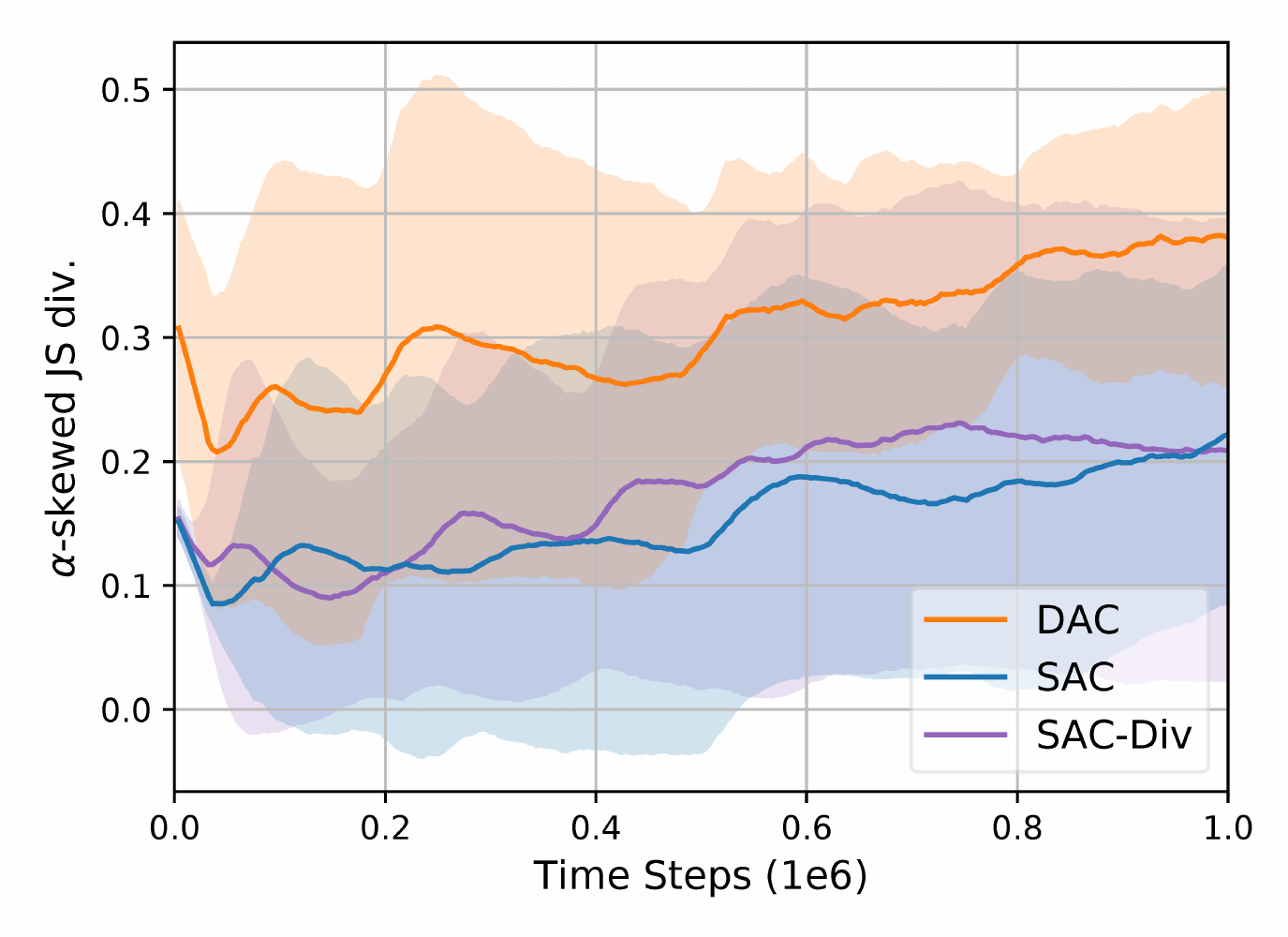}
	\label{fig:compdivsparsehalf}}
	\subfigure[Number of state visitation]{\includegraphics[width=0.23\textwidth]{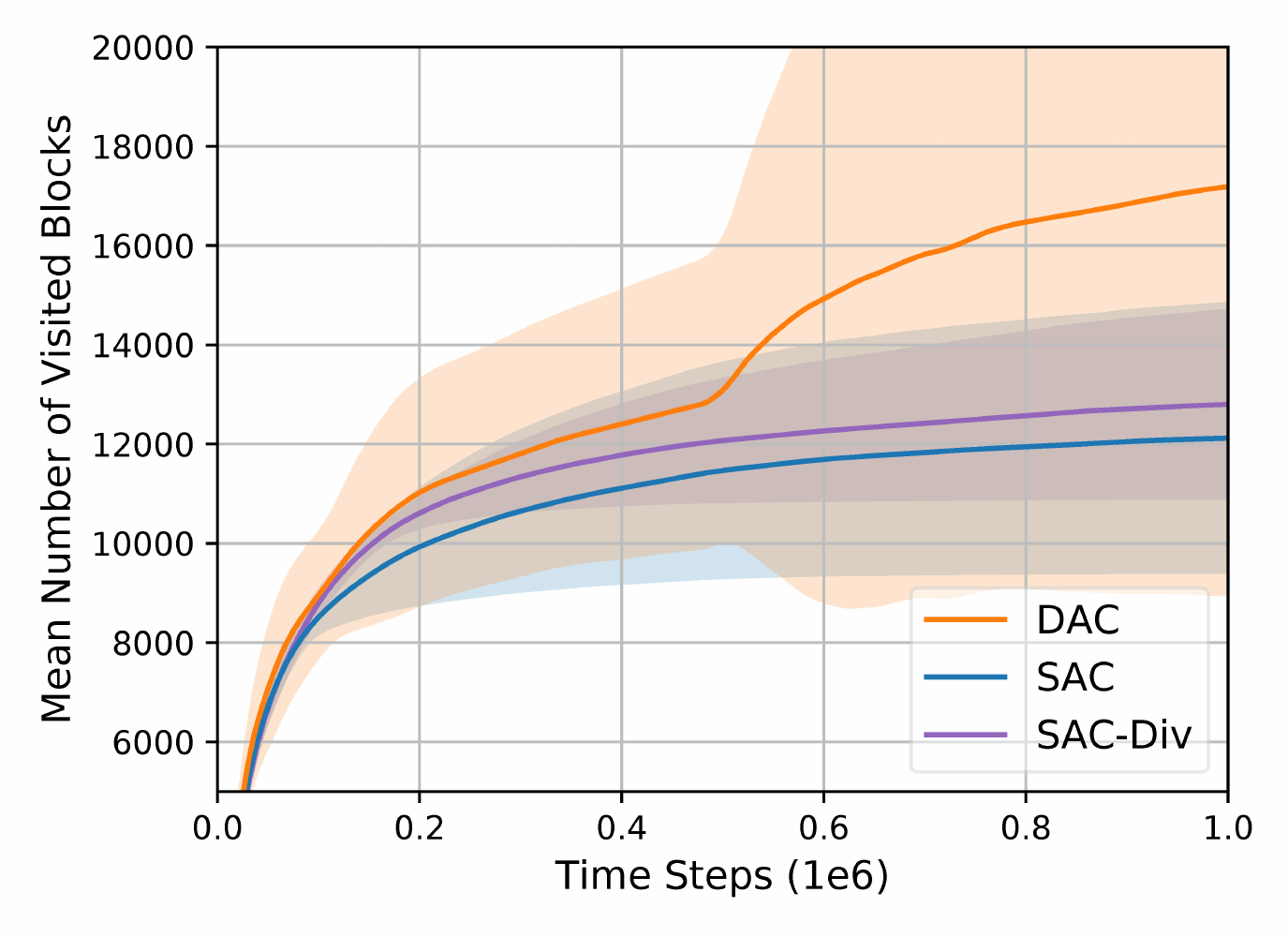}
	\label{fig:compcovsparsehalf}}
	\caption{(a) $\alpha$-skewed JS symmetrization of KLD $D_{JS}^\alpha(\pi||q)$ with $\alpha=0.5$  and (b) the corresponding mean number of state visitation}
	\label{fig:compdivcovsparsehalf}
	\vspace{-0.8em}
\end{figure}

\begin{figure*}[!h]
	\centering
	\subfigure[HumanoidStandup-v1]{\includegraphics[width=0.19\textwidth]{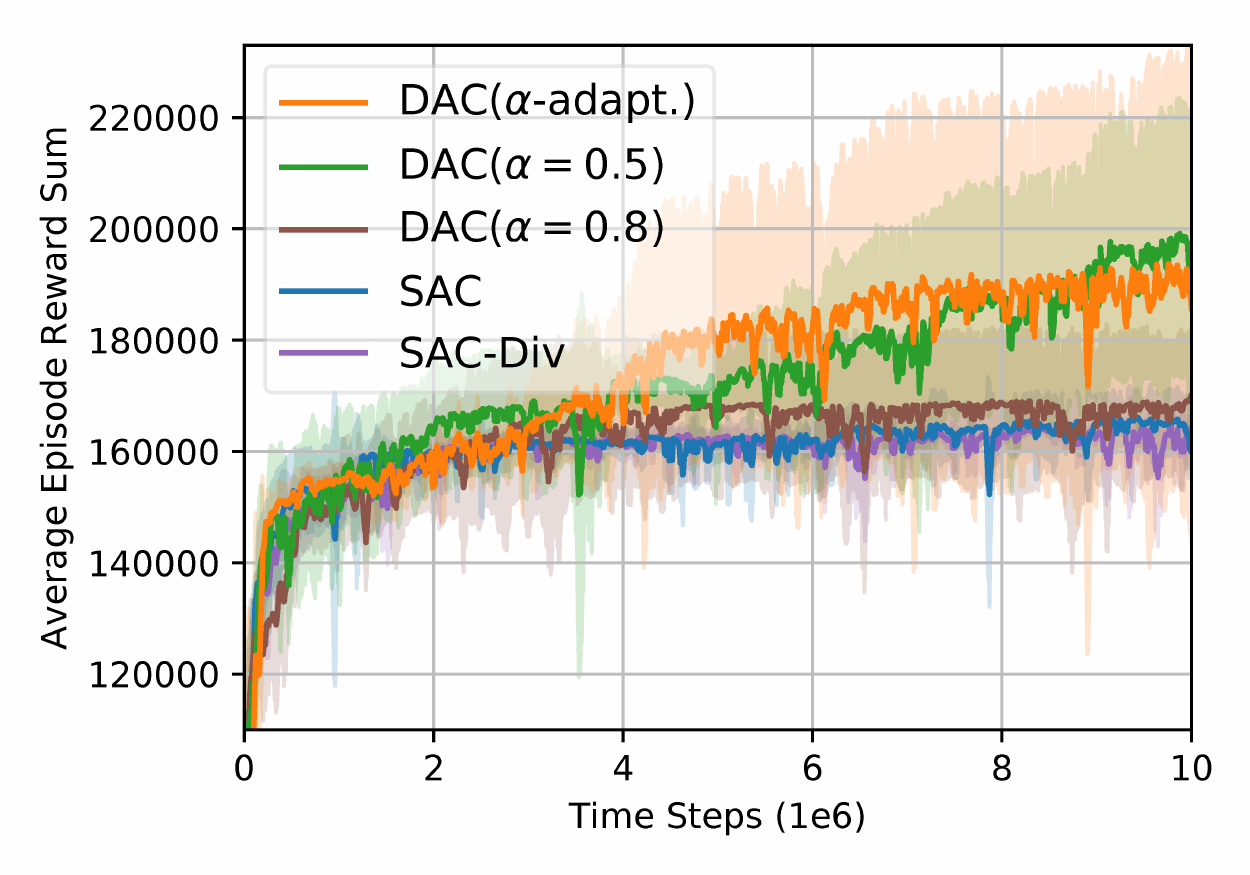}}
	\subfigure[Del.HalfCheetah-v1]{\includegraphics[width=0.19\textwidth]{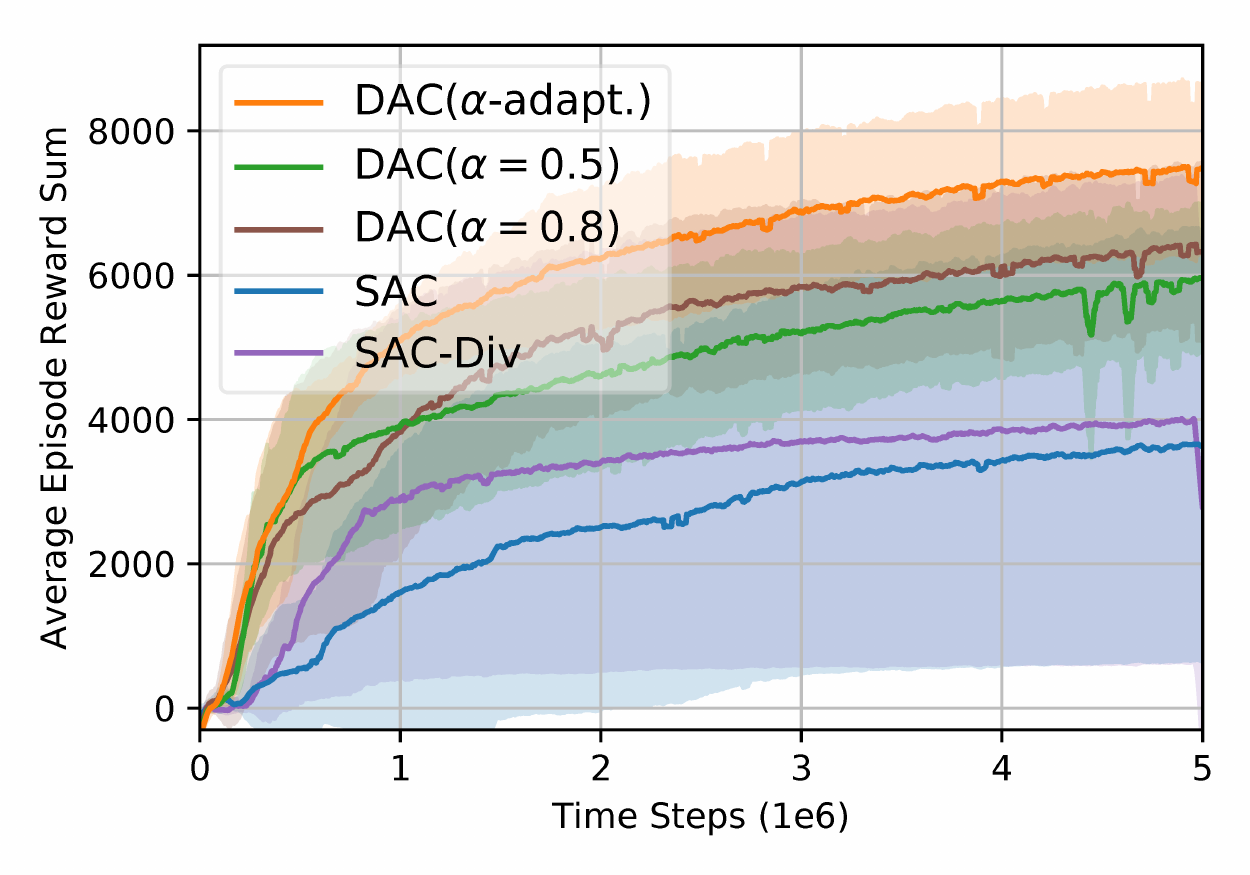}}
	\subfigure[Del.Hopper-v1]{\includegraphics[width=0.19\textwidth]{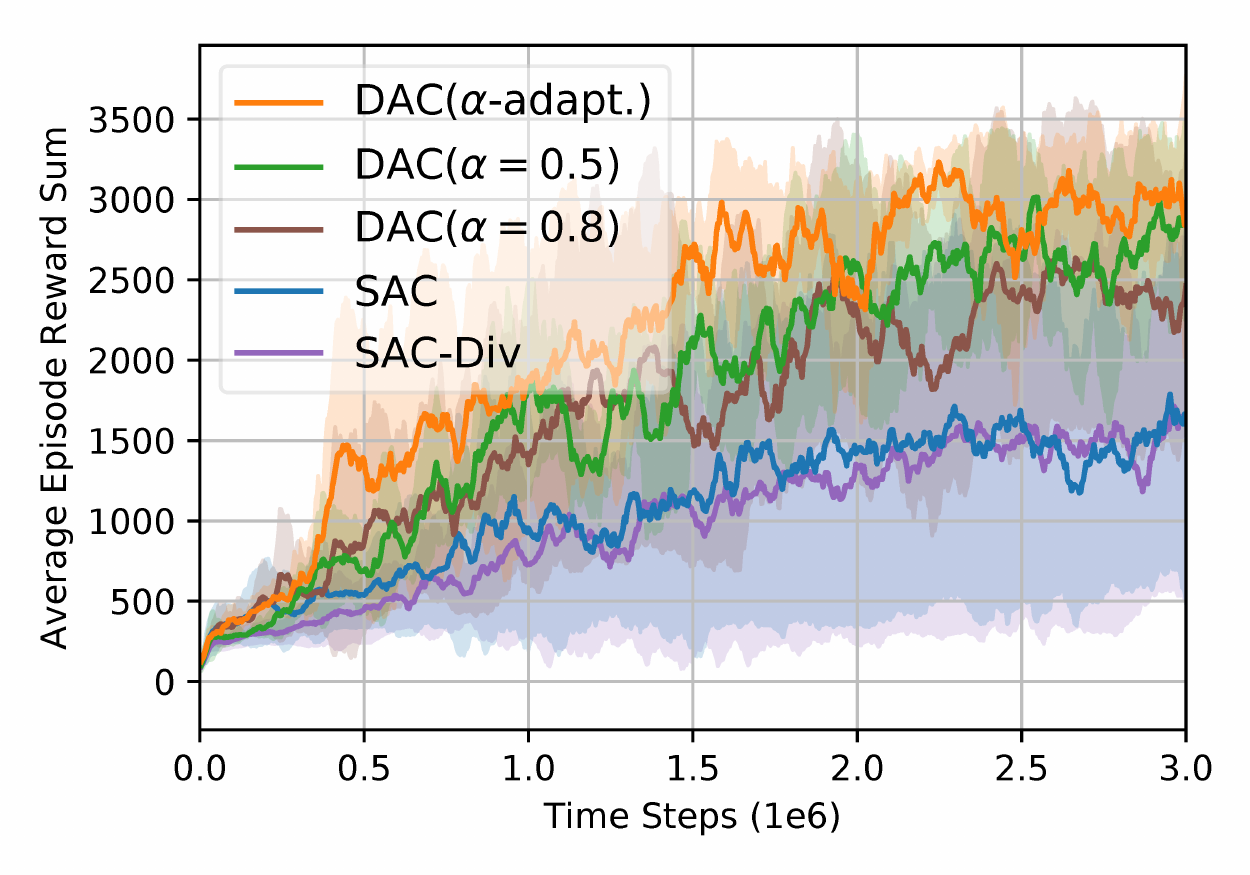}}
	\subfigure[Del.Walker2d-v1]{\includegraphics[width=0.19\textwidth]{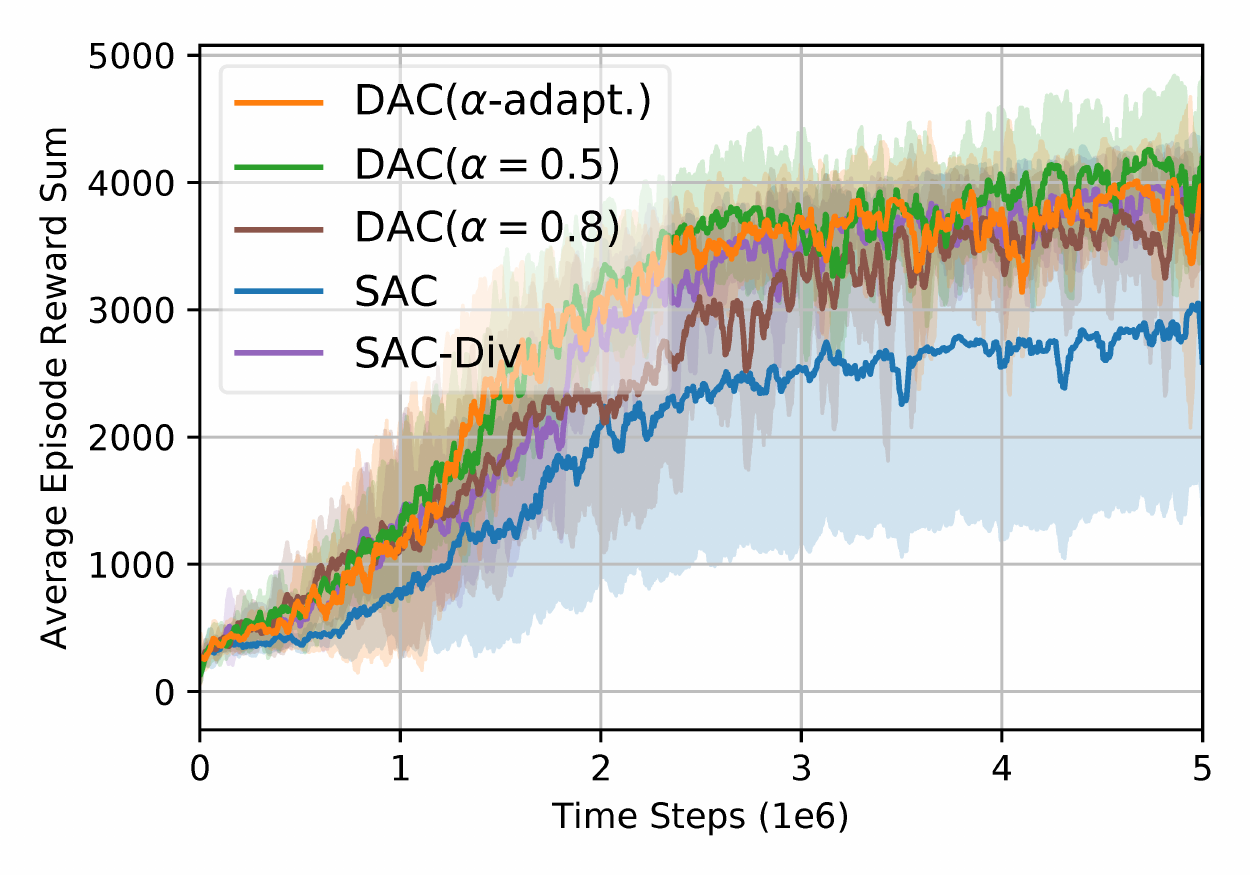}}
	\subfigure[Del.Ant-v1]{\includegraphics[width=0.19\textwidth]{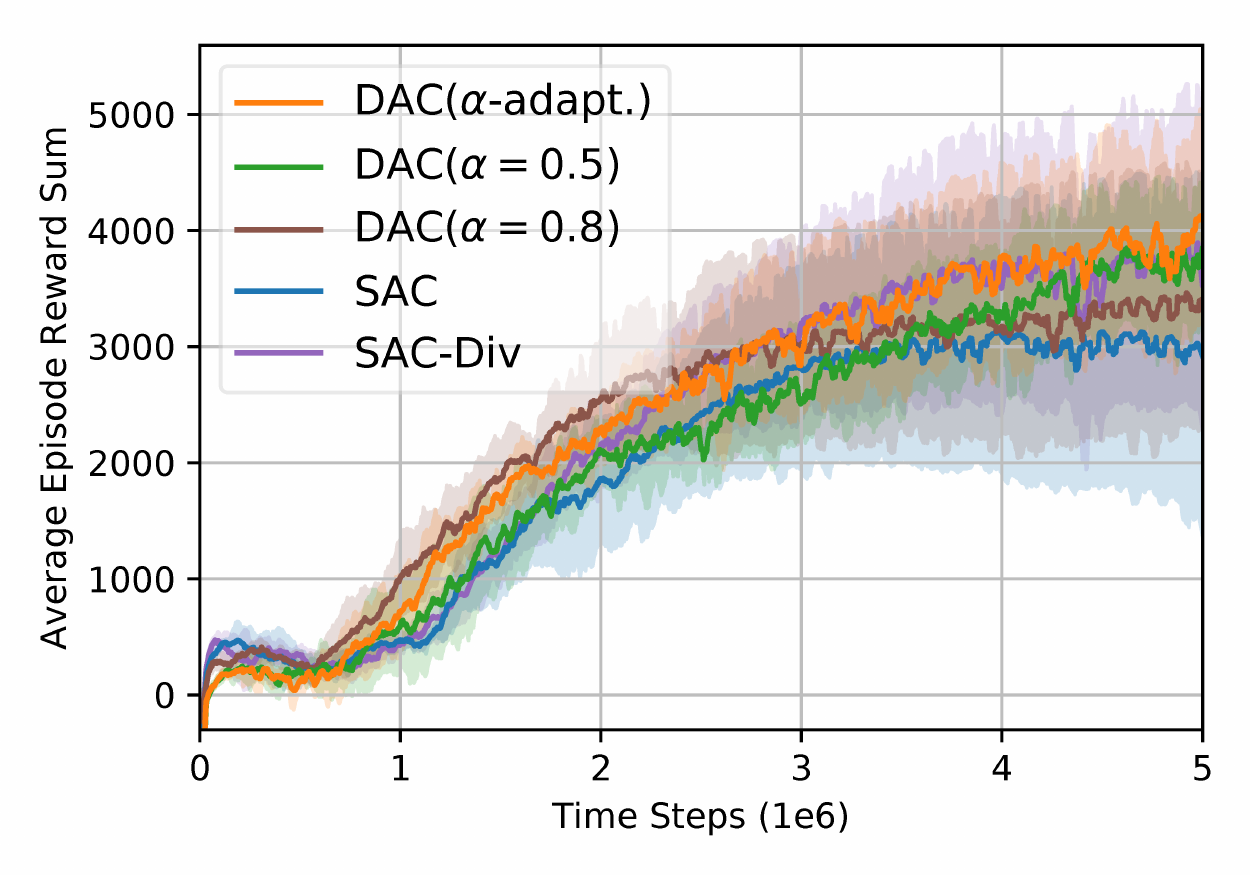}}
	\vspace{-.2em}
	\caption{Performance comparison on HumanoidStandup and Delayed Mujoco tasks (A zoomed version of the figure is available at Figure	\ref{fig:compmujoco2} in Appendix \ref{appendSection:PerComp}.)}
	\label{fig:compmujoco}
	\vspace{-0.8em}
\end{figure*}

\subsection{Performance on Sparse-Reward Mujoco Tasks}
\label{subsec:performSAC}

Then, we  evaluated the performance on  sparse-reward Mujoco tasks, which have been widely used as difficult environments for RL in many previous studies \citep{hong2018diversity, mazoure2019leveraging, burda2018exploration}.
We considered two versions. One was SparseMujoco, which is a sparse version of Mujoco \citep{todorov2012mujoco} in OpenAI Gym \citep{brockman2016openai}, and the reward is 1 if the agent satisfies a certain condition, otherwise 0 \citep{hong2018diversity,mazoure2019leveraging}.  The other was DelayedMujoco \citep{zheng2018learning, guo2018generative}, which  has the same state-action spaces as original Mujoco tasks but reward is sparsified. That is, rewards for $D$ time steps are accumulated and the accumulated reward sum is delivered to the agent once every $D$ time steps, so the agent receives no reward during the accumulation time.

First, we fixed  $\alpha=0.5$ for DAC and tested DAC on SparseMujoco. The result is shown in Fig. \ref{fig:compsparsemujoco}, which  shows the  performance  averaged over $10$ random seeds.  For all performance plots, we used deterministic evaluation which generated an episode by deterministic policy for each iteration, and the shaded region in the figure represents one standard deviation (1$\sigma$) from the mean.  
It is seen that DAC has significant performance gain over the competitive SAC and SAC-Div baselines. Figs. \ref{fig:compdivcovsparsehalf} (a) and (b)
show 
the divergence $D_{JS}^\alpha(\pi||q)$ curve 
and the corresponding number of discretized state visitation curve, respectively, on SparseHalfCheetah shown in Fig.	\ref{fig:compsparsemujoco}(a).  (The curves for the other tasks are provided  in Appendix \ref{subsec:appbaseline}. See Appendix \ref{subsec:appbaseline} about the discretization.) It is seen in Fig. \ref{fig:compdivsparsehalf} that  the divergence of DAC is much higher than those of SAC/SAC-Div throughout the learning time.
This implies that the policy of DAC choose more diverse actions from the policy distribution far away from the sample action distribution $q$, so DAC visits more diverse states than the baselines, as seen in Fig. \ref{fig:compcovsparsehalf}.

Next, we tested DAC on DelayedMujoco and HumanoidStandup. Note that HumanoidStandup is one of the difficult high-action dimensional Mujoco tasks, so its reward is not sparsified for test.  We considered three cases for $\alpha$ of DAC: $\alpha=0.5$, $0.8$, and $\alpha$-adaptation. Fig. \ref{fig:compmujoco} shows the result. Again, we can observe significant performance improvement by DAC over the SAC baselines. We can also  observe that the best $\alpha$ depends on tasks. For example, $\alpha=0.8$ is the best for DelayedHalfCheetah, but $\alpha=0.5$ is the best for DelayedAnt. Thus, the result shows that $\alpha$-adaptation method is necessary  in order to adapt $\alpha$ properly for each task. Although the proposed $\alpha$-adaptation in Section \ref{sec:autoalpha} is sub-optimal, DAC with our $\alpha$-adaptation method has  top-level performance across all the considered tasks and further enhances the performance in some cases such as DelayedHalfCheetah and DelayedHopper tasks.

\begin{figure}[!t]
	\subfigure[DelayedHalfCheetah]{\includegraphics[width=0.23\textwidth]{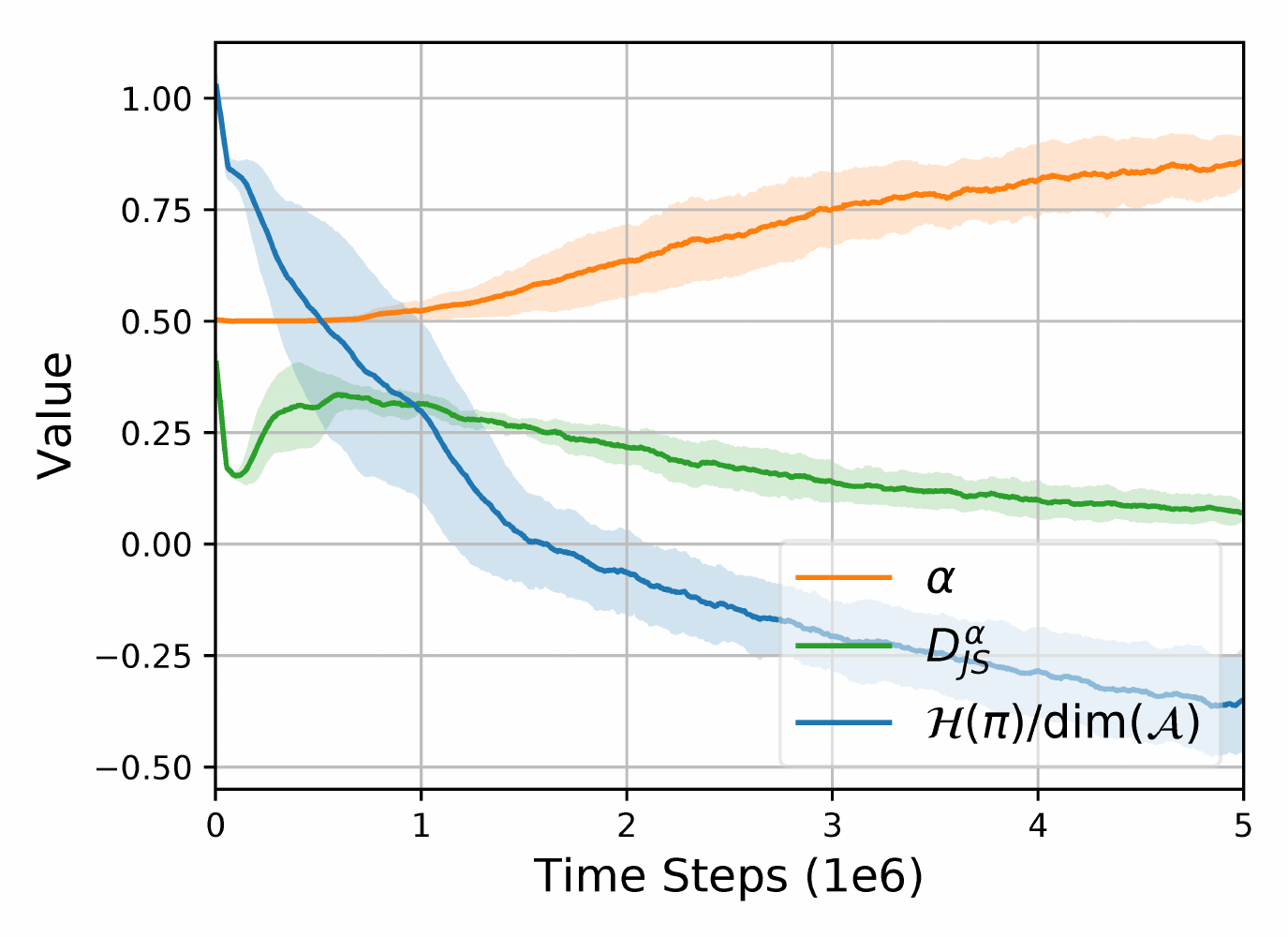}\label{fig:alphadhalf}}
 	\subfigure[DelayedHopper-v1]{\includegraphics[width=0.23\textwidth]{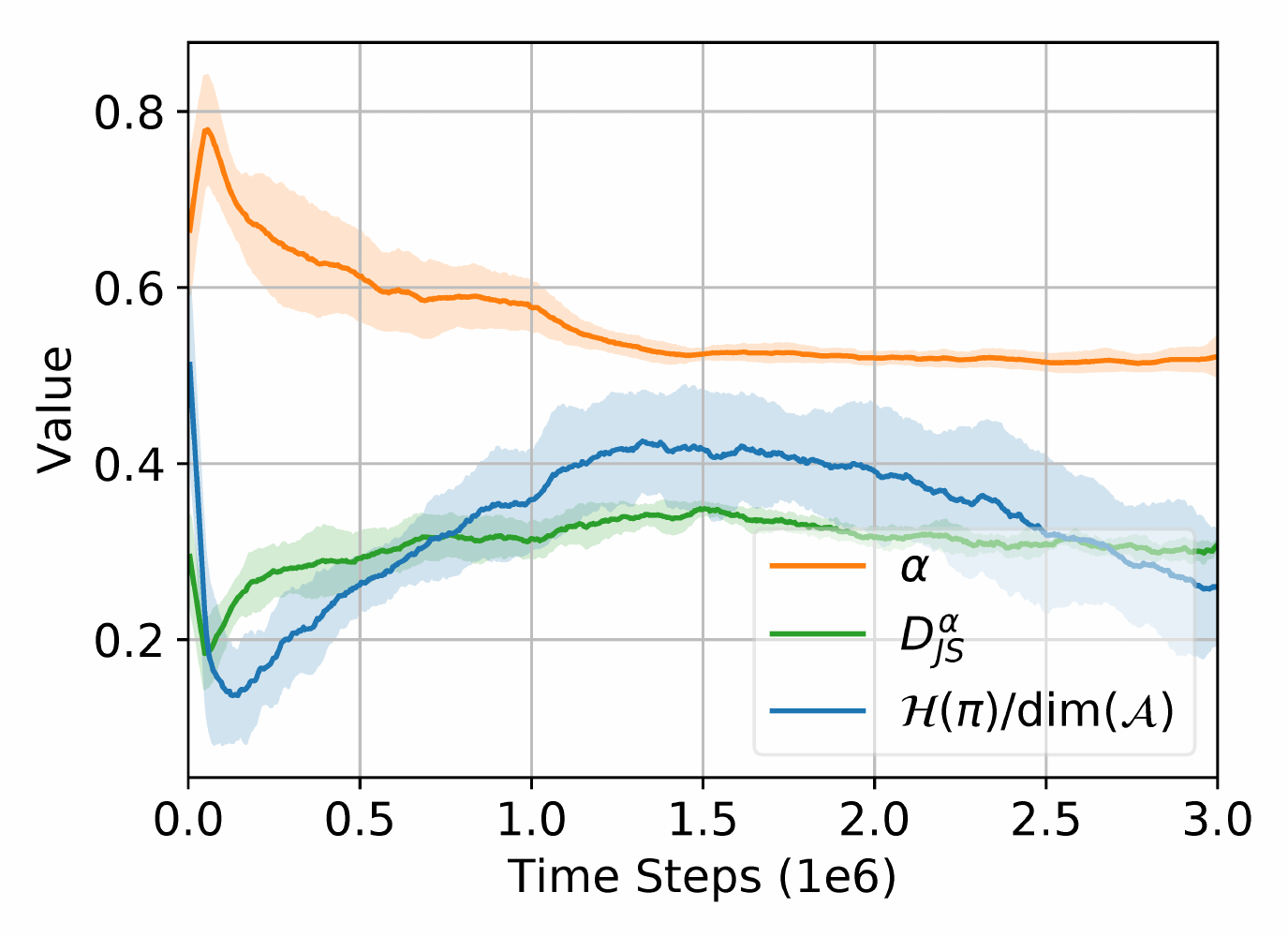}\label{fig:alphadhop}}
	\caption{Averaged learning curve for $\alpha$-adaptation}
	\label{fig:adapt}
\end{figure}

We studied more on the $\alpha$-adaptation proposed in Section \ref{sec:autoalpha}  and the behavior of sample-awareness over the learning phase.  
Fig. \ref{fig:adapt} shows 
the learning curve of $\alpha$, $D_{JS}(\pi||q)$ and the policy entropy $\mathcal{H}(\pi)$, which are intertwined in the DAC objective function as seen in \eqref{eq:HmixDecomp}.
In the case of DelayedHalfCheetah, $\alpha$ increases to one as time step goes on, and the initially nonzero JS divergence term $D_{JS}(\pi||q)$ diminishes to zero as time goes. This means that the sample action distribution is exploited in the early phase of learning, and DAC operates like SAC as time goes.  On the other hand, in the case of DelayedHopper, the learned $\alpha$ gradually settle down aroung 0.5, and the JS divergence term $D_{JS}(\pi||q)$  is non-zero throughout the learning phase. Thus, it is seen that the proposed $\alpha$-adaptation learns the weighting factor $\alpha$ with a completely different strategy depending on the task, and this leads to better overall performance for each task as seen in Fig. \ref{fig:compmujoco}.

\subsection{Analysis on the Change of $q$}
\label{subsec:furtheranal}

We assumed that the sample action distribution $q$ of the replay buffer $\mathcal{D}$ is fixed for theoretical development and  proof of diverse policy iteration in Section \ref{subsec:pisem}. However, $q$ changes as iteration goes in real situation, so there exist a gap between the assumption and the real situation for DAC.  Changing distribution was considered in some previous work. \citet{hazan2019provably} considered the
change of previous distributions to guarantee convergence,
but they still have a common objective function (i.e., the entropy of state distribution $d^\pi$ induced by policy $\pi$) to maximize.  
In our case, on the other hand, the objective function \eqref{eq:objpi} itself changes over time as $q$ changes, so it is difficult to show convergence with incorporation of the change of $q$. 
Thus, we assumed locally fixed $q$ because $q$ changes slowly when the buffer size is large.  In order to investigate the impact of the lapse in the assumption and check the robustness of DAC with respect to the change speed of $q$,  we performed an additional study. In the study, we maintained the buffer size of the replay buffer $\mathcal{D}$ as $N = $1000k. Then, instead of using original $q$, i.e., the sample action distribution of whole  $\mathcal{D}$, we now used $q'$, which is the sample action distribution of the latest $N'$ samples (we call $\mathcal{D}'$) stored in $\mathcal{D}$ with $N' \leq $1000k. The smaller $N'$ is, the faster $q'$ changes. Then, with others remaining the same, the ratio objective function $\hat{J}_{R^\alpha}(\eta)$ and the target value $\hat{V}(s_t)$ in DAC were changed to incorporate $q'$ as

\vspace{-1.5em}

{\small\begin{align}
&\hat{J}_{R^\alpha}(\eta)=\mathbb{E}_{s_t\sim\mathcal{D}}[\alpha\mathbb{E}_{a_t\sim\pi_\theta}[\log R_\eta^\alpha(s_t,a_t)]\nonumber\\
&\hspace{6.5em}+ (1-\alpha)\mathbb{E}_{a_t\sim q'}[\log(1-R_\eta^\alpha(s_t,a_t))]],\nonumber\\
&\hat{V}(s_t)=\mathbb{E}_{a_t\sim\pi_\theta}[Q_\phi(s_t,a_t)\nonumber\\
&\hspace{6.5em}+\alpha\log R_\eta^\alpha (s_t,a_t) -\alpha\log \alpha\pi_\theta(a_t|s_t)] \nonumber\\
&+ (1-\alpha)\mathbb{E}_{a_t\sim q'}[\log R_\eta^\alpha(s_t,a_t)-\log \alpha \pi_\theta(a_t|s_t)], \nonumber 
\end{align}}where $s_t$ is still drawn from the original buffer $\mathcal{D}$ and only $q'$ considers samples distribution of $\mathcal{D'}$. Hence, $s_t$ drawn from $\mathcal{D}$ may not belong to $\mathcal{D}'$ used to compute $q'$. So, we used generalization: To sample actions from $q'$ for arbitrary states in $\mathcal{D}$, we learned $q'$ by using variational auto-encoder. Fig. 6 shows the corresponding performance. As shown, the performance degrades as $q'$ changes faster by decreasing $N'$ from 1000k to 1k (Note that the original DAC is the case when $N'=$1000k), but the performance is quite robust against the $q$ change speed. Note that the performance is better than SAC even for $N'=$1k.

\begin{figure}[h]
    \centering
    \includegraphics[width=0.3\textwidth]{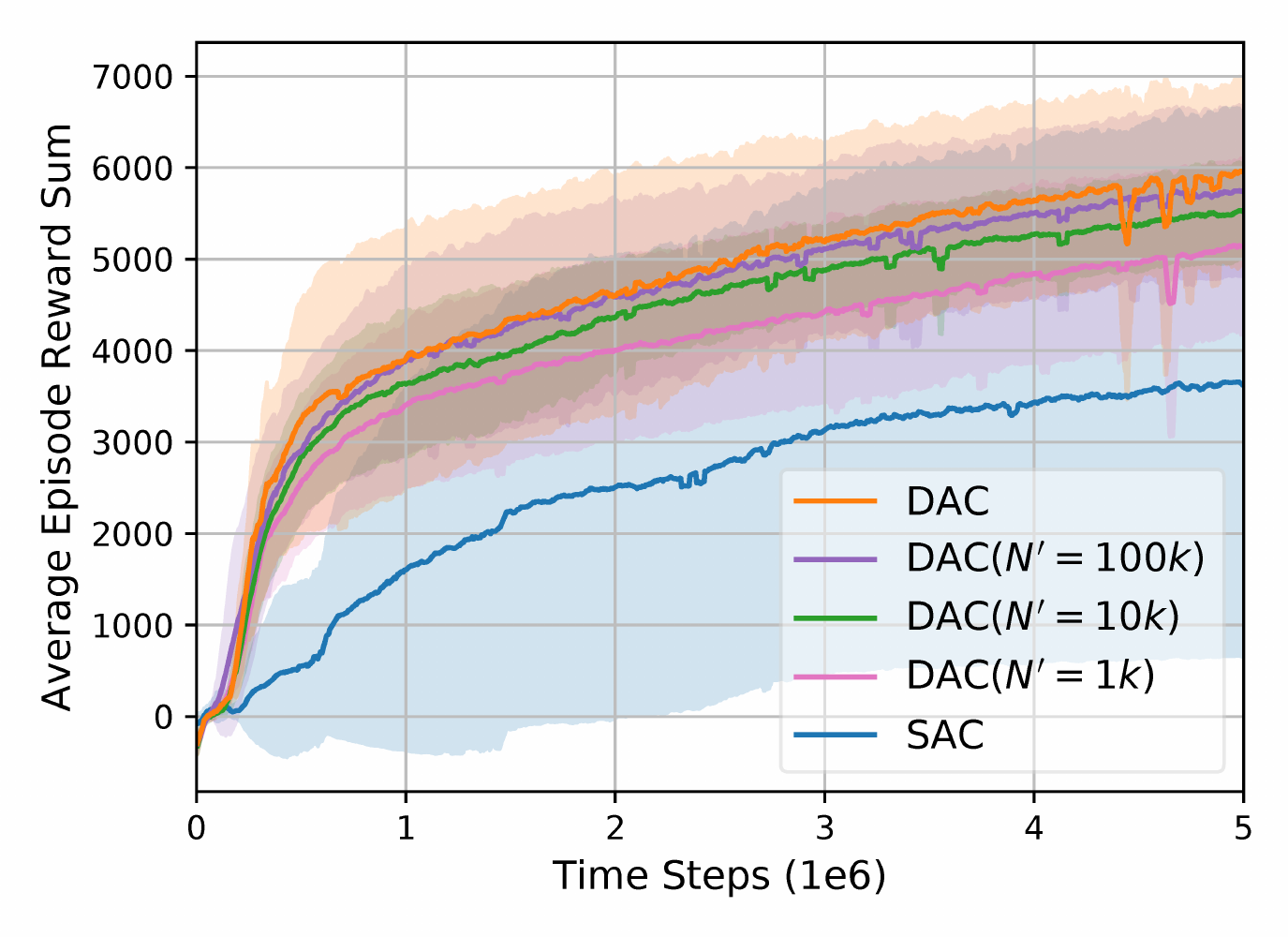}\label{fig:qspeed}
    \vspace{-0.5em}
    \caption{Robustness against the change speed of $q$}
\end{figure}


We provided more results including the max average return table, more ablation study (control coefficient $c$, entropy coefficient $\beta$, and the effect of JS divergence)  and the performance comparison  with various state-of-the-art RL algorithms in Appendix \ref{appendSection:PerComp}. 
The results there also show that DAC yields top level performance.

\section{Conclusion} \label{sec:conclusion}

In this paper, we have proposed a sample-aware entropy framework for off-policy RL to overcome the limitation of simple policy entropy regularization.
With the sample-aware entropy regularization, we can achieve diversity gain by exploiting sample history in the replay buffer in addition to policy entropy for  sample-efficient exploration.
For practical implementation of sample-aware entropy regularized RL, we have used the ratio function to make  computation of the sample action distribution from the replay buffer unnecessary, and have proposed the DAC algorithm with convergence proof.
We have also provided an adaptation method for DAC to automatically control the ratio of the sample action distribution to the policy distribution. Numerical results show that the proposed DAC algorithm significantly outperforms other state-of-the-art RL algorithms. 

\section{Acknowledgement} \label{sec:ack}

This work was supported in part by the ICT R\&D program of MSIP/IITP(2016-0-00563, Research
on Adaptive Machine Learning Technology Development for Intelligent Autonomous Digital Companion) and in part by Basic Science Research Program through the National Research Foundation of Korea (NRF) funded by the Ministry of Science, ICT \& Future Planning(NRF2017R1E1A1A03070788).

\newpage

\nocite{langley00}

\bibliography{reference}
\bibliographystyle{icml2021}

\newpage
\appendix
\onecolumn
\counterwithin{table}{section}
\counterwithin{figure}{section}
\counterwithin{equation}{section}
\renewcommand{\theequation}{\thesection.\arabic{equation}}

\vspace{1em}

\section{A Simple Example of Efficiency of Sample-Aware Entropy Maximization}
\label{sec:toy}

\vspace{1em}

Here, we provide a toy example showing the effectiveness of  maximizing the sample-aware entropy defined as the entropy of a mixture distribution $q_{mix}^{\pi,\alpha}=\alpha\pi+(1-\alpha)q$, where $q$ is the sample action distribution of the replay buffer. For this simple toy example, we consider a discrete MDP case in order to show the intuition of sample-aware entropy maximization.

Let us consider a simple 1-step MDP in which $s_0$ is the unique initial state, there exist $N_a$ actions ($\mathcal{A} = \{A_1,\cdots,A_{N_a}\}$), $s_1$ is the terminal state, and $r$ is a deterministic reward function. Then, there exist $N_a$ state-action pairs in total.  Let us assume that we already have $N_a-1$ state-action samples in the replay buffer as $\mathbf{R}=\{(s_0,A_1,r(s_0,A_1)),\cdots,(s_0,A_{N_a-1},r(s_0,A_{N_a-1}))\}$. In order to estimate the Q-function for all state-action pairs, the policy should sample the last action $A_{N_a}$ (Then, we can reuse all samples infinitely to estimate $Q$). Here, we will compare two exploration methods.

1)	First, if we consider the simple entropy maximization, the policy that maximizes its entropy will choose all actions with equal probability $1/{N_a}$ (uniformly). Then, $N_a$ samples should be taken on average by the policy to visit the action $A_{N_a}$.

2)	Second, consider the sample-aware entropy maximization. Here, the sample action distribution $q$ in the buffer becomes $q(a_0|s_0)=1/(N_a-1)$ for $a_0\in\{A_1,\cdots,A_{N_a-1}\}$ and $q(A_{N_a}|s_0)=0$, the mixture distribution becomes $q_{mix}^{\pi,\alpha}=\alpha\pi+(1-\alpha) q$, and we set $\alpha=1/{N_a}$. Then, the policy that maximizes the sample-aware entropy is given by  $\pi(A_{N_a}|s_0)=1$ because this policy makes $q_{mix}^{\pi,\alpha}$ uniform and the sample-aware entropy is maximized.  In this case, we only need one sample to visit the action $A_{N_a}$. In this way, the proposed sample-aware entropy maximization can enhance sample-efficiency for exploration by using the previous sample distribution and choosing a proper $\alpha$. With this motivation, we propose the sample-aware entropy regularization for off-policy RL and an $\alpha$-adaptation method.

\vspace{1em}

\newpage
\section{Proofs}
\label{sec:proofs}

\subsection{Proof of Theorem \ref{thm1}}
\label{pfthm1}

To prove Theorem \ref{thm1}, we first provide two lemmas. For a fixed policy $\pi$, $Q^\pi$ can be estimated by repeating the Bellman backup operator, as stated in Lemma \ref{lem:eval} below. Lemma \ref{lem:eval} is based on usual policy evaluation but has a new ingredient of the ratio function in the proposed sample-aware entropy case.

\begin{customlem}{1}[Diverse Policy Evaluation]
	Define a sequence of diverse $Q$-functions as $Q_{k+1} = \mathcal{T}^\pi Q_k$,  $k\geq0$, where $\pi$ is a fixed policy and $Q_0$ is a real-valued initial $Q$. Assume that the action space is bounded, and $R^{\pi,\alpha}(s_t,a_t) \in (0,1)$ for all $(s_t,a_t)\in\mathcal{S}\times\mathcal{A}$. Then, the sequence $\{Q_k\}$ converges to the true diverse state-action value $Q^\pi$.
	\label{lem:eval}
\end{customlem}

\emph{Proof.} Let $r_{\pi,t} := \frac{1}{\beta}r_t + \gamma \mathbb{E}_{s_{t+1}\sim P}[ \mathbb{E}_{a_{t+1}\sim\pi}[\alpha\log R^{\pi,\alpha}(s_{t+1},a_{t+1}) - \alpha\log\alpha\pi(a_{t+1}|s_{t+1})] + (1-\alpha)\mathbb{E}_{a_{t+1}\sim q}[\log  R^{\pi,\alpha}(s_{t+1},a_{t+1})-\log \alpha\pi(a_{t+1}|s_{t+1})]]$. Then, we can rewrite the modified Bellman equation  \eqref{eq:bellman} into the standard Bellman equation form for the true $Q^\pi$ as follows:
\begin{equation}
\mathcal{T}^\pi Q(s_t,a_t) = r_{\pi,t} + \gamma\mathbb{E}_{s+1\sim P,~a_{t+1\sim\pi}}[Q(s_{t+1},a_{t+1})]
\end{equation}
Under the assumption of a bounded action space and $R^{\pi,\alpha} \in (0,1)$, the reward $r_{\pi,t}$ is bounded and  the convergence is guaranteed as the usual policy evaluation \citep{sutton1998reinforcement, haarnoja2018soft}.  \hfill{$\square$}

\vspace{2em}

Lemma \ref{lem:improve} is about  diverse policy improvement.

\begin{customlem}{2}[Diverse Policy Improvement]
	Let $\pi_{new}$ be the updated policy obtained by solving $\pi_{new} = \underset{\pi}{\arg\max}~J_{\pi_{old}}(\pi)$, where  $J_{\pi_{old}}(\pi)$ is given in \eqref{eq:pracobj}. 	Then, $Q^{\pi_{new}}(s_t,a_t)\geq Q^{\pi_{old}}(s_t,a_t)$,  $\forall~ (s_t,a_t)\in\mathcal{S}\times\mathcal{A}$.
	\label{lem:improve}
\end{customlem}

\emph{Proof.} Since $\pi_{new} = \underset{\pi}{\arg\max}~J_{\pi_{old}}(\pi)$, we have  $J_{\pi_{old}}(\pi_{new})\geq J_{\pi_{old}}(\pi_{old})$. Expressing 
$J_{\pi_{old}}(\pi_{new})$ and $J_{\pi_{old}}(\pi_{old})$ by 
using the definition of $J_{\pi_{old}}(\pi)$  in \eqref{eq:pracobj}, we have
\begin{align}
J_{\pi_{old}}(\pi_{new}(\cdot|s_t))&=\mathbb{E}_{a_t\sim\pi_{new}}[Q^{\pi_{old}}(s_t,a_t) + \alpha\log R^{\pi_{new},\alpha}(s_t,a_t) - \alpha\log\alpha\pi_{new}(a_t|s_t)] \nonumber\\
&\quad\quad\quad\quad+ (1-\alpha)\mathbb{E}_{a_t\sim q}[\log R^{\pi_{new},\alpha}(s_t,a_t)-\log \alpha \pi_{new}(a_t|s_t)]\nonumber\\
&\geq J_{\pi_{old}}(\pi_{old}(\cdot|s_t))\nonumber\\
&=\mathbb{E}_{a_t\sim\pi_{old}}[Q^{\pi_{old}}(s_t,a_t) + \alpha\log R^{\pi_{old},\alpha}(s_t,a_t) - \alpha\log\alpha\pi_{old}(a_t|s_t)] \nonumber\\
&\quad\quad\quad\quad+ (1-\alpha)\mathbb{E}_{a_t\sim q}[\log R^{\pi_{old},\alpha}(s_t,a_t)-\log \alpha  \pi_{old}(a_t|s_t)]\nonumber\\
&=V^{\pi_{old}}(s_t)  \label{eq:appendLem2B2}
\end{align}
by the definition of $V^\pi(s_t)$ in \eqref{eq:Vestst}. Then, based on  \eqref{eq:appendLem2B2}, we obtain the following inequality: {\small
\begin{align}
&Q^{\pi_{old}}(s_t,a_t) = \frac{1}{\beta}r_t + \gamma\mathbb{E}_{s_{t+1}\sim P}[V^{\pi_{old}}(s_{t+1})]\nonumber\\
&\quad\stackrel{(a)}{\le} \frac{1}{\beta}r_t + \gamma\mathbb{E}_{s_{t+1}\sim P}\{\mathbb{E}_{a_{t+1}\sim\pi_{new}}[
\underbrace{Q^{\pi_{old}}(s_{t+1},a_{t+1})}_{=\frac{1}{\beta}r_{t+1}+\gamma \mathbb{E}_{s_{t+2}\sim P}[V^{\pi_{old}}(s_{t+2})]} + \alpha\log R^{\pi_{new},\alpha}(s_{t+1},a_{t+1})- \alpha\log\alpha\pi_{new}(a_{t+1}|s_{t+1})]  \nonumber\\
&\hspace{10em}+(1-\alpha)\mathbb{E}_{a_{t+1}\sim q}[\log R^{\pi_{new},\alpha}(s_{t+1},a_{t+1})-\log \alpha \pi_{new}(a_{t+1}|s_{t+1})]\}\nonumber\\
&\quad~~\vdots\nonumber\\
&\quad\leq Q^{\pi_{new}}(s_t,a_t), ~~~\mbox{for each $(s_t,a_t)\in\mathcal{S}\times\mathcal{A}$},  \label{eq:appendLem2B3}
\end{align}
}where Inequality (a) is obtained by applying  Inequality \eqref{eq:appendLem2B2} on $V^{\pi_{old}}(s_{t+1})$, and $Q^{\pi_{old}}(s_{t+1},a_{t+1})$ in the RHS of  Inequality (a) is expressed as 
$\frac{1}{\beta}r_{t+1}+\gamma \mathbb{E}_{s_{t+2}\sim P}[V^{\pi_{old}}(s_{t+2})]$ and Inequality \eqref{eq:appendLem2B2} is then applied on  $V^{\pi_{old}}(s_{t+2})$; this procedure is repeated  to obtain Inequality \eqref{eq:appendLem2B3}.  By \eqref{eq:appendLem2B3}, we have the claim. This concludes proof.
\hfill{$\square$}

\newpage
Now, we prove Theorem \ref{thm1} based on the previous lemmas.

\begin{customthm}{1}[Diverse Policy Iteration]
	By repeating iteration of the diverse policy evaluation and the diverse policy improvement, any initial policy  converges to the optimal policy $\pi^*$ s.t. $Q^{\pi^*}(s_t,a_t)\geq Q^{\pi'}(s_t,a_t)$, $\forall~\pi'\in\Pi$, $\forall~ (s_t,a_t)\in\mathcal{S}\times\mathcal{A}$. Also, such $\pi^*$ achieves maximum $J$, i.e., $J_{\pi^*}(\pi^*)\geq J_{\pi}(\pi)$ for any $\pi\in\Pi$.
\end{customthm}

\emph{Proof.} Let $\Pi$ be the space of policy distributions and let $\{\pi_i,i=0,1,2,\cdots|~\pi_i \in\Pi\}$ be a sequence of policies generated by the following recursion:
\begin{equation}  \label{eq:appendThpiip1def}
\pi_{i+1}=\mathop{\arg\max}_{\pi \in \Pi} J_{\pi_{i}}(\pi) ~~~~~~~~\mbox{with an arbitrary initial policy}~ \pi_0,
\end{equation}
where the objective function $J_{\pi_{i}}(\pi)$ is defined in \eqref{eq:pracobj}.

Proof of convergence of the sequence $\{\pi_i,i=0,1,2,\cdots\}$ to a local optimum is for arbitrary state space $\mathcal{S}$. On the other hand, for proof of convergence of  $\{\pi_i,i=0,1,2,\cdots\}$ to the global optimum, we  assume finite MDP,  as typically assumed for convergence proof in usual policy iteration  (Sutton \& Barto, 1998).

\vspace{1em}

For any state-action pair $(s,a)\in\mathcal{S}\times\mathcal{A}$, each $Q^{\pi_i}(s,a)$ is bounded due to the discount factor $\gamma$ (see \eqref{eq:TrueQpiFunc}), and the sequence $\{Q^{\pi_{i}}(s,a),i=0,1,2,\cdots\}$ is monotonically increasing by Lemma \ref{lem:improve}.
Now, consider two terms  $J_{\pi_{i+1}}(\pi_{i+1}(\cdot|s))$ and $J_{\pi_i}(\pi_{i+1}(\cdot|s))$, which are expressed by the definition of $J_{\pi_{old}}(\pi)$ in \eqref{eq:pracobj}  as follows:
\begin{align}
J_{\pi_{i+1}}(\pi_{i+1}(\cdot|s))&=\beta\{\mathbb{E}_{a\sim\pi_{i+1}}\left[Q^{\pi_{i+1}}(s,a) + \alpha(\log R^{\pi_{i+1},\alpha}(s,a) - \log\alpha\pi_{i+1}(a|s))\right]\nonumber\\
&~~\hspace{2em}+(1-\alpha)\mathbb{E}_{a\sim q}\left[\log R^{\pi_{i+1},\alpha}(s,a)-\log\alpha\pi_{i+1}(a|s)\right]\}\label{eq:appendJpipi1}\\
J_{\pi_{i}}(\pi_{i+1}(\cdot|s))&=\beta\{\mathbb{E}_{a\sim\pi_{i+1}}\left[Q^{\pi_{i}}(s,a) + \alpha(\log R^{\pi_{i+1},\alpha}(s,a) - \log\alpha\pi_{i+1}(a|s))\right]\nonumber\\
&~~\hspace{2em}+(1-\alpha)\mathbb{E}_{a\sim q}\left[\log R^{\pi_{i+1},\alpha}(s,a)-\log\alpha\pi_{i+1}(a|s)\right]\}.\label{eq:appendJpipi2}
\end{align}
Note in \eqref{eq:appendJpipi1} and \eqref{eq:appendJpipi2} that all the terms are the same for $J_{\pi_{i+1}}(\pi_{i+1}(\cdot|s))$ and $J_{\pi_i}(\pi_{i+1}(\cdot|s))$ except $\beta \mathbb{E}_{a\sim \pi_{i+1}}[Q^{\pi_{i+1}}(s,a)]$ in $J_{\pi_{i+1}}(\pi_{i+1}(\cdot|s))$ and $\beta \mathbb{E}_{a\sim \pi_{i+1}}[Q^{\pi_i}(s,a)]$ in $J_{\pi_i}(\pi_{i+1}(\cdot|s))$.  Because $\{Q^{\pi_i}(s,a),i=0,1,2,\cdots\}$ is monotonically increasing by Lemma \ref{lem:improve}, comparing \eqref{eq:appendJpipi1} and \eqref{eq:appendJpipi2} yields  
\begin{equation} \label{eq:appendProofTheo1B7}
J_{\pi_{i+1}}(\pi_{i+1}(\cdot|s)) \geq J_{\pi_i}(\pi_{i+1}(\cdot|s)).
\end{equation} 
 Furthermore, we have  for any $s \in \mathcal{S}$,
\begin{equation} \label{eq:appendProofTheo1B4}
J_{\pi_i}(\pi_{i+1}(\cdot|s))\geq J_{\pi_i}(\pi_i(\cdot|s))
\end{equation}
by the definition of $\pi_{i+1}$ in \eqref{eq:appendThpiip1def}.
Combining  \eqref{eq:appendProofTheo1B7} and \eqref{eq:appendProofTheo1B4}, we have 
\begin{equation}  \label{eq:appendPfLocalMono}
J_{\pi_{i+1}}(\pi_{i+1}(\cdot|s)) \geq J_{\pi_i}(\pi_{i+1}(\cdot|s)) \geq J_{\pi_i}(\pi_i(\cdot|s))
\end{equation}
for any state $s\in\mathcal{S}$. Therefore,  the sequence $\{J_{\pi_i}(\pi_i(\cdot|s)), i=0,1,2,\cdots\}$  is monotonically increasing for any  $s\in\mathcal{S}$. Furthermore, note from \eqref{eq:appendJpipi1} that  $J_{\pi_i}(\pi_i(\cdot|s))$ is bounded for all $i$, because the $Q$-function and the entropy of the mixture distribution are bounded. (Note that the RHS of \eqref{eq:appendJpipi1} except the term $\mathbb{E}_{a \sim \pi_{i+1}}[Q^{\pi_{i+1}}(s,a)]$  is nothing but the entropy of the mixture distribution $\mathcal{H}(q_{mix}^{\pi_{i+1},\alpha})$. Please see \eqref{eq:HmixInRateFunction2} for this.) Note that $J_{\pi_i}(\pi_i)$, which is obtained by setting $\pi_{old}=\pi_i$ and $\pi=\pi_i$ in \eqref{eq:pracobj},  is nothing but  $J(\pi_i)$ with the desired original $J$ defined in \eqref{eq:objpi}.
Hence, by \eqref{eq:appendPfLocalMono} and the boundedness of the sequence $\{J_{\pi_i}(\pi_i)\}$,  convergence to a local optimum of $J$ by the sequence $\{\pi_i, i=0,1,2,\cdots\}$ is guaranteed  by the monotone convergence theorem.

\vspace{1em}

Now, consider convergence to the global optimum. By the monotone convergence theorem, $\{Q^{\pi_i}(s,a),i=0,1,2,\cdots\}$ and $\{J_{\pi_i}(\pi_i(\cdot|s)),i=0,1,2,\cdots\}$ pointwisely converge to their limit functions $Q^*:\mathcal{S}\times\mathcal{A}\rightarrow\mathbb{R}$ and $J^*:\mathcal{S}\rightarrow\mathbb{R}$, respectively. Here, note that $J^*(s)\geq J_{\pi_i}(\pi_i(\cdot|s))$ for any $i$ because the sequence $\{J_{\pi_i}(\pi_i(\cdot|s)),i=0,1,2,\cdots\}$ is monotonically  increasing by \eqref{eq:appendPfLocalMono}. 
By the definition of pointwise convergence,  for any $s\in \mathcal{S}$, for any $\epsilon>0$, there exists a sufficiently large ${N}(s) (> 0)$ depending on $s$ such that  $ J_{\pi_i}(\pi_i(\cdot|s)) \geq J^*(s) -\frac{ \epsilon(1-\gamma)}{\gamma}$  for all $i\geq {N}(s)$.  When $\mathcal{S}$ is finite, we set $\bar{N}= \max_{s}N(s)$. Then, we have
\begin{equation}  \label{eq:appendTh1B1010}
    J_{\pi_i}(\pi_i(\cdot|s)) \geq J^*(s) -\frac{ \epsilon(1-\gamma)}{\gamma}, ~~~\forall s \in \mathcal{S},~\forall i \ge \bar{N}
\end{equation}
Furthermore,  we have  
\begin{equation}  \label{eq:appendTh1PfforJpiforallpiprime}
J_{\pi_i}(\pi_i(\cdot|s)) \geq J_{\pi_i}(\pi'(\cdot|s)) -  \frac{\epsilon(1-\gamma)}{\gamma},~~~~\forall s \in \mathcal{S},~\forall i \ge \bar{N},~\forall \pi' \in \Pi.
\end{equation}
\eqref{eq:appendTh1PfforJpiforallpiprime} is valid by the following reason. Suppose that \eqref{eq:appendTh1PfforJpiforallpiprime} is not true. Then, there exist some   $s'\in\mathcal{S}$ and some $\pi'\in\Pi$ such that 
\begin{equation}  \label{eq:appendTh1B1212}
J_{\pi_i}(\pi'(\cdot|s')) \stackrel{(b)}{>} J_{\pi_i}(\pi_i(\cdot|s')) + \frac{\epsilon(1-\gamma)}{\gamma}\stackrel{(c)}{\geq} J^*(s'),
\end{equation}
where Inequality (b) is obtained by negating \eqref{eq:appendTh1PfforJpiforallpiprime} and Inequality (c) is obtained by \eqref{eq:appendTh1B1010}.
Moreover, we have 
\begin{equation}  \label{eq:appendTh1B1313}
J_{\pi_{i+1}}(\pi_{i+1}(\cdot|s')) \stackrel{(d)}{\geq} J_{\pi_i}(\pi_{i+1}(\cdot|s'))= \max_{\pi} J_{\pi_i}(\pi(\cdot|s'))  \stackrel{(e)}{\geq} J_{\pi_i}(\pi'(\cdot|s')), 
\end{equation}
where Inequality (d) is valid due to \eqref{eq:appendProofTheo1B7} and Inequality (e) is valid by the definition of $\pi_{i+1}$ given in \eqref{eq:appendThpiip1def}. Combining 
\eqref{eq:appendTh1B1212} and
  \eqref{eq:appendTh1B1313} yields
\begin{equation}  
J_{\pi_{i+1}}(\pi_{i+1}(\cdot|s')) \ge J_{\pi_i}(\pi_{i+1}(\cdot|s'))  \ge J_{\pi_i}(\pi'(\cdot|s'))  > J_{\pi_i}(\pi_i(\cdot|s')) + \frac{\epsilon(1-\gamma)}{\gamma}\geq J^*(s').
\end{equation}
However, this contradicts to the fact that $J^*(s')$ is the limit of the monotone-increasing sequence  $J_{\pi_i}(\pi_i(\cdot|s'))$. Therefore, \eqref{eq:appendTh1PfforJpiforallpiprime} is valid.

\vspace{1em} Based on  \eqref{eq:appendTh1PfforJpiforallpiprime}, we have the following inequality regarding $Q^{\pi_i}(s_t,a_t)$: For any $(s_t,a_t)$, for all $i \ge \bar{N}$,
\begin{align}
Q^{\pi_i}(s_t,a_t) &= \frac{1}{\beta}r_t + \gamma\mathbb{E}_{s_{t+1}\sim P}[V^{\pi_i}(s_{t+1})]\nonumber\\
&= \frac{1}{\beta}r_t + \gamma\mathbb{E}_{s_{t+1}\sim P}[J_{\pi_i}(\pi_i(\cdot|s_{t+1}))]  \nonumber\\
&\stackrel{(f)}{\geq} \frac{1}{\beta}r_t + \gamma\mathbb{E}_{s_{t+1}\sim P}\left[J_{\pi_i}(\pi'(\cdot|s_{t+1}))-\frac{\epsilon(1-\gamma)}{\gamma}\right],~~~\forall \pi' \in \Pi,\nonumber\\
&\stackrel{(g)}{=} \frac{1}{\beta}r_t +  \gamma\mathbb{E}_{s_{t+1}\sim P}\{\mathbb{E}_{a_{t+1}\sim\pi}[Q^{\pi_i}(s_{t+1},a_{t+1}) + \alpha\log R^{\pi',\alpha}(s_{t+1},a_{t+1})- \alpha\log\alpha\pi'(a_{t+1}|s_{t+1})] \nonumber\\
&\hspace{9em} + (1-\alpha)\mathbb{E}_{a_{t+1}\sim q}[\log R^{\pi',\alpha}(s_{t+1},a_{t+1})-\log \alpha \pi'(a_{t+1}|s_{t+1})]\} - \epsilon(1-\gamma)\nonumber\\
&~~\vdots\nonumber\\
&\stackrel{(h)}{\geq} Q^{\pi'}(s_t,a_t) - \epsilon,~~~\forall \pi' \in \Pi,  \label{eq:appendTh1PfineqH}
\end{align}
where Inequality (f) is valid due to \eqref{eq:appendTh1PfforJpiforallpiprime}; Equality (g) is obtained by explicitly expressing $J_{\pi_i}(\pi')$ using \eqref{eq:pracobj}; we express $Q^{\pi_i}(s_{t+1},a_{t+1})$ as $Q^{\pi_i}(s_{t+1},a_{t+1}) = \frac{1}{\beta}r_{t+1} + \gamma\mathbb{E}_{s_{t+2}\sim P}[V^{\pi_i}(s_{t+2})]$ and repeat the same procedure on $V^{\pi_i}(s_{t+2})=J_{\pi_i}(\pi_i(\cdot|s_{t+2}))$; and we obtain the last Inequality (h) by repeating this iteration.  Here, the resulting constant term is $-\epsilon(1-\gamma)-\epsilon \gamma(1-\gamma)-\epsilon \gamma^2(1-\gamma) - \cdots = -\epsilon$,  as shown in the RHS of Inequality (g).  Note that the uniformity condition "$\forall s \in \mathcal{S}$" in the  Inequality \eqref{eq:appendTh1PfforJpiforallpiprime} is required because we need to express $J_{\pi_i}(\pi_i(\cdot|s_{t+1})),~J_{\pi_i}(\pi_i(\cdot|s_{t+2})),~J_{\pi_i}(\pi_i(\cdot|s_{t+3})),~\cdots$ in terms of $J_{\pi_i}(\pi'(\cdot|s_{t+1})),~J_{\pi_i}(\pi'(\cdot|s_{t+2})),~J_{\pi_i}(\pi'(\cdot|s_{t+3})),~\cdots$, respectively, by using  \eqref{eq:appendTh1PfforJpiforallpiprime} in the above recursive procedure and the support of each element of the sequence  $s_{t+1},~s_{t+2},~s_{t+3},\cdots$ is $\mathcal{S}$ in general.  Since $\epsilon > 0$ is arbitrary in the above, by taking $i\rightarrow \infty$ on both sides of \eqref{eq:appendTh1PfineqH}, we have
\begin{equation}  \label{eq:appendTh1QlastIneq}
    Q^{\pi_\infty}(s,a)\geq Q^{\pi'}(s,a),~~~\forall \pi'\in\Pi, ~~~\forall~(s,a)\in\mathcal{S}\times\mathcal{A}
\end{equation}
since the sequence $\{ Q^{\pi_i}(s,a),i=0,1,2,\cdots\}$ is monotonically increasing.

\vspace{1em}
Now, let us compare  $J_{\pi'}(\pi'(\cdot|s))$ and  $J_{\pi_\infty}(\pi'(\cdot|s))$.  These two terms can be expressed in similar forms  to 
\eqref{eq:appendJpipi1} and \eqref{eq:appendJpipi2}, respectively.  Then, only $Q^{\pi_\infty}(s,a)$ and $Q^{\pi'}(s,a)$ are different in the expressed forms.  Comparing $J_{\pi'}(\pi'(\cdot|s))$ and  $J_{\pi_\infty}(\pi'(\cdot|s))$ as we did for \eqref{eq:appendProofTheo1B7}, we have
\begin{equation}  \label{eq:appendTh1FinalEq1}
 J_{\pi_\infty}(\pi'(\cdot|s)) \ge    J_{\pi'}(\pi'(\cdot|s)) 
\end{equation}
due to   Inequality \eqref{eq:appendTh1QlastIneq}. In addition, we have  $J_{\pi_i}(\pi_i(\cdot|s))  \geq J_{\pi_i}(\pi'(\cdot|s))- \frac{\epsilon(1-\gamma)}{\gamma}$ due to \eqref{eq:appendTh1PfforJpiforallpiprime}.
Since $\epsilon >0$ is arbitrary, by taking $i \rightarrow \infty$, we have
\begin{equation} \label{eq:appendTh1FinalEq2}
J_{\pi_\infty}(\pi_\infty(\cdot|s))  \geq J_{\pi_\infty}(\pi'(\cdot|s)).
\end{equation}
Finally, combining \eqref{eq:appendTh1FinalEq1} and  \eqref{eq:appendTh1FinalEq2} yields
\begin{equation}
    J_{\pi_\infty}(\pi_\infty(\cdot|s))\geq J_{\pi_\infty}(\pi'(\cdot|s)) \geq J_{\pi'}(\pi'(\cdot|s)),~~\forall~\pi'\in\Pi,~~~\forall~s\in\mathcal{S}.
\end{equation}
Recall that $J_{\pi}(\pi)$, which is obtained by setting $\pi_{old}=\pi$ and $\pi=\pi$ in \eqref{eq:pracobj},  is nothing but  $J(\pi)$ of the desired original $J$ defined in \eqref{eq:objpi}. Therefore, 
 $\pi_\infty$ is the optimal policy $\pi^*$ maximizing $J$, and  $\{\pi_i\}$ converges to the optimal policy $\pi^*$.  This concludes the proof.
 \hfill{$\square$}

\textbf{Remark:} Note that what we actually need for proof of convergence to the global optimum is the uniform convergence of  $J_{\pi_i}(\pi_i(\cdot|s)) \rightarrow J^*(s)$ as functions of $s$ to obtain \eqref{eq:appendTh1PfforJpiforallpiprime}. 
The finite state assumption is one sufficient condition for this. 
In order to guarantee convergence to global optimum in non-finite MDP (e.g. continuous state-space), we  need more assumption as considered in \cite{puterman1979convergence,santos2004convergence}. Here, we do not further detail. In this paper, we just consider function approximation for the policy and the value functions  to implement the diverse policy iteration in continuous state and action spaces, based on the convergence proof in finite MDP.

\subsection{Proof of Theorem \ref{thm2}}
\label{pfthm2}

\vspace{1em}
\textbf{Remark:}  We defined $J_{\pi_{old}}(\pi)$ as \eqref{eq:pracobj}, which is restated below:
\begin{align}
&J_{\pi_{old}}(\pi(\cdot|s_t)):=\beta\{\mathbb{E}_{a_t\sim\pi}\left[Q^{\pi_{old}}(s_t,a_t) + \alpha(\log R^{\pi,\alpha}(s_t,a_t) - \log\alpha\pi(a_t|s_t))\right]\nonumber\\
&~~\hspace{11em}+(1-\alpha)\mathbb{E}_{a_t\sim q}\left[\log R^{\pi,\alpha}(s_t,a_t)-\log\alpha\pi(a_t|s_t)\right]\},
\end{align}
where $\pi$ in the $R^{\pi,\alpha}$ terms inside the expectations is the optimization argument. As mentioned in the main part of the paper, this facilitates proof of Lemma \ref{lem:improve} and proof of Theorem \ref{thm1}, especially in Steps
\eqref{eq:appendLem2B2},
  \eqref{eq:appendLem2B3},
  \eqref{eq:appendJpipi1},  \eqref{eq:appendJpipi2}, and    
 \eqref{eq:appendProofTheo1B7}. However, as explained in the main part of the paper, implementing the function $R^{\pi,\alpha}(s_t,a_t)$ with optimization argument $\pi$ is difficult. Hence, we replaced $J_{\pi_{old}}(\pi)$ with $\tilde{J}_{\pi_{old}}(\pi)$ in 
  \eqref{eq:finalObj} by considering  the ratio function $R^{\pi_{old},\alpha}(s_t,a_t)$ for only the current policy $\pi_{old}$. Now, we prove the gradient equivalence of $J_{\pi_{old}}(\pi)$ and $\tilde{J}_{\pi_{old}}(\pi)$ at $\theta=\theta_{old}$ for parameterized policy $\pi_\theta$.

\vspace{1em}

\begin{customlem}{3} For the ratio function $R^{\pi,\alpha}(s_t,a_t)$ defined in \eqref{eq:Rphialpha}, we have the following:
\begin{equation}  
\log R^{\pi,\alpha}(s_t,a_t)-\log \alpha\pi(a_t|s_t) =  \log (1-R^{\pi,\alpha}(s_t,a_t))-\log ((1-\alpha)q(a_t|s_t))
\end{equation}
\end{customlem}

\emph{Proof.} From the definition of the ratio function:
\begin{equation}  
R^{\pi,\alpha}(s_t,a_t) = \frac{\alpha\pi(a_t|s_t)}{\alpha \pi(a_t|s_t) + (1-\alpha)q(a_t|s_t)},
\end{equation}
we have
\begin{equation}  
1-R^{\pi,\alpha}(s_t,a_t) = \frac{(1-\alpha)q(a_t|s_t)}{\alpha \pi(a_t|s_t) + (1-\alpha)q(a_t|s_t)}.
\end{equation}
Hence, we have
\begin{align}
\log \frac{1}{\alpha \pi(a_t|s_t) + (1-\alpha)q(a_t|s_t)}&=\log R^{\pi,\alpha}(s_t,a_t)-\log (\alpha\pi(a_t|s_t))\\
&=\log (1-R^{\pi,\alpha}(s_t,a_t))-\log ((1-\alpha)q(a_t|s_t)).
\end{align}
This concludes  proof. \hfill{$\square$}

\vspace{1em}

\begin{customthm}{2}
   Consider the new objective function for policy improvement $\tilde{J}_{\pi_{old}}(\pi(\cdot|s_t))$ in \eqref{eq:finalObj}, where the ratio function inside the expectation in \eqref{eq:finalObj} is the ratio function for the given current policy $\pi_{old}$. Suppose that the policy is parameterized with parameter $\theta$. Then, for parameterized policy $\pi_\theta$, the two objective functions $J_{\pi_{\theta_{old}}}(\pi_\theta(\cdot|s_t))$ and $\tilde{J}_{\pi_{\theta_{old}}}(\pi_\theta(\cdot|s_t))$ have the same gradient direction for $\theta$ at $\theta=\theta_{old}$ for all $s_t\in\mathcal{S}$, where $\theta_{old}$ is the parameter of the given current policy $\pi_{old}$.
\end{customthm}

\emph{Proof.}  With the parameterized $\pi_\theta$, the two objective functions are expressed as 
\begin{align}
J_{\pi_{\theta_{old}}}(\pi_\theta(\cdot|s_t))&=\beta(\mathbb{E}_{a_t\sim\pi_\theta}[Q^{\pi_{\theta_{old}}}(s_t,a_t) + \alpha\log R^{\pi_\theta,\alpha}(s_t,a_t) - \alpha\log \alpha\pi_\theta(a_t|s_t)]\nonumber\\
&+(1-\alpha)\mathbb{E}_{a_t\sim q}[\log R^{\pi_\theta,\alpha}(s_t,a_t)-\log \alpha\pi_\theta(a_t|s_t)])\nonumber\\
&\stackrel{(1)}{=}\beta(\mathbb{E}_{a_t\sim\pi_\theta}[Q^{\pi_{\theta_{old}}}(s_t,a_t) + \alpha\log R^{\pi_\theta,\alpha}(s_t,a_t) - \alpha\log \alpha\pi_\theta(a_t|s_t)]\nonumber\\
&+(1-\alpha)\mathbb{E}_{a_t\sim q}[\log (1-R^{\pi_\theta,\alpha}(s_t,a_t))-\log (1-\alpha)q(a_t|s_t)])\label{eq:appendTh2Jorg}\\
\tilde{J}_{\pi_{\theta_{old}}}(\pi_\theta(\cdot|s_t))&=\beta\mathbb{E}_{a_t\sim\pi_\theta}[Q^{\pi_{\theta_{old}}}(s_t,a_t)+\alpha\log R^{\pi_{\theta_{old}},\alpha}(s_t,a_t)-\alpha\log\pi_\theta(a_t|s_t)], \label{eq:appendTh2Jtilde}
\end{align}
where  Step $(1)$ is valid by Lemma 3.
Comparing \eqref{eq:appendTh2Jorg} and \eqref{eq:appendTh2Jtilde},
we can ignore the common $Q^{\pi_{\theta_{old}}}$ and  $\log \pi_\theta$ terms, and the constant terms w.r.t. $\theta$ that yield zero gradient in \eqref{eq:appendTh2Jorg} and \eqref{eq:appendTh2Jtilde}. Therefore, we  only need to show
\begin{equation}
\label{eq:thm1}
\nabla_\theta\{\alpha\mathbb{E}_{a_t\sim\pi_\theta}[\log R^{\pi_\theta,\alpha}]+(1-\alpha)\mathbb{E}_{a_t\sim q}[\log (1-R^{\pi_\theta,\alpha})]\} =\nabla_\theta \mathbb{E}_{a_t\sim\pi_\theta}[\alpha\log R^{\pi_{\theta_{old}},\alpha}]
\end{equation}
 at $\theta=\theta_{old}$.
The gradient of the left-hand side (LHS)  in \eqref{eq:thm1} at $\theta=\theta_{old}$ is  expressed as
\begin{align}
&\nabla_\theta\{\alpha\mathbb{E}_{a_t\sim\pi_\theta}[\log R^{\pi_\theta,\alpha}]+(1-\alpha)\mathbb{E}_{a_t\sim q}[\log (1-R^{\pi_\theta,\alpha})]\}\nonumber\\
&= \nabla_\theta\left\{\alpha \int_{a_t} \pi_\theta \log R^{\pi_\theta,\alpha} da_t+(1-\alpha)\int_{a_t} q \log (1-R^{\pi_\theta,\alpha})da_t\right\}\nonumber\\ 
&= \alpha \int_{a_t} (\nabla_\theta\pi_\theta) \log R^{\pi_\theta,\alpha} da_t+\alpha \int_{a_t} \pi_\theta (\nabla_\theta\log R^{\pi_\theta,\alpha}) da_t+(1-\alpha)\int_{a_t} q \nabla_\theta\log (1-R^{\pi_\theta,\alpha})da_t\nonumber\\ 
&=\alpha \int_{a_t} (\nabla_\theta\pi_\theta)|_{\theta=\theta_{old}} \log R^{\pi_\theta,\alpha}|_{\theta=\theta_{old}} da_t+\alpha \int_{a_t} \pi_\theta (\nabla_\theta\log R^{\pi_\theta,\alpha}) da_t+(1-\alpha)\int_{a_t} q \nabla_\theta\log (1-R^{\pi_\theta,\alpha})da_t\nonumber\\
&=\alpha \nabla_\theta \int_{a_t} \pi_\theta \log R^{\pi_{\theta_{old}},\alpha} da_t+\alpha \int_{a_t} \pi_\theta (\nabla_\theta\log R^{\pi_\theta,\alpha}) da_t+(1-\alpha)\int_{a_t} q \nabla_\theta\log (1-R^{\pi_\theta,\alpha})da_t\nonumber\\ 
&=\nabla_\theta\mathbb{E}_{a_t\sim\pi_\theta}[\alpha\log R^{\pi_{\theta_{old}},\alpha}]+ \alpha\mathbb{E}_{a_t\sim\pi_\theta}[\nabla_\theta\log R^{\pi_\theta,\alpha}]+(1-\alpha)\mathbb{E}_{a_t\sim q}[\nabla_\theta\log (1-R^{\pi_\theta,\alpha})].\label{eq:thmprf}
\end{align}
Here, the gradient of the last two terms in the RHS of  \eqref{eq:thmprf} becomes zero, as shown below:
\begin{align}\label{eq:thmprf2}
&\alpha\mathbb{E}_{a_t\sim\pi_\theta}[\nabla_\theta\log R^{\pi_\theta,\alpha}]+(1-\alpha)\mathbb{E}_{a_t\sim q}[\nabla_\theta\log (1-R^{\pi_\theta,\alpha})]\nonumber\\
&=\alpha\mathbb{E}_{a_t\sim\pi_\theta}\left[\frac{\nabla_\theta R^{\pi_\theta,\alpha}}{ R^{\pi_\theta,\alpha}}\right]+(1-\alpha)\mathbb{E}_{a_t\sim q}\left[\frac{\nabla_\theta (1-R^{\pi_\theta,\alpha})}{(1-R^{\pi_\theta,\alpha})}\right]\nonumber\\
&=\alpha\mathbb{E}_{a_t\sim\pi_\theta}\left[\frac{\nabla_\theta R^{\pi_\theta,\alpha}}{ R^{\pi_\theta,\alpha}}\right]-(1-\alpha)\mathbb{E}_{a_t\sim q}\left[\frac{\nabla_\theta R^{\pi_\theta,\alpha}}{(1-R^{\pi_\theta,\alpha})}\right]\nonumber\\
&=\alpha\mathbb{E}_{a_t\sim\pi_\theta}\left[\frac{\nabla_\theta R^{\pi_\theta,\alpha}}{ R^{\pi_\theta,\alpha}}\right]-(1-\alpha)\mathbb{E}_{a_t\sim q}\left[\frac{\alpha\pi_\theta + (1-\alpha)q}{(1-\alpha) q}\cdot\nabla_\theta R^{\pi_\theta,\alpha}\right]\nonumber\\
&=\alpha\mathbb{E}_{a_t\sim\pi_\theta}\left[\frac{\nabla_\theta R^{\pi_\theta,\alpha}}{ R^{\pi_\theta,\alpha}}\right]-\mathbb{E}_{a_t\sim q}\left[\frac{\alpha\pi_\theta + (1-\alpha)q}{q}\cdot\nabla_\theta R^{\pi_\theta,\alpha}\right]\nonumber\\
&\stackrel{(2)}{=}\alpha\mathbb{E}_{a_t\sim\pi_\theta}\left[\frac{\nabla_\theta R^{\pi_\theta,\alpha}}{ R^{\pi_\theta,\alpha}}\right]-\mathbb{E}_{a_t\sim \pi_\theta}\left[\frac{\pi_\theta + (1-\alpha)q}{ \pi_\theta}\cdot\nabla_\theta R^{\pi_\theta,\alpha}\right]\nonumber\\
&=\alpha\mathbb{E}_{a_t\sim\pi_\theta}\left[\frac{\nabla_\theta R^{\pi_\theta,\alpha}}{ R^{\pi_\theta,\alpha}}\right]-\alpha\mathbb{E}_{a_t\sim \pi_\theta}\left[\frac{\pi_\theta + (1-\alpha)q}{\alpha \pi_\theta}\cdot\nabla_\theta R^{\pi_\theta,\alpha}\right]\nonumber\\
&=\alpha\mathbb{E}_{a_t\sim\pi_\theta}\left[\frac{\nabla_\theta R^{\pi_\theta,\alpha}}{ R^{\pi_\theta,\alpha}}\right]-\alpha\mathbb{E}_{a_t\sim \pi_\theta}\left[\frac{\nabla_\theta R^{\pi_\theta,\alpha}}{R^{\pi_\theta,\alpha}}\right] = 0,
\end{align}
where we used an importance sampling technique (i.e., measure change) $\mathbb{E}_{a_t\sim q}[f(s_t,a_t)]=\mathbb{E}_{a_t\sim\pi_\theta}\left[\frac{q(a_t|s_t)}{\pi_\theta(a_t|s_t)}f(s_t,a_t)\right]$ for Step $(2)$. By \eqref{eq:thmprf} and \eqref{eq:thmprf2}, $J_{\pi_{\theta_{old}}}(\pi_\theta(\cdot|s_t))$ and $J_{\pi_{\theta_{old}}}(\pi_\theta(\cdot|s_t))$ have the same gradient at $\theta=\theta_{old}$.  This concludes proof. \hfill{$\square$}

\newpage

\section{Detailed DAC Implementation}
\label{sec:imple}

\vspace{1em}

We defined the target value $\hat{V}(s_t)=\mathbb{E}_{a_t\sim\pi_\theta}[Q_\phi(s_t,a_t)+\alpha\log R_\eta^\alpha (s_t,a_t) -\alpha\log \alpha\pi_\theta(a_t|s_t)]
+ (1-\alpha)\mathbb{E}_{a_t\sim \mathcal{D}}[\log R_\eta^\alpha(s_t,a_t)-\log \alpha \pi_\theta(a_t|s_t)]$ in \eqref{eq:tarv}. However, the probability of $\pi$ for actions sampled from $\mathcal{D}$ can have high variance, so we clip the term inside the expectation over $a_t\sim\mathcal{D}$ by action dimension for stable learning. Thus, the final target value is given by
\begin{align}
\hat{V}(s_t)&=\mathbb{E}_{a_t\sim\pi_\theta}[Q_\phi(s_t,a_t)+\alpha\log R_\eta^\alpha (s_t,a_t) -\alpha\log \alpha\pi_\theta(a_t|s_t)] \nonumber\\
&\quad+ (1-\alpha)\mathbb{E}_{a_t\sim \mathcal{D}}[\mathrm{clip}(\log R_\eta^\alpha(s_t,a_t)-\log \alpha \pi(a_t|s_t);-d,d)],
\label{eq:tarvc}
\end{align}
where $d=\mathrm{dim}(\mathcal{A})$ is the action dimension and $\mathrm{clip}(x;-d,d)$ is the clipping function to fit into the range $[-d,d]$. We  use \eqref{eq:tarvc} for actual implementation. 

In addition, we require $R^{\pi_\theta,\alpha} \in (\epsilon,1-\epsilon)$  in the proofs of Theorems 1 and 2 so that $\log R^{\pi_\theta,\alpha}$ and $\log (1-R^{\pi_\theta,\alpha} )$ appearing in the proofs do not diverge.  For practical implementation, we clipped the ratio function $R^\alpha$ as $(\epsilon,1-\epsilon)$ for small $\epsilon>0$ since some $q$ values can be close to zero before the replay buffer stores a sufficient amount of samples. However, $\pi$ is always non-zero since we consider Gaussian policy.

To compute the gradient of $\hat{J}_{\pi}(\theta)$ in  \eqref{eq:Jpihat}, we use the  reparameterization trick proposed by \citep{kingma2013auto, haarnoja2018soft}. Note that the policy action $a_t \sim \pi_\theta$ is the output of the policy neural network with parameter $\theta$. So, it can be viewed as $a_t = f_\theta (\epsilon_t;s_t)$, where $f$ is a function parameterized by $\theta$ and $\epsilon_t$ is a noise vector sampled from spherical normal distribution $\mathcal{N}$. Then, the gradient of $\hat{J}_{\pi}(\theta)$ is represented as $\nabla_{\theta}\hat{J}_{\pi}(\theta) = \mathbb{E}_{s_t\sim\mathcal{D},~\epsilon_t\sim\mathcal{N}}[\nabla_a(Q_\phi(s_t,a) + \alpha \log R_\eta^\alpha(s_t,a)- \alpha \log \pi_\theta (a|s_t))|_{a=f_\theta (\epsilon_t;s_t)} \nabla_\theta f_\theta (\epsilon_t;s_t) - \alpha (\nabla_\theta\log\pi_\theta)(f_\theta(\epsilon_t;s_t)|s_t)]$.

\vspace{1em}

\subsection{Detailed Implementation of the $\alpha$-Adaptation}
\label{subsec:gradalpha}

In order to learn $\alpha$, we parameterize $\alpha$ as a function of $s_t$ using parameter $\xi$, i.e.,  $\alpha= \alpha_\xi(s_t)$, and implement $\alpha_\xi(s_t)$ with a neural network. Then,  $\xi$ is updated to minimize the following loss function of $\alpha$ obtained from \eqref{eq:alphamaxmin}:
\begin{align}\label{eq:lossalpha}
\hat{L}_{\alpha}(\xi)&=\mathbb{E}_{s_t\sim\mathcal{D}}[\mathcal{H}(q_{mix}^{\pi_\theta,\alpha_\xi}) - \alpha_\xi c]
\end{align}
In the $\alpha$ adaptation case,  all the updates for diverse policy iteration are the same except that $\alpha$ is replaced with $\alpha_\xi(s_t)$. The gradient of $\hat{L}_{\alpha}(\xi)$ with respect to $\xi$ can be estimated as below:

\sbox0{\begin{adjustbox}{max height=.8\textheight, max width=\textwidth}
\parbox{\linewidth}{
\begin{align}\label{eq:gradalpha}
&\nabla_\xi \hat{L}_{\alpha}(\xi) = \nabla_\xi \mathbb{E}_{s_t\sim\mathcal{D}}[\mathcal{H}(q_{mix}^{\pi_\theta,\alpha_\xi}) - \alpha_\xi c]\nonumber\\
=&\nabla_\xi \mathbb{E}_{s_t\sim\mathcal{D}}[\alpha_\xi\mathbb{E}_{a_t\sim\pi_\theta}[-\log(\alpha_\xi \pi_\theta + (1-\alpha_\xi)q) - c] +(1-\alpha_\xi)\mathbb{E}_{a_t\sim q}[- \log(\alpha_\xi\pi_\theta + (1-\alpha_\xi)q)]]\nonumber\\
=&\mathbb{E}_{s_t\sim\mathcal{D}} [(\nabla_\xi \alpha_\xi)(\mathbb{E}_{a_t\sim\pi_\theta}[-\log(\alpha_\xi \pi_\theta + (1-\alpha_\xi)q)   - c ] - \mathbb{E}_{a_t\sim q}[- \log(\alpha_\xi\pi_\theta + (1-\alpha_\xi)q)])]\nonumber\\
&+ \mathbb{E}_{s_t\sim\mathcal{D}}[\alpha_\xi\mathbb{E}_{a_t\sim\pi_\theta}[-\nabla_\xi\log(\alpha_\xi \pi_\theta + (1-\alpha_\xi)q) ] +(1-\alpha_\xi)\mathbb{E}_{a_t\sim q}[-\nabla_\xi \log(\alpha_\xi\pi_\theta + (1-\alpha_\xi)q)]]\nonumber \nonumber\\
=&\mathbb{E}_{s_t\sim\mathcal{D}} [(\nabla_\xi \alpha_\xi)(\mathbb{E}_{a_t\sim\pi_\theta}[-\log\alpha_\xi\pi_\theta + \log R^{\pi_\theta,\alpha_\xi}  - c  ] - \mathbb{E}_{a_t\sim q}[\log R^{\pi_\theta,\alpha_\xi} -\log \alpha_\xi \pi_\theta])]\nonumber\\
&+\mathbb{E}_{s_t\sim\mathcal{D}}\bigg[\underset{=0}{\underbrace{\int_{a_t\in\mathcal{A}}(\alpha_\xi \pi_\theta + (1-\alpha_\xi)q)[-\nabla_\xi\log(\alpha_\xi \pi_\theta + (1-\alpha_\xi)q) }}]\bigg]\nonumber \\
=&\mathbb{E}_{s_t\sim\mathcal{D}} [(\nabla_\xi \alpha_\xi)(\mathbb{E}_{a_t\sim\pi_\theta}[-\log\alpha_\xi\pi_\theta + \log R^{\pi_\theta,\alpha_\xi} - c  ] - \mathbb{E}_{a_t\sim q}[\log R^{\pi_\theta,\alpha_\xi}-\log \alpha_\xi \pi_\theta])]
\end{align}
}
\end{adjustbox}}\usebox0
\typeout{\the\textheight,\the\ht0,\the\dp0}

Note that $R^{\pi_\theta,\alpha_\xi}$ can be estimated by the ratio function $R_\eta^{\alpha_{\xi}}$. Here, we use the same clipping technique as used in \eqref{eq:tarvc} for the last term of \eqref{eq:gradalpha}. For $\alpha$-adaptation, we used regularization for $\alpha$ learning and restricted the range of $\alpha$ as $0.5\leq\alpha\leq0.99$ for $\alpha$ adaptation in order to maintain a certain level of entropy regularization and prevent saturation of $R_\eta^\alpha$.

\newpage
\section{Simulation Setup}
\label{sec:simsetup}

\vspace{1em}

We here provide the detailed simulation setup of DAC, SAC baselines, RND, and MaxEnt(State). For fair comparison,  we use the common hyperparameter setup for DAC and SAC baselines except for the parts regarding entropy or divergence. 

The hyperparameter setup basically follows the setup in \citep{haarnoja2018soft}, which is given by Table \ref{table:parameter}. Here, the entropy coefficient $\beta$ is selected based on the ablation study in Section \ref{sec:mainablation}. For the policy space $\Pi$, we considered a Gaussian policy set widely considered in usual continuous RL. Also, we provide Table \ref{table:env}, which shows the environment description, the corresponding entropy control coefficient $\beta$, threshold for sparse Mujoco tasks, and reward delay $D$ for delayed Mujoco tasks.

\begin{table}[!h]
	\centering
	\begin{tabular}{l|C{9em}C{9em}}
	\hline
	 & SAC / SAC-Div & DAC \\
	\hline
	Learning rate $\delta$ & \multicolumn{2}{c}{$3\cdot 10^{-4}$} \\
	Discount factor $\gamma$ & \multicolumn{2}{c}{$0.99$ ($0.999$ for pure exploration)} \\
	Horizon $N$ & \multicolumn{2}{c}{$1000$} \\
	Mini-batch size $M$ & \multicolumn{2}{c}{$256$} \\
	Replay buffer length & \multicolumn{2}{c}{$10^6$} \\
	Smoothing coefficient of EMA for $V_{\bar{\psi}}$ & \multicolumn{2}{c}{$0.005$} \\
	Optimizer & \multicolumn{2}{c}{Adam} \\
	Num. of hidden layers (all networks) & \multicolumn{2}{c}{2} \\
	Size of hidden layers (all networks) & \multicolumn{2}{c}{256} \\
	Policy distribution & \multicolumn{2}{c}{Independent Gaussian distribution} \\
	Activation layer  & \multicolumn{2}{c}{ReLu} \\
	Output layer for $\pi_\theta$, $Q_\phi$, $V_\psi$ , $V_{\bar{\psi}}$ & \multicolumn{2}{c}{Linear} \\
	Output layer for $\alpha_\xi$, $R_\eta^\alpha$  & $\cdot$ & Sigmoid \\
	Regularize coefficient for $\alpha_\xi$  & $\cdot$ & $10^{-3}$ \\
	Control coefficient $c$ for $\alpha$-adaptation & $\cdot$ & $-2.0 \cdot\mathrm{dim}(\mathcal{A})$ \\
	\hline
	\end{tabular}
	\caption{Hyperparamter setup}
	\label{table:parameter}
\end{table}

\begin{table}[!h]
	\centering
	\begin{tabular}{l|C{6em}C{6em}C{6em}C{6em}}
		\hline
		 & State dim. & Action dim. & $\beta$ & Threshold \\
		\hline
		SparseHalfCheetah-v1 & 17 & 6 & 0.02 & 5.0 \\
		SparseHopper-v1 & 11 & 3 & 0.04 & 1.0 \\
		SparseWalker2d-v1 & 17 & 6 & 0.02 & 1.0 \\
		SparseAnt-v1 & 111 & 8 & 0.01 & 1.0 \\
		\hline
		 & State dim. & Action dim. & $\beta$ & Delay $D$ \\
		\hline
		HumanoidStandup-v1 & 376 & 17 & 1 & $\cdot$ \\
		DelayedHalfCheetah-v1 & 17 & 6 & 0.2 & 20 \\
		DelayedHopper-v1 & 11 & 3 & 0.2 & 20 \\
		DelayedWalker2d-v1 & 17 & 6 & 0.2 & 20 \\
		DelayedAnt-v1 & 111 & 8 & 0.2 & 20 \\
		\hline
	\end{tabular}
	\caption{State and action dimensions of Mujoco tasks and the corresponding $\beta$}
	\label{table:env}
\end{table}

In addition, we also compared the performance of DAC to two recent state-based exploration methods, RND \citep{burda2018exploration} and MaxEnt(State) \citep{hazan2019provably}, in Section \ref{sec:experiments}. State-based exploration methods aim to find rare states to enhance exploration performance.

In order to explore rare states, RND adds an intrinsic reward based on prediction error  $r_t^{int} = ||\hat{f}(s_{t+1}) - f(s_{t+1})||^2 $ to the extrinsic reward $r_t^{ext}$ so that the total reward becomes  $r_t=r_t^{ext}+c^{int}r_t^{int}$, where $\hat{f}$ is a prediction network  and $f$ is a randomly fixed target network. Then, the agent goes to rare states since rare states have higher prediction errors. For our simulation, we considered MLP with $2$ ReLu hidden layers of size $256$ with $20$-dimensional output for both networks of RND, and we used $c^{int}=5$ that performed well for  considered tasks. 

On the other hand, MaxEnt(State) aims to maximize the entropy of state mixture distribution $\mathcal{H}(d^{\pi^{mix}})$ to explore rare states, where $d^\pi$ is the state distribution of a trajectory generated from $\pi$. In order to do that, MaxEnt(State) uses the reward  $r_{MaxEnt(State)}(s)=-(\log d^{\pi_{mix}}(s) + c_s)$, where $c_s$ is a smoothing constant. MaxEnt(State) mainly considers large or continuous state space, so $d^{\pi_{mix}}$ is computed by projection/Kernel density estimation. Then, MaxEnt(State) explores the state space better than a simple random policy on various tasks in continuous state spaces. For our simulation, we use previous $100$K states stored in the buffer to estimate $d^{\pi_{mix}}$. Note that MaxEnt(State) is originally designed for pure exploration, but we use its reward functional as an intrinsic reward in order to learn sparse-rewarded tasks. In this case, we found that  $c^{int}=0.02$ worked  well for the considered tasks. For both RND and MaxEnt(State), we basically consider the same simulation setup with DAC and SAC baselines but use Gaussian policy with fixed standard deviation $\sigma=0.3$ for both RND and MaxEnt(State) to make  fair comparison between action-based exploration and state-based exploration.

\newpage

\section{More Results on Performance Comparison}
\label{appendSection:PerComp}

\vspace{.5em}

We provide more numerical results in this section. In Appendix \ref{subsec:appbaseline}, we provide the remaining learning curves and max average return tables for the performance comparisons in the main paper. In Appendix \ref{subsec:rnd}, we provide the performance comparison between DAC and RND/MaxEnt(State) on SparseMujocot tasks. In Appendix \ref{subsec:appothers}, we compare the DAC with $\alpha$ adaptation to other general RL algorithms on HumanoidStandup and DelayedMujoco tasks.


\subsection{Performance Comparison with the SAC Baselines}
\label{subsec:appbaseline}

\begin{figure}[!h]
	\centering
	\subfigure[SparseHopper-v1]{\includegraphics[width=0.31\textwidth]{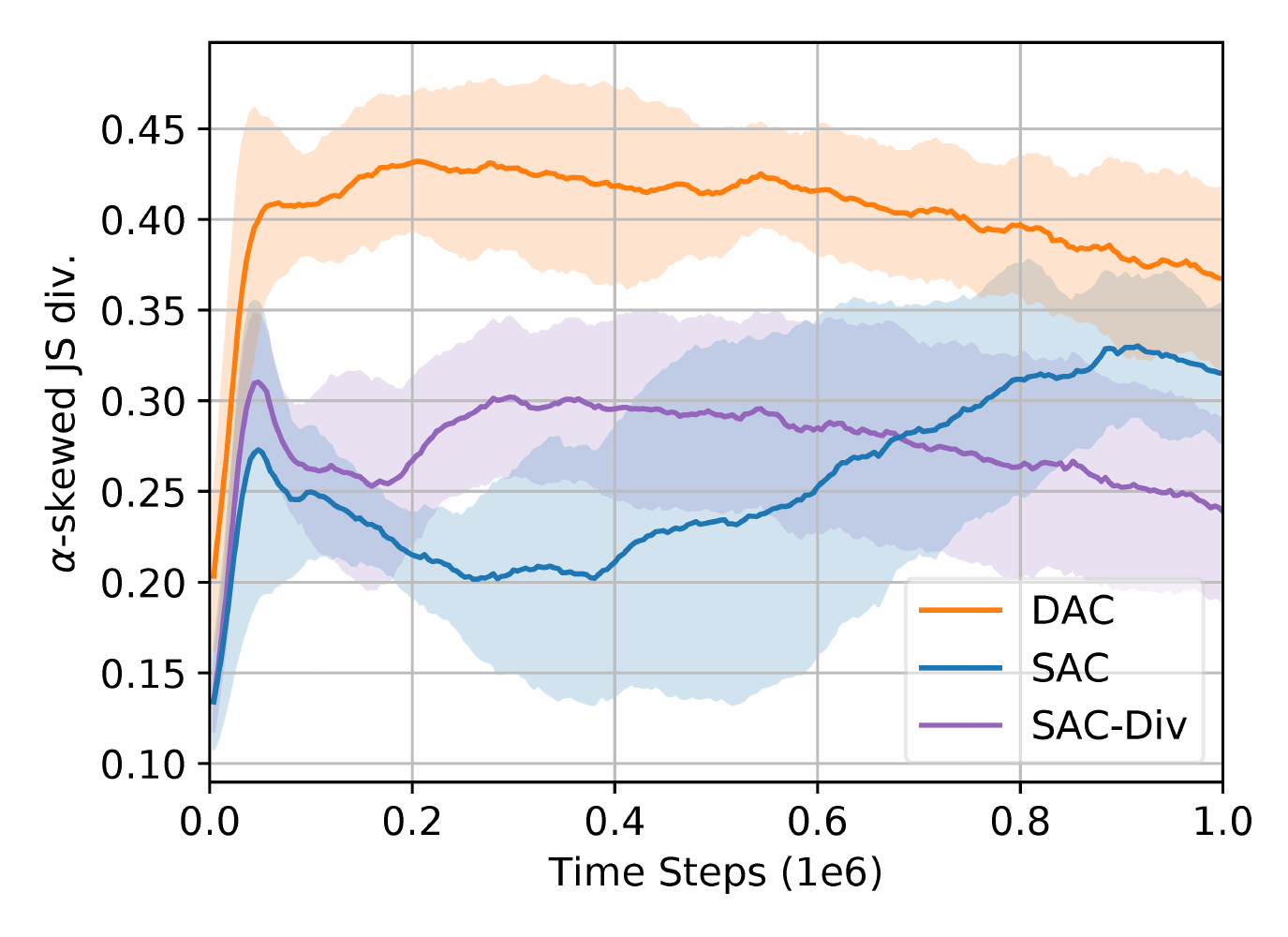}}
	\subfigure[SparseWalker-v1]{\includegraphics[width=0.31\textwidth]{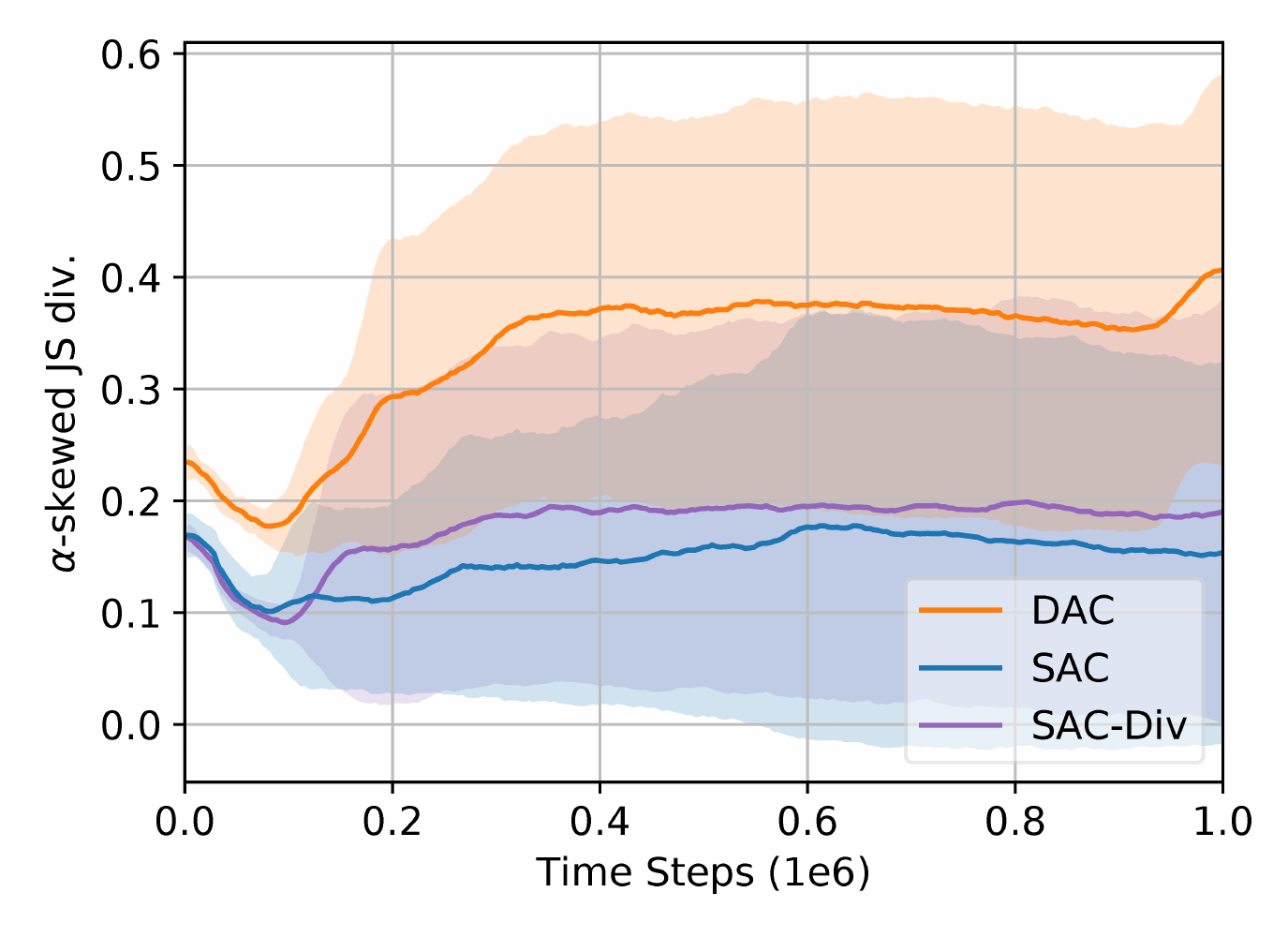}}
	\subfigure[SparseAnt-v1]{\includegraphics[width=0.31\textwidth]{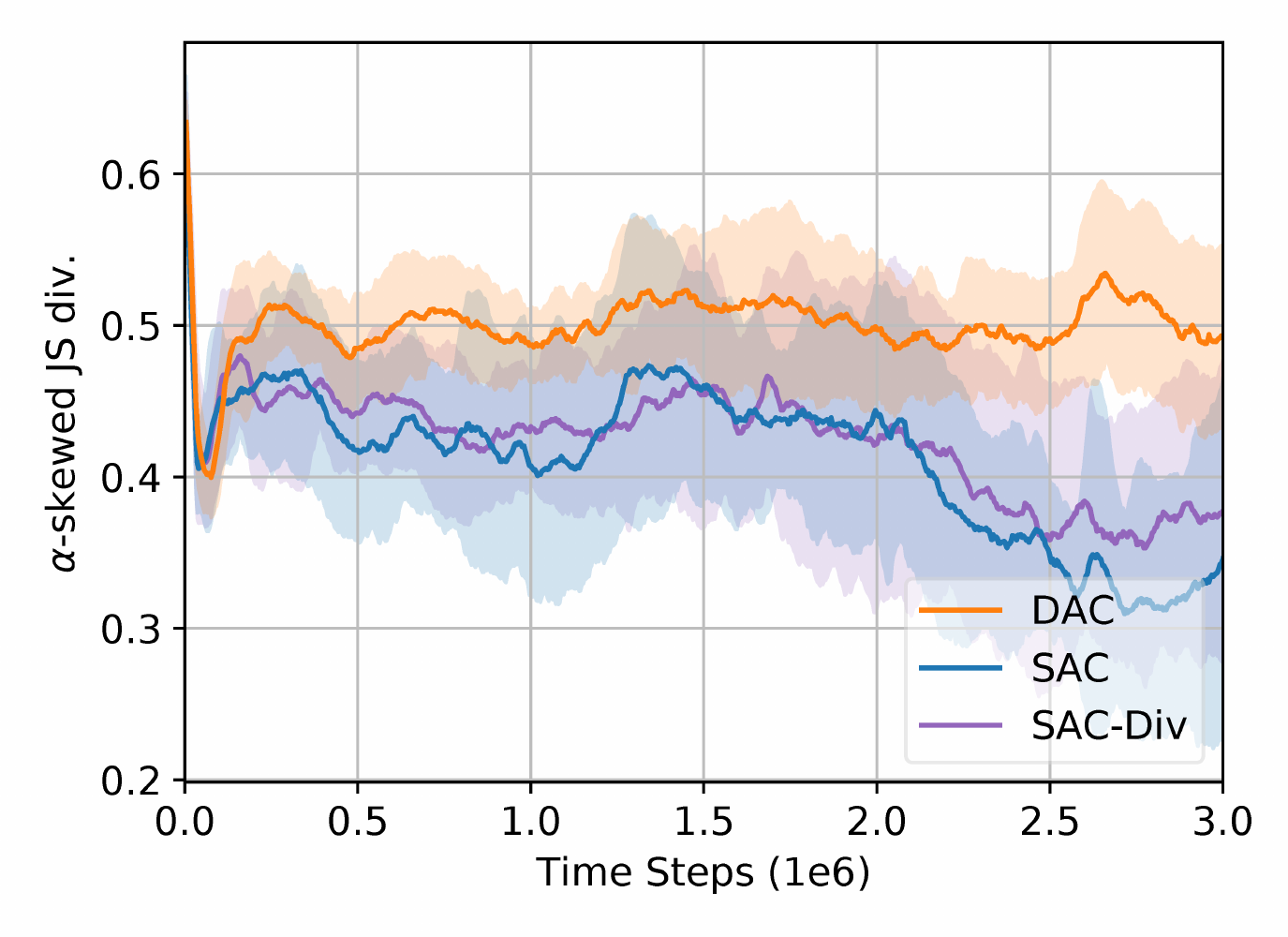}}
	\caption{$\alpha$-skewed JS symmetrization of KLD $D_{JS}^\alpha$ for DAC and SAC/SAC-Div}
	\label{fig:compdivmujoco}
	\vspace{-0.8em}
\end{figure}

\begin{figure}[!h]
	\centering
	\subfigure[SparseHopper-v1]{\includegraphics[width=0.31\textwidth]{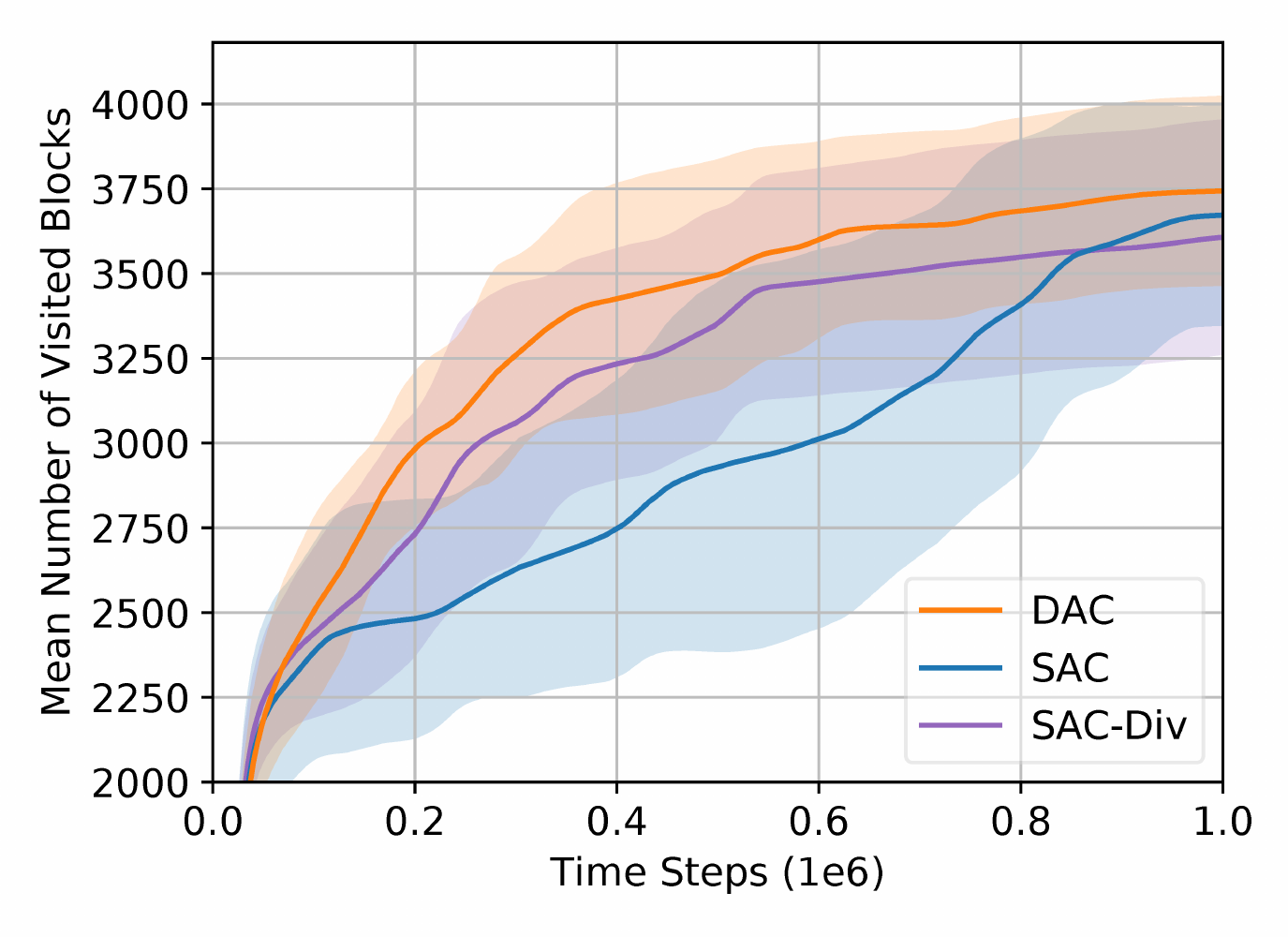}}
	\subfigure[SparseWalker2d-v1]{\includegraphics[width=0.31\textwidth]{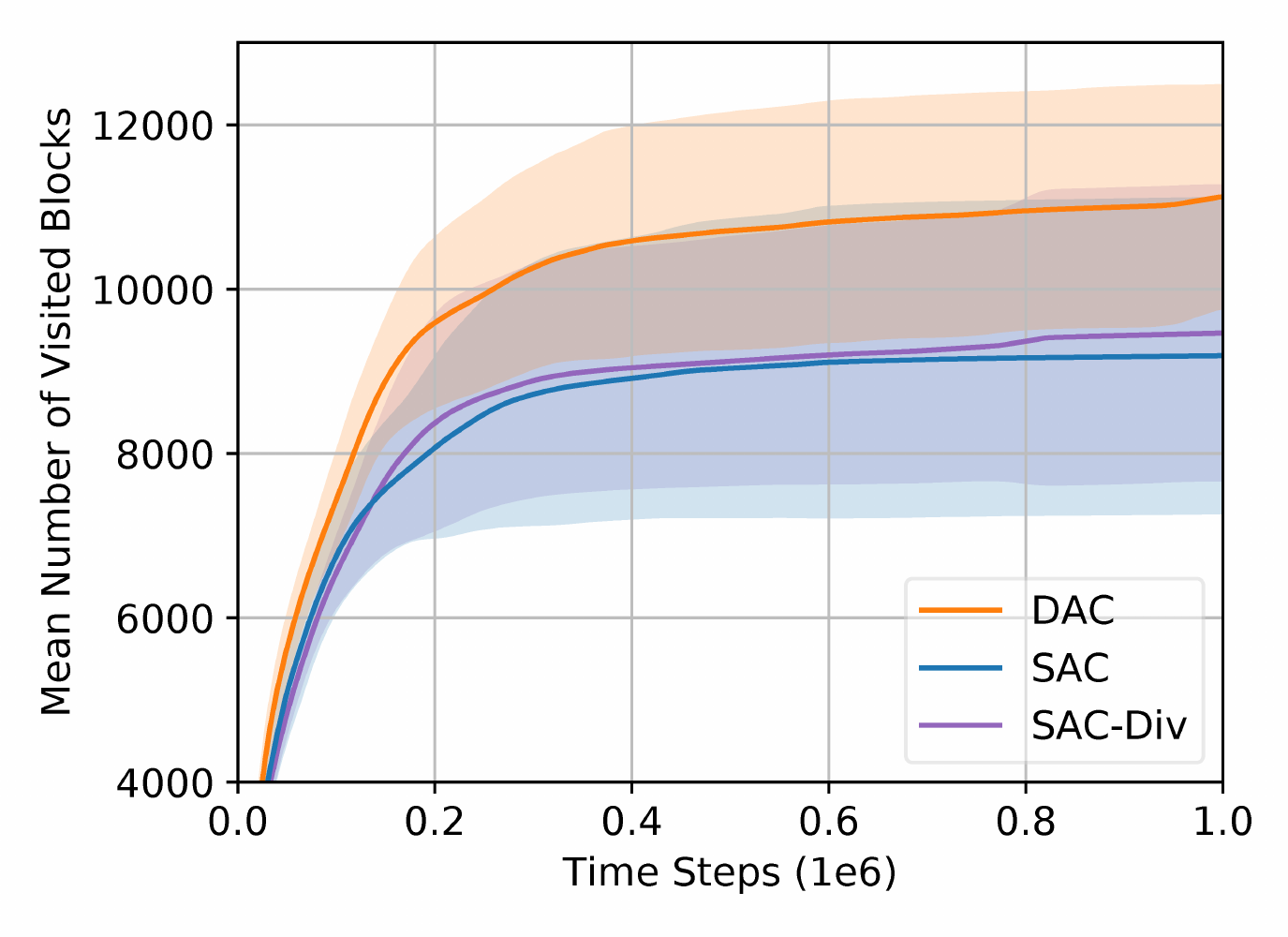}}
	\subfigure[SparseAnt-v1]{\includegraphics[width=0.31\textwidth]{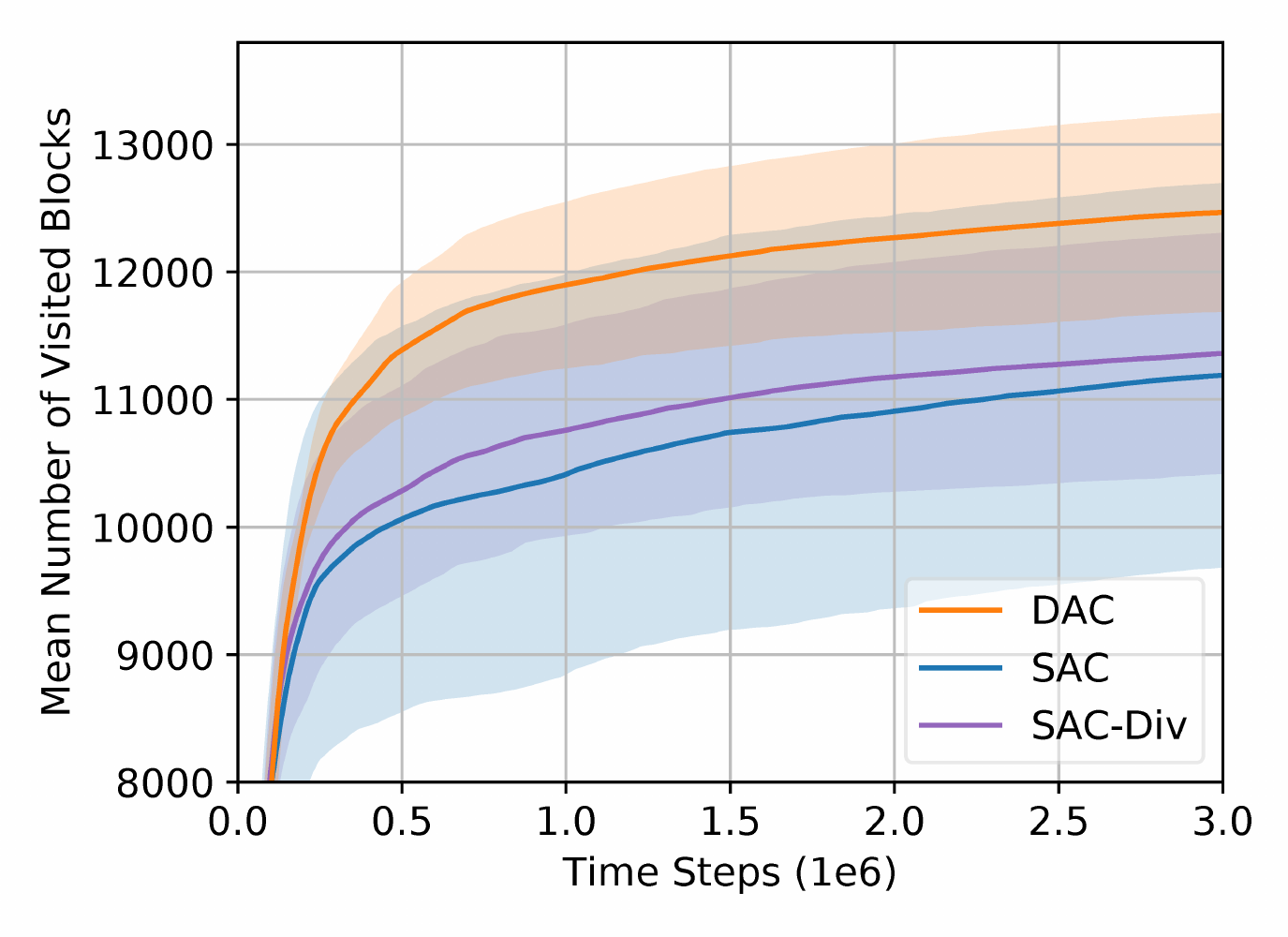}}
	\caption{The number of discretized state visitation on sparse Mujoco tasks}
	\label{fig:compcovmujoco}
\end{figure}

\begin{figure*}[!t]
	\centering
	\subfigure[HumanoidStandup-v1]{\includegraphics[width=0.33\textwidth]{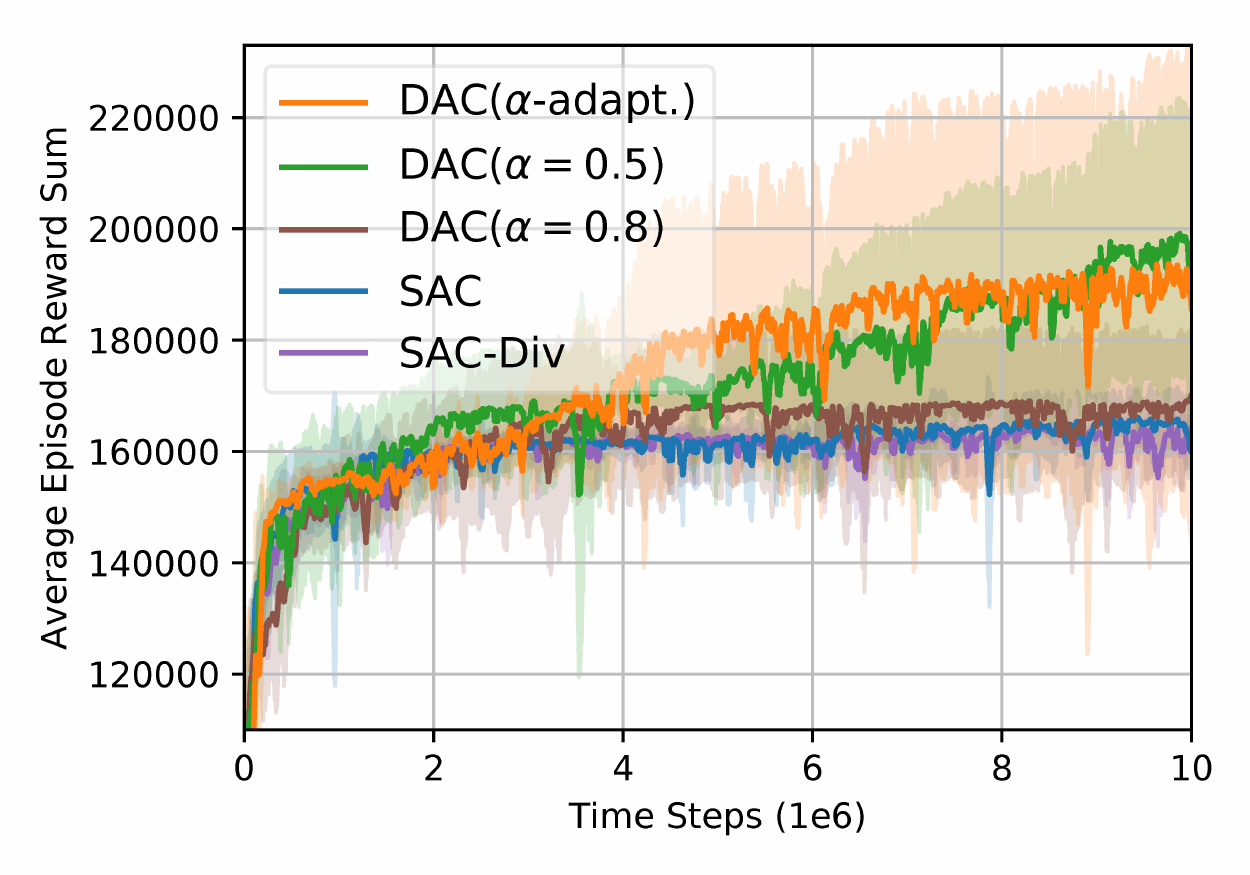}}
	\subfigure[Del.HalfCheetah-v1]{\includegraphics[width=0.33\textwidth]{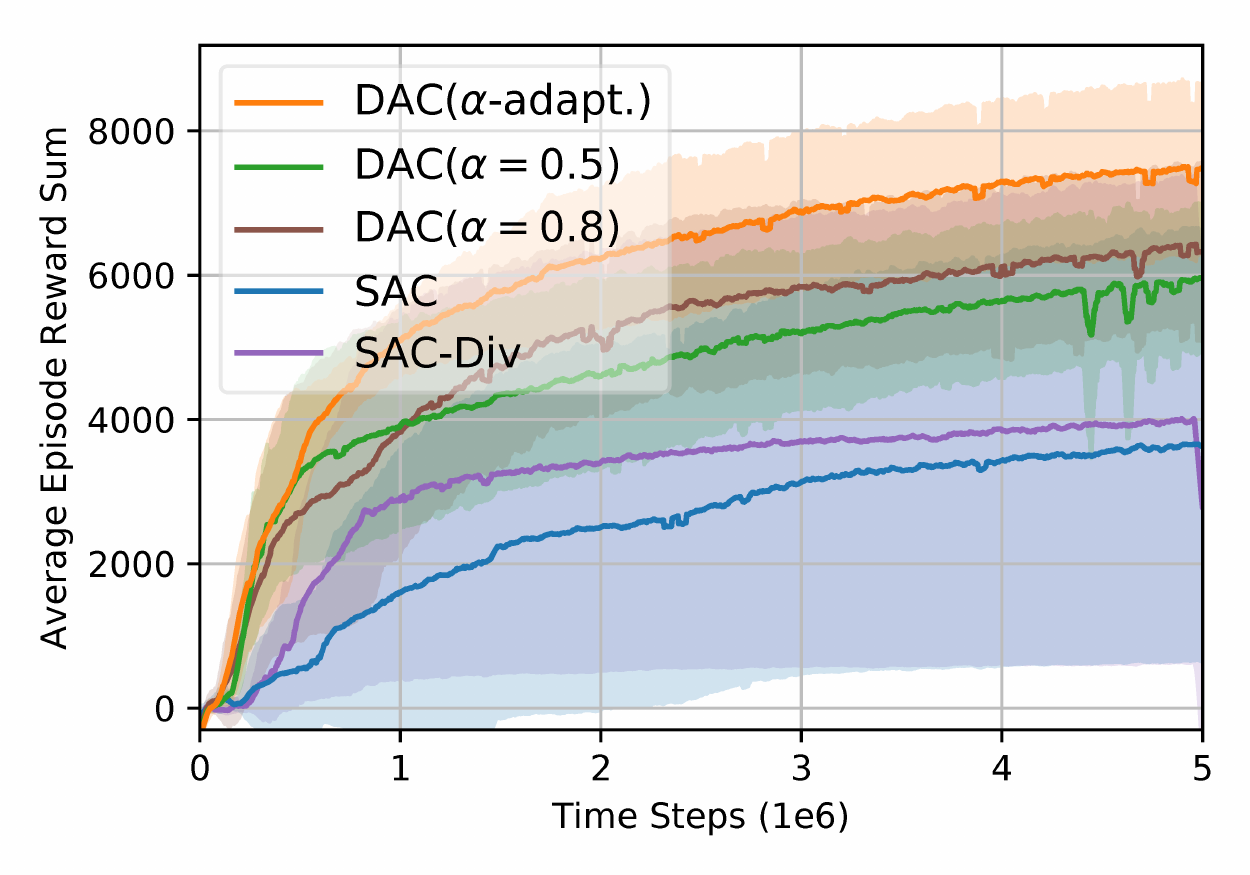}}
	\subfigure[Del.Hopper-v1]{\includegraphics[width=0.33\textwidth]{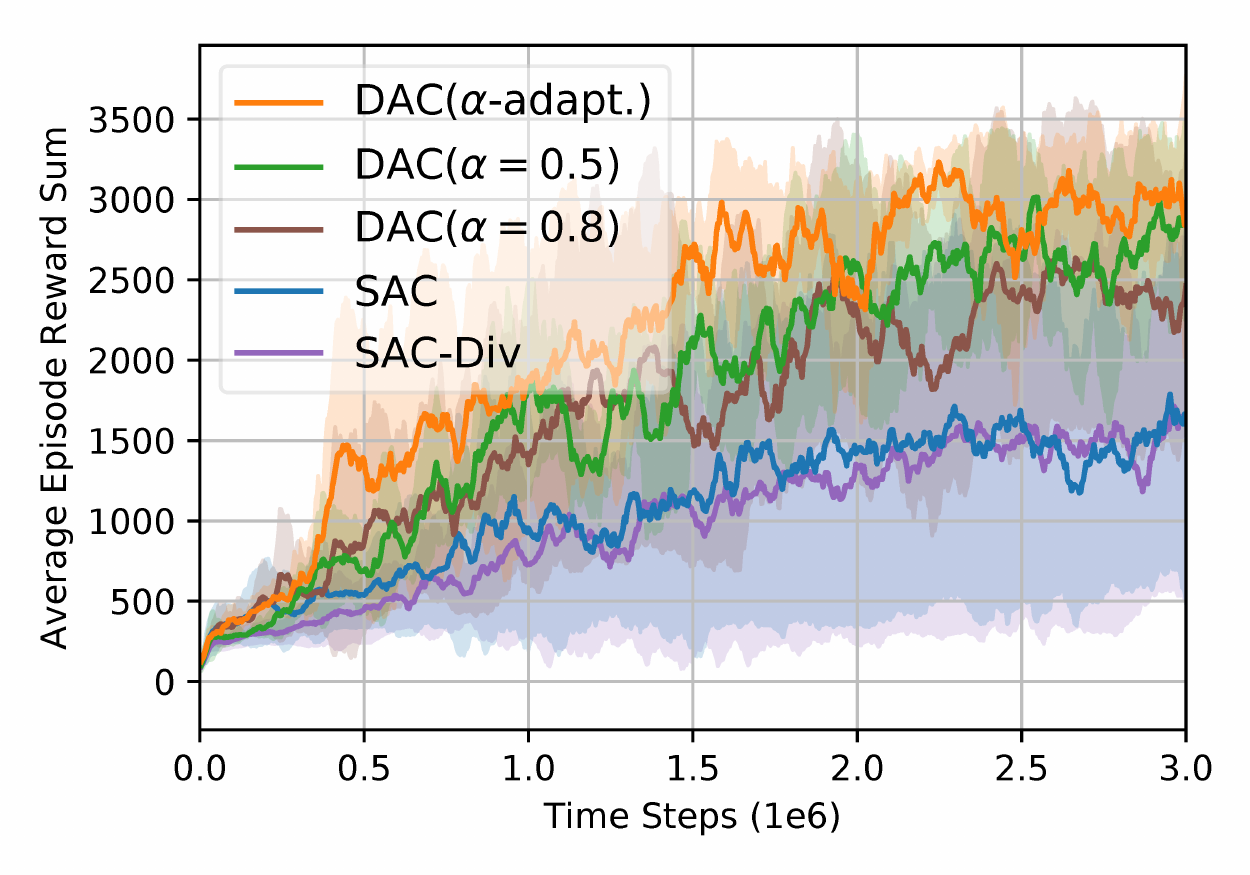}}
	\subfigure[Del.Walker2d-v1]{\includegraphics[width=0.33\textwidth]{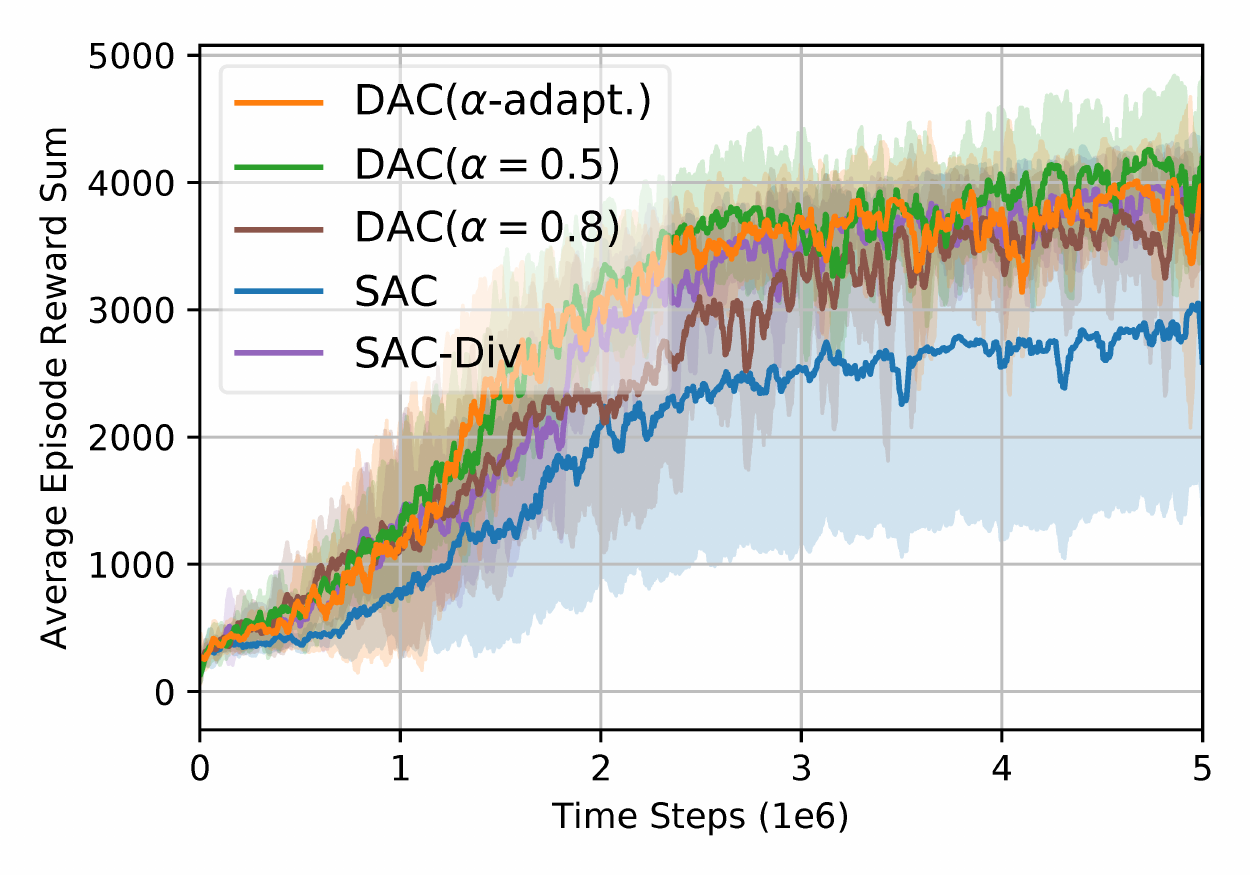}}
	\subfigure[Del.Ant-v1]{\includegraphics[width=0.33\textwidth]{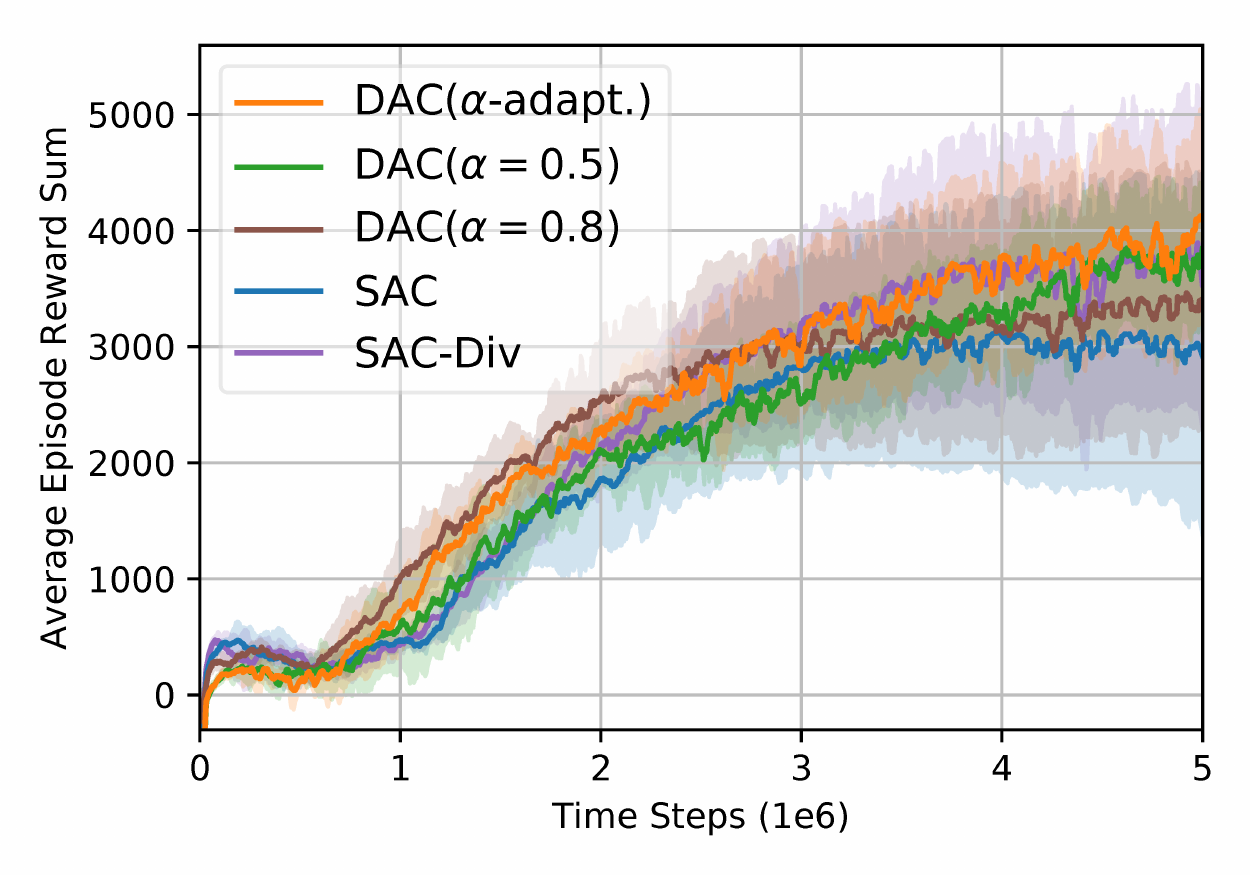}}
	\caption{Performance comparison on HumanoidStandup and DelayedMujoco tasks}
	\label{fig:compmujoco2}
	\vspace{3em}
\end{figure*}

In this subsection, we provide more performance plots and tables for the performance comparison between DAC and SAC baselines. Fig. \ref{fig:compdivmujoco} shows the divergence $D_{JS}^{\alpha}$ curve ($\alpha=0.5$) and Fig. \ref{fig:compcovmujoco} shows the mean number of discretized state visitation curve for remaining SparseMujoco tasks. Table. \ref{table:marsparse} shows the corresponding max average return performance on sparse Mujoco tasks. Fig. \ref{fig:compmujoco2} shows the scaled version of the performance plots in Fig. \ref{fig:compcovmujoco}, and Table \ref{table:marbase} shows the corresponding max average return performance.

Here, in order to show the tendency of state visitation in Fig. \ref{fig:compcovmujoco}, we discretized the state of each SparseMujoco task. For discretization, we simply consider 2 components of observations for each task: $x,y$ axis position for SparseAnt, and $x,z$ axis position for the other SparseMujoco tasks. We discretize the position by setting the grid spacing per axis to $0.01$ in the range of $(-10,10)$. For SAC/SAC-Div, the ratio function $R$ is estimated separately by the same way with DAC.

\vspace{2em}

\newpage

\begin{table*}[!t]
	\centering
	\begin{adjustbox}{width=0.8\textwidth}
		\begin{tabular}{L{8em}|C{8em}C{8em}C{8em}}
			\hline
			& DAC ($\alpha=0.5$) & SAC & SAC-Div\\
			\hline
			SparseHalfCheetah & {\bf915.90$\pm$50.71} & 386.90$\pm$404.70 & 394.70$\pm$405.53\\
			SparseHopper & {\bf900.30$\pm$3.93} & 823.70$\pm$215.35 & 817.40$\pm$253.54\\
			SparseWalker2d & {\bf665.10$\pm$355.66} & 273.30$\pm$417.51 & 278.50$\pm$398.23\\
			SparseAnt & 935.80$\pm$37.08 & {\bf963.80$\pm$42.51} & 870.70$\pm$121.14\\
			\hline
		\end{tabular}
	\end{adjustbox}
	\caption{Max average return of DAC algorithm and SAC baselines on SparseMujoco tasks}
	\label{table:marsparse}
	\vspace{3em}
\end{table*}

\begin{table*}[!t]
	\centering
	\begin{adjustbox}{width=1\textwidth}
		\begin{tabular}{L{7.1em}|C{7.5em}C{7.5em}C{7.5em}C{7.5em}C{7.5em}}
			\hline
			& DAC ($\alpha=0.5$) & DAC ($\alpha=0.8$) & DAC ($\alpha$-adapt.) & SAC & SAC-Div\\
			\hline
			HumanoidS & {\bf202491.81 $\pm$25222.77} & 170832.05 $\pm$12344.71 & 197302.37 $\pm$43055.31 & 167394.36 $\pm$7291.99 & 165548.76 $\pm$2005.85 \\
			Del. HalfCheetah & 6071.93$\pm$1045.64 & 6552.06$\pm$1140.18 & {\bf7594.70$\pm$1259.23} & 3742.33$\pm$3064.55 & 4080.67$\pm$3418.07 \\
			Del. Hopper & 3283.77$\pm$112.04 & 2836.81$\pm$679.05 & {\bf3428.18$\pm$69.08} & 2175.31$\pm$1358.39 & 2090.64$\pm$1383.83\\
			Del. Walker2d & {\bf4360.43$\pm$507.58} & 3973.37$\pm$273.63 & 4067.11$\pm$257.81 & 3220.92$\pm$1107.91 & 4048.11$\pm$290.48 \\
			Del. Ant & 4088.12$\pm$578.99 & 3535.72$\pm$1164.76 & {\bf4243.19$\pm$795.49} & 3248.43$\pm$1454.48 & 3978.34$\pm$1370.23 \\
			\hline
		\end{tabular}
	\end{adjustbox}
	\caption{Max average return of DAC algorithms and SAC baselines on HumanoidStandup and DelayedMujoco tasks}
	\label{table:marbase}
	\vspace{3em}
\end{table*}

\newpage
\subsection{Comparison to State-based Exploration Methods on Sparse Mujoco Tasks}
\label{subsec:rnd}


We compared the performance of DAC ($\alpha=0.5$) with RND/MaxEnt(State) on SparseMujoco tasks, and the performance  of DAC ($\alpha$-adapt.) with RND/MaxEnt(State) on DelayedMujoco tasks. Fig. \ref{fig:otherrlrnd} shows the performance learning curve, and the corresponding max average return table in Table \ref{table:marsparsernd}. From the results, it is seen that DAC  has better performance than RND/MaxEnt(State) on most Sparse/DelayedMujoco tasks.  DAC has superiority not only in pure exploration but also in learning sparse rewarded tasks as compared to recent state-based exploration methods.

\vspace{1em}
\begin{figure}[!h]
	\centering
	\subfigure[SparseHalfCheetah-v1]{\includegraphics[width=0.45\textwidth]{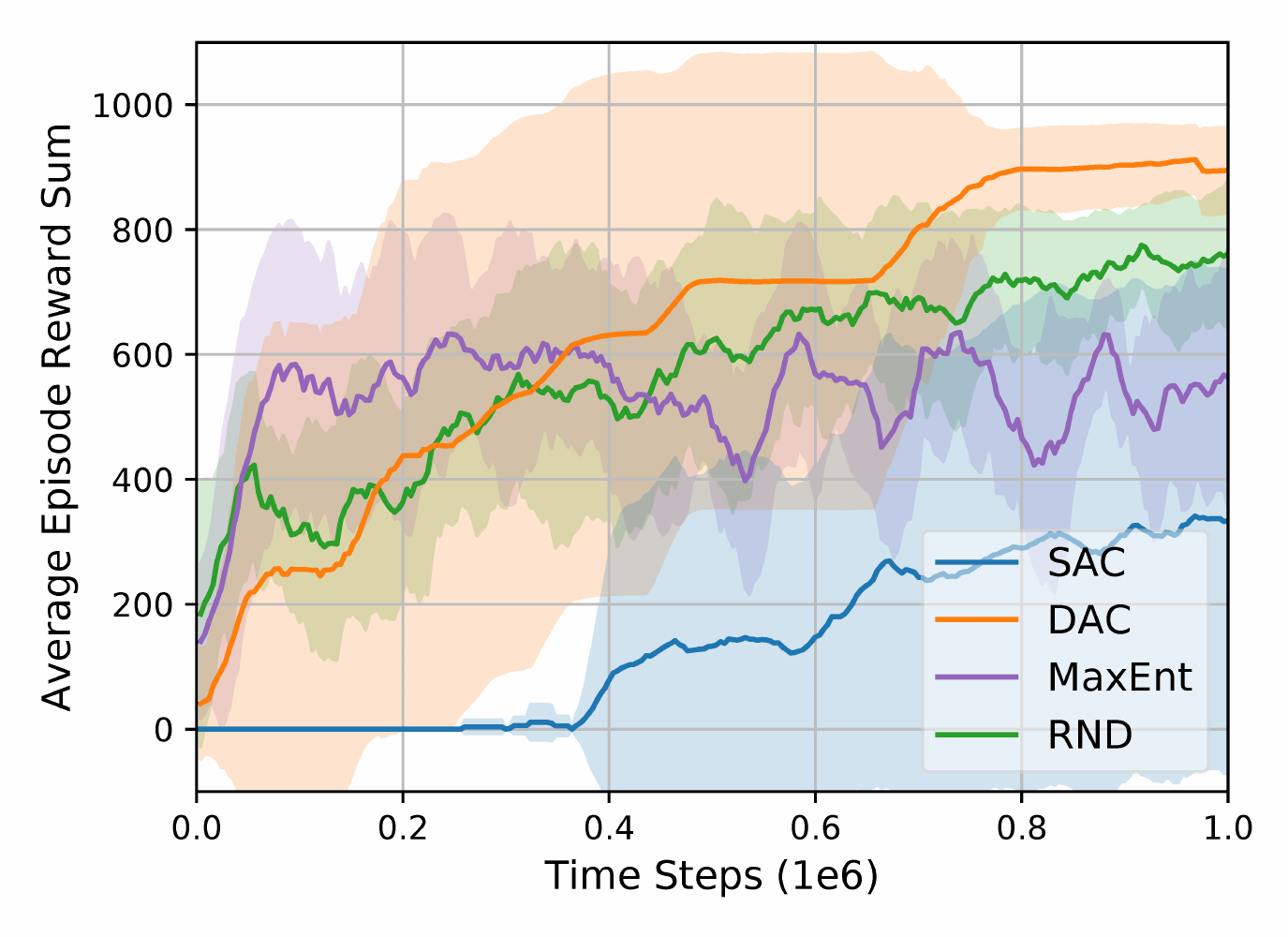}}
	\subfigure[SparseHopper-v1]{\includegraphics[width=0.45\textwidth]{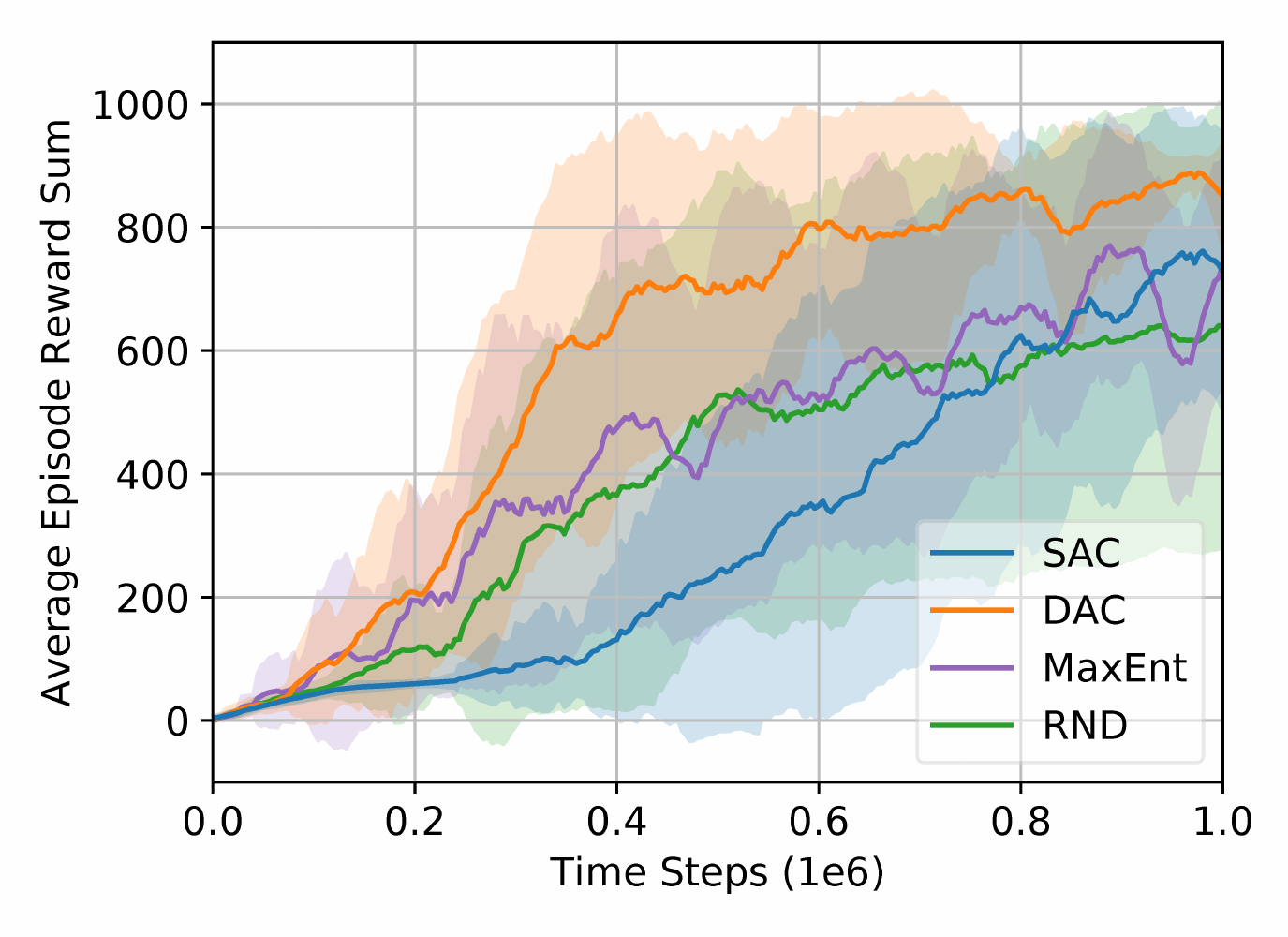}}
	\subfigure[SparseWalker2d-v1]{\includegraphics[width=0.45\textwidth]{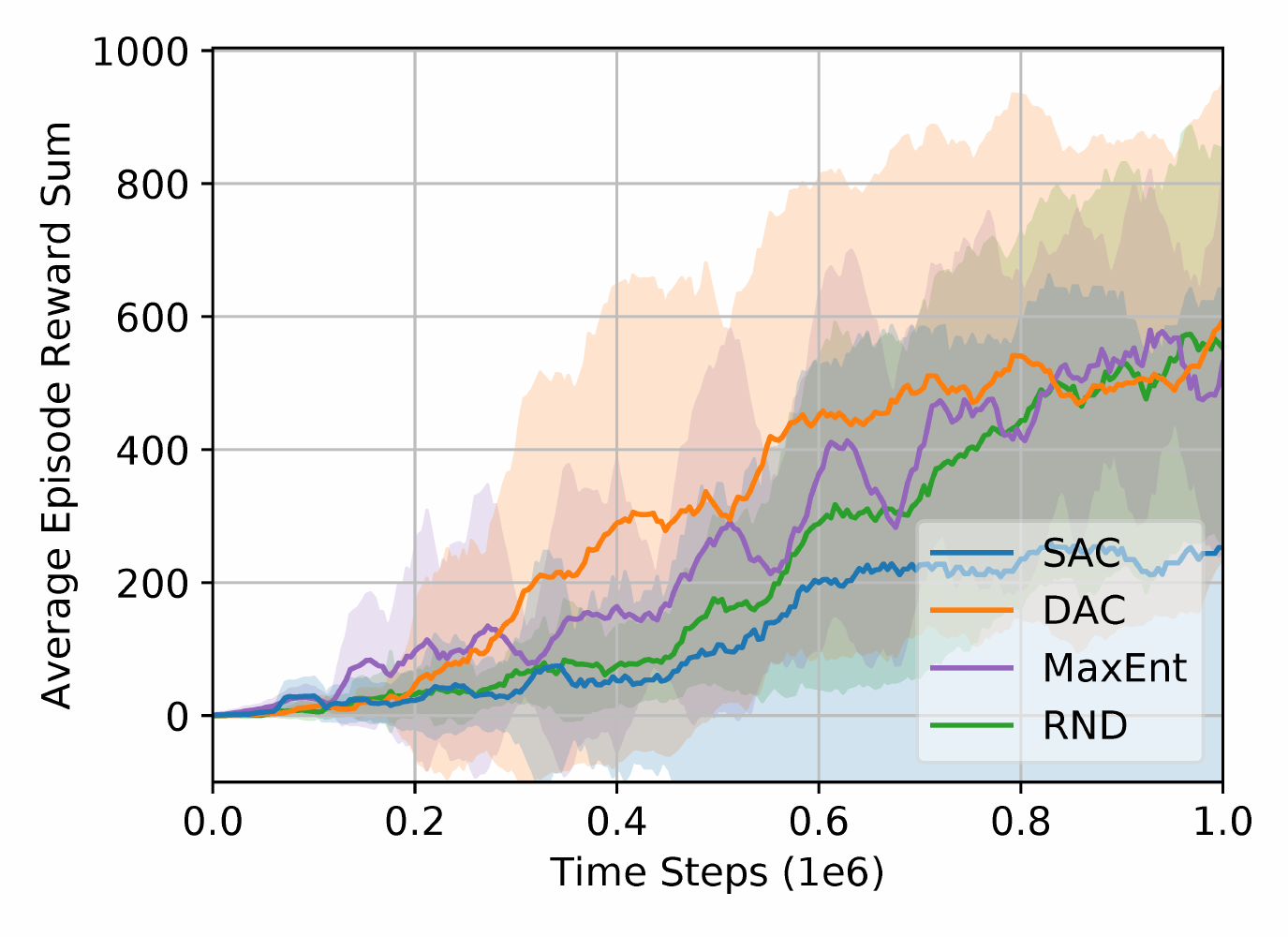}}
	\subfigure[SparseAnt-v1]{\includegraphics[width=0.45\textwidth]{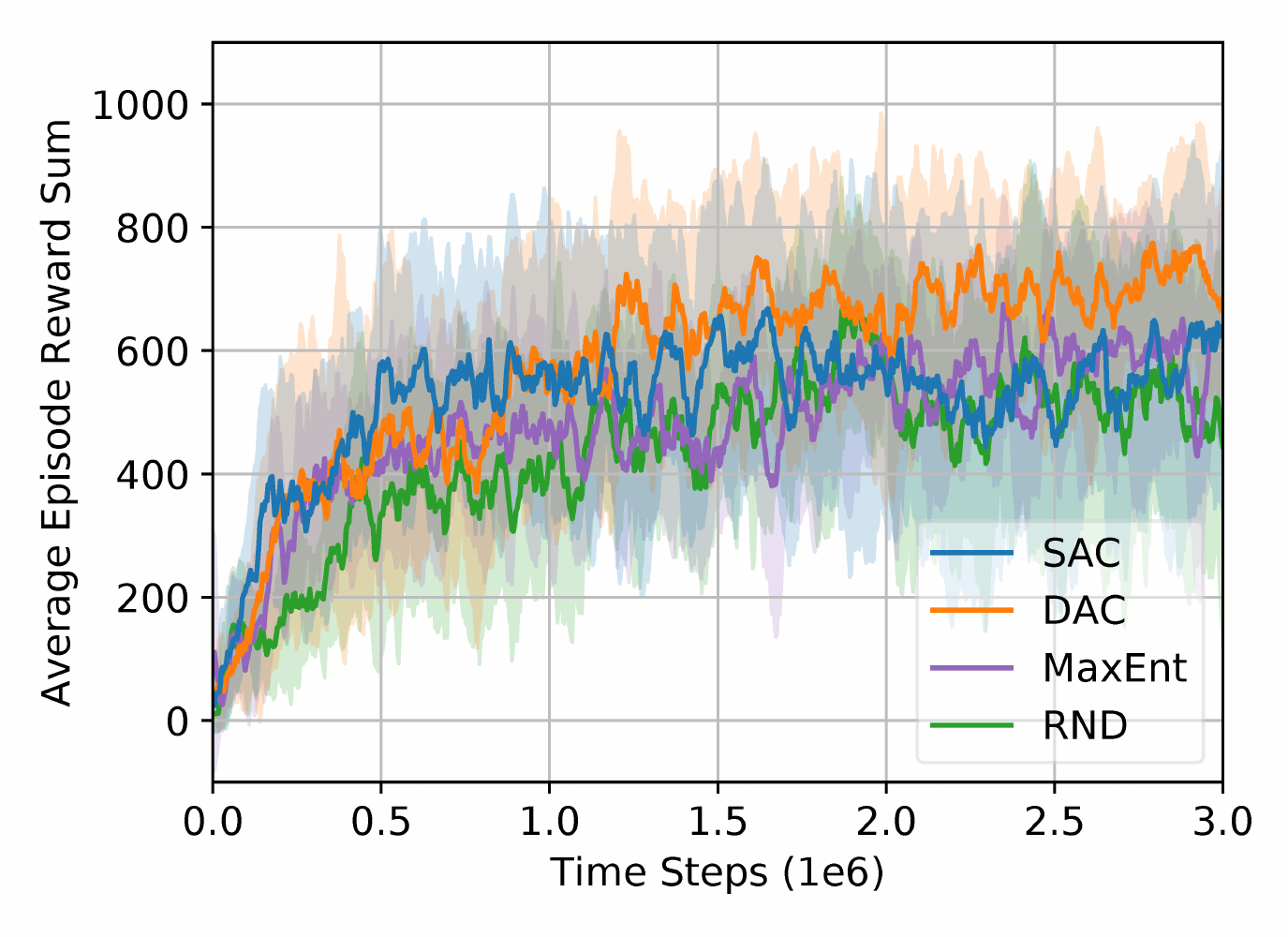}}
	\caption{Performance comparison to RND/MaxEnt(State) on SparseMujoco tasks}
\end{figure}
\begin{figure}[!h]
    \centering
	\subfigure[Del.HalfCheetah-v1]{\includegraphics[width=0.45\textwidth]{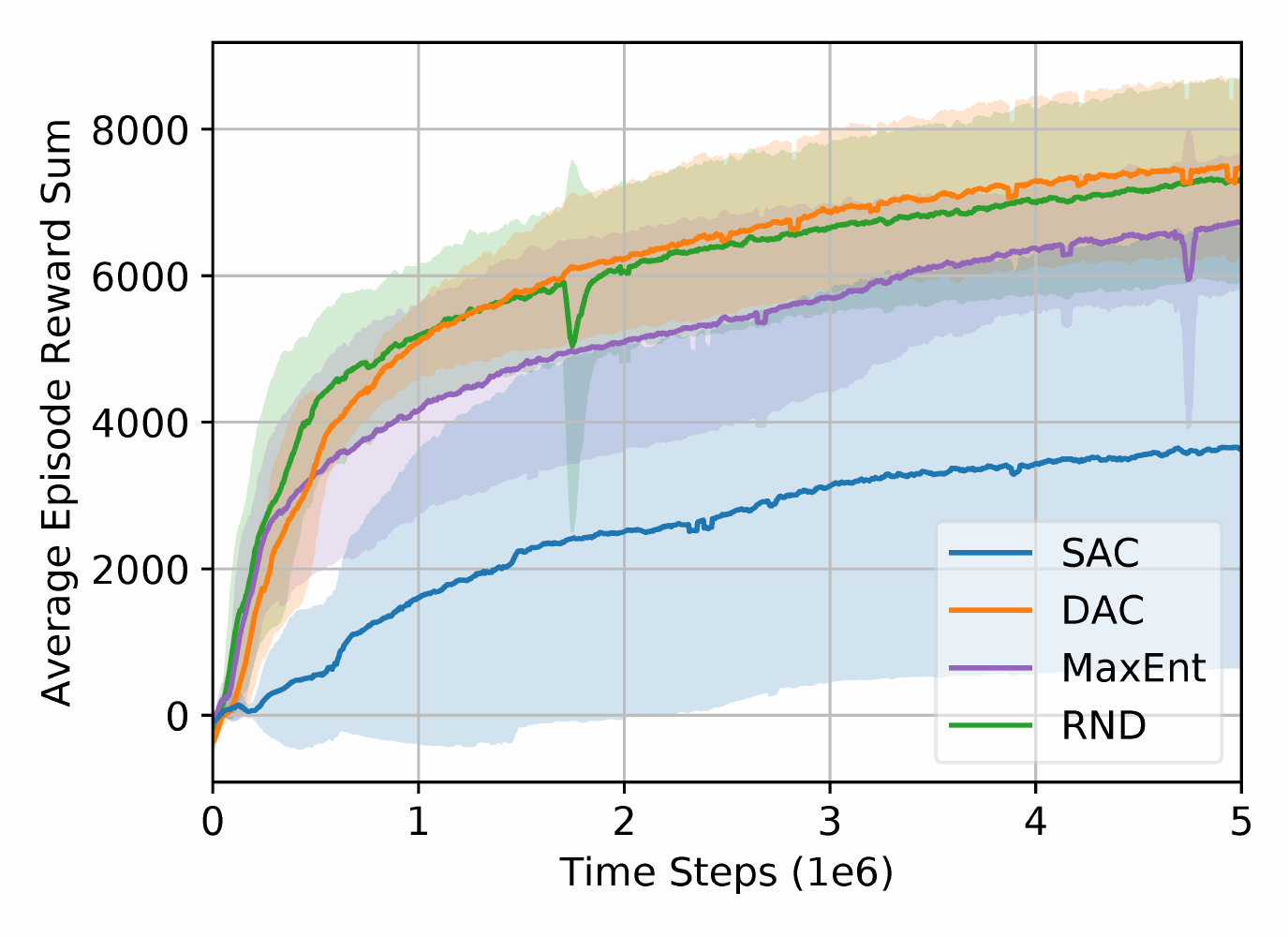}}
	\subfigure[Del.Hopper-v1]{\includegraphics[width=0.45\textwidth]{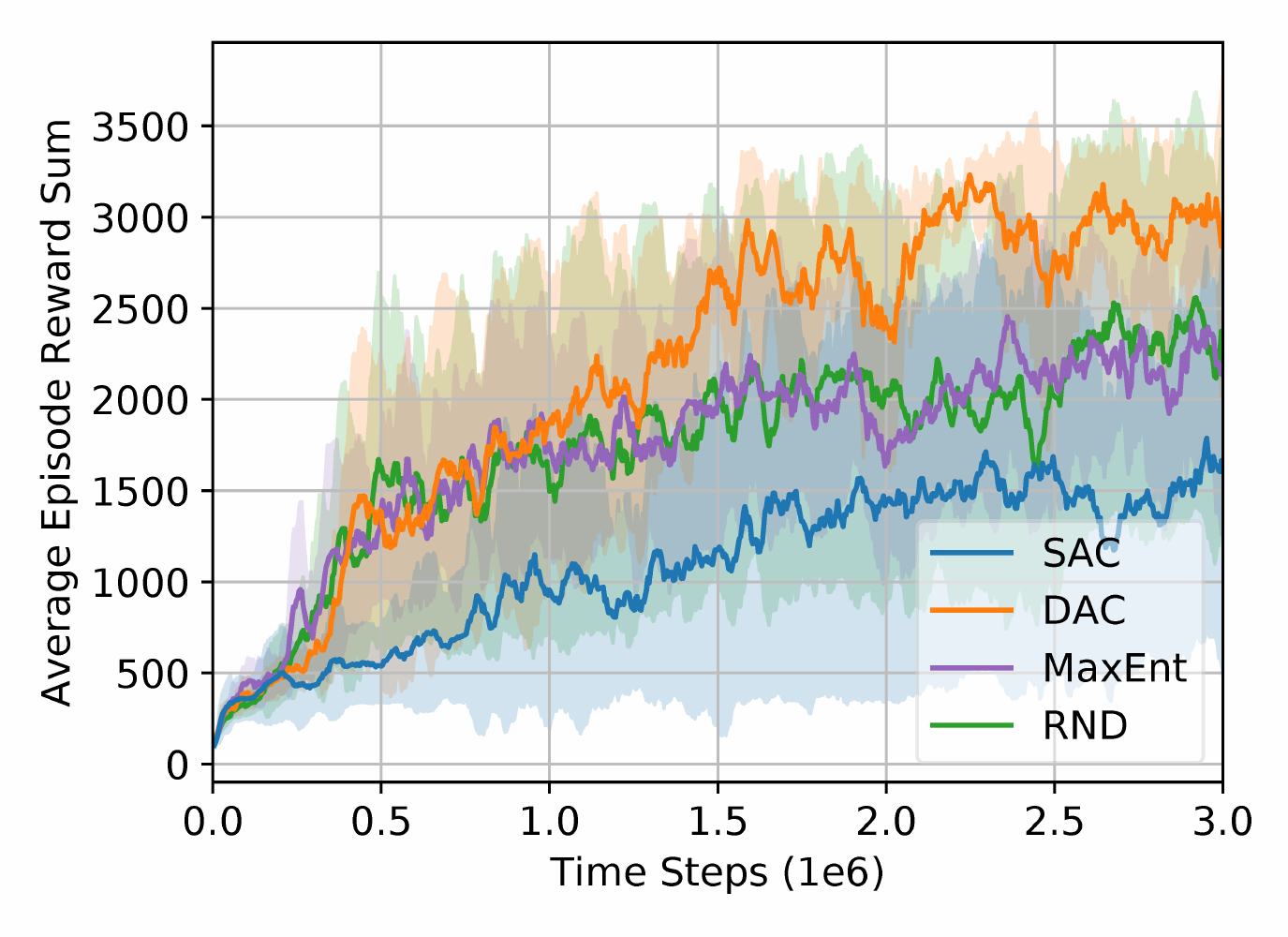}}
	\subfigure[Del.Walker2d-v1]{\includegraphics[width=0.45\textwidth]{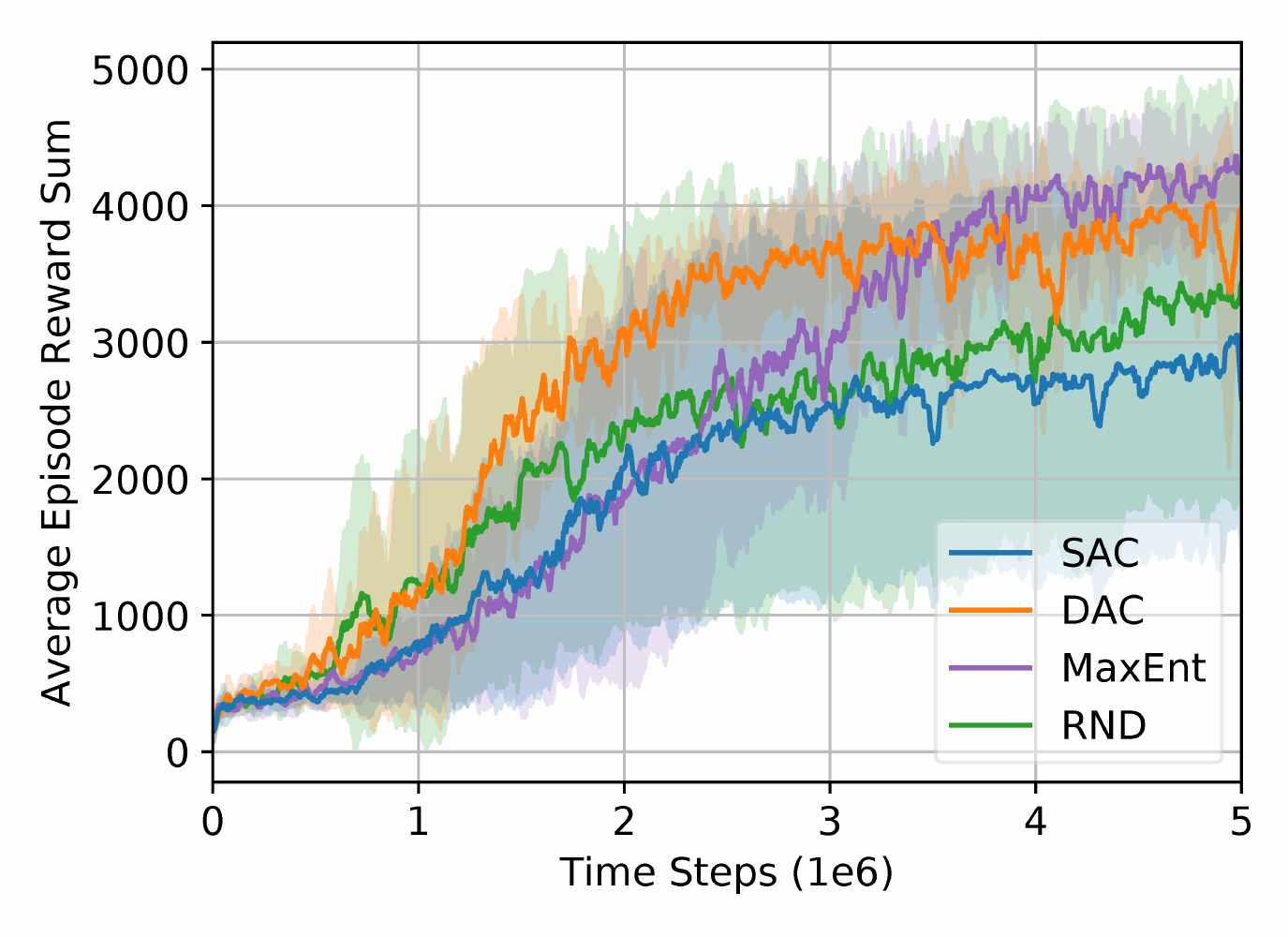}}
	\subfigure[Del.Ant-v1]{\includegraphics[width=0.45\textwidth]{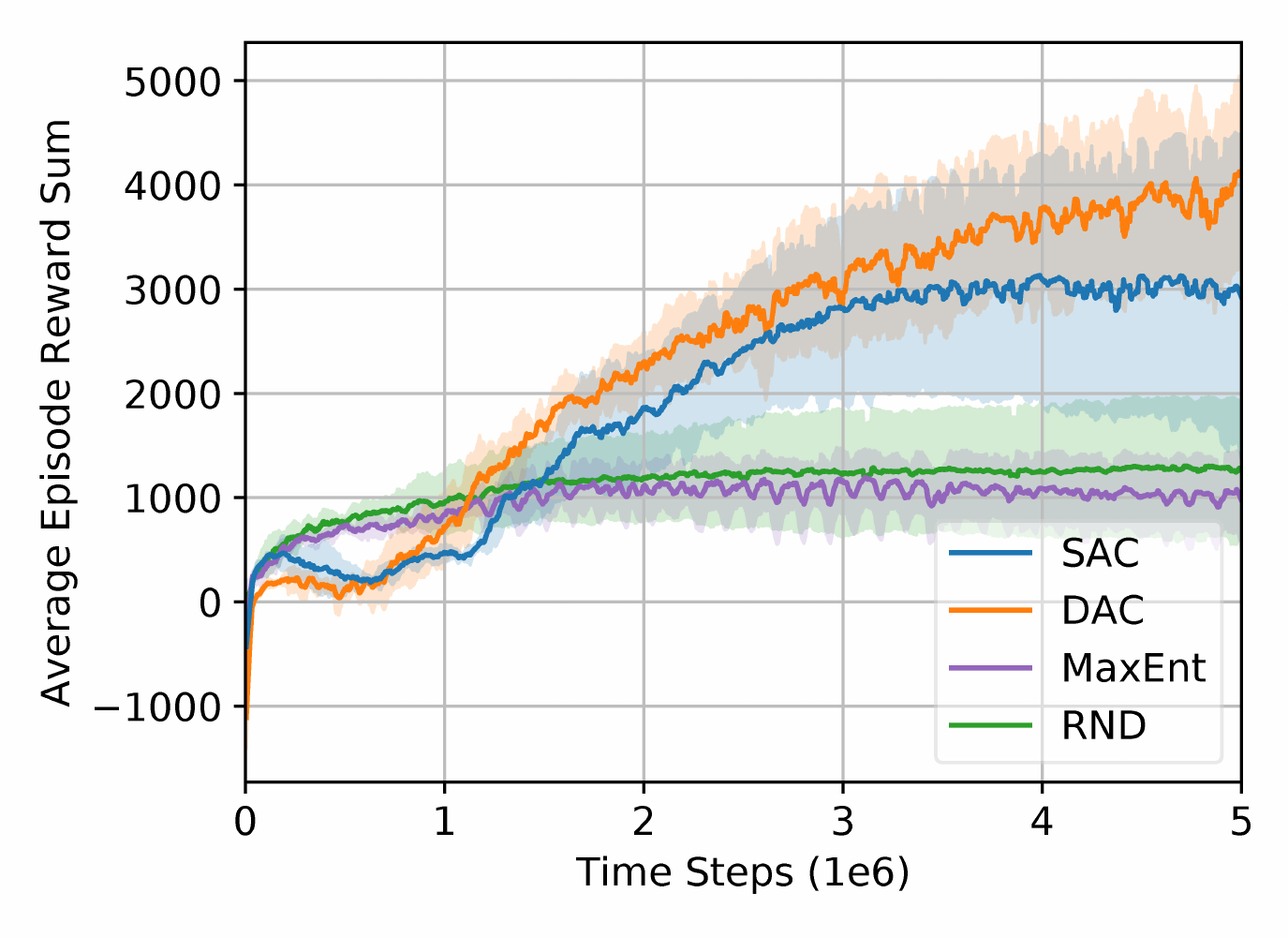}}
	\caption{Performance comparison to RND/MaxEnt(State) on DelayedMujoco tasks}
	\label{fig:otherrlrnd}
	\vspace{-0.8em}
\end{figure}

\begin{table*}[!h]
	\centering
	\begin{adjustbox}{width=0.8\textwidth}
		\begin{tabular}{L{8em}|C{8em}C{8em}C{8em}C{8em}}
			\hline
			& DAC ($\alpha=0.5$) & RND & MaxEnt(State) & SAC \\
			\hline
			SparseHalfCheetah & {\bf915.90$\pm$50.71} & 827.80$\pm$85.61 & 800.20$\pm$127.11 & 386.90$\pm$404.70\\
			SparseHopper & {\bf900.30$\pm$3.93} & 648.10$\pm$363.75 & 879.50$\pm$30.96 & 823.70$\pm$215.35\\
			SparseWalker2d & 665.10$\pm$355.66 & 663.00$\pm$356.39 & {\bf705.30$\pm$274.88} & 273.30$\pm$417.51 \\
			SparseAnt & 935.80$\pm$37.08 & 920.60$\pm$107.50 & 900.00$\pm$70.02 & {\bf963.80$\pm$42.51}\\
			\hline
			& DAC ($\alpha$-adapt.) & RND & MaxEnt(State) & SAC\\
			\hline
			Del.HalfCheetah & {\bf7594.70$\pm$1259.23} & 7429.94$\pm$1383.75 & 6823.37$\pm$882.25 & 3742.33$\pm$3064.55 \\
			Del.Hopper & {\bf3428.18$\pm$69.08} & 2764.06$\pm$1220.86 & 3254.10$\pm$30.75 & 2175.31$\pm$1358.39 \\
			Del.Walker2d & 4067.11$\pm$257.81 & 3514.97$\pm$1536.04 & {\bf4430.61$\pm$347.02} & 3220.92$\pm$1107.91 \\
			Del.Ant & {\bf4243.19$\pm$795.49} & 1361.36$\pm$704.69 & 1246.80$\pm$323.50 & 3248.43$\pm$1454.48 \\
			\hline
		\end{tabular}
	\end{adjustbox}
	\caption{Max average return of DAC, RND, and MaxEnt(State)}
	\label{table:marsparsernd}
\end{table*}

\newpage
\subsection{Comparison to Recent General RL Algorithms}
\label{subsec:appothers}

\begin{figure}[!h]
	\centering
	\subfigure[HumanoidStandup-v1]{\includegraphics[width=0.33\textwidth]{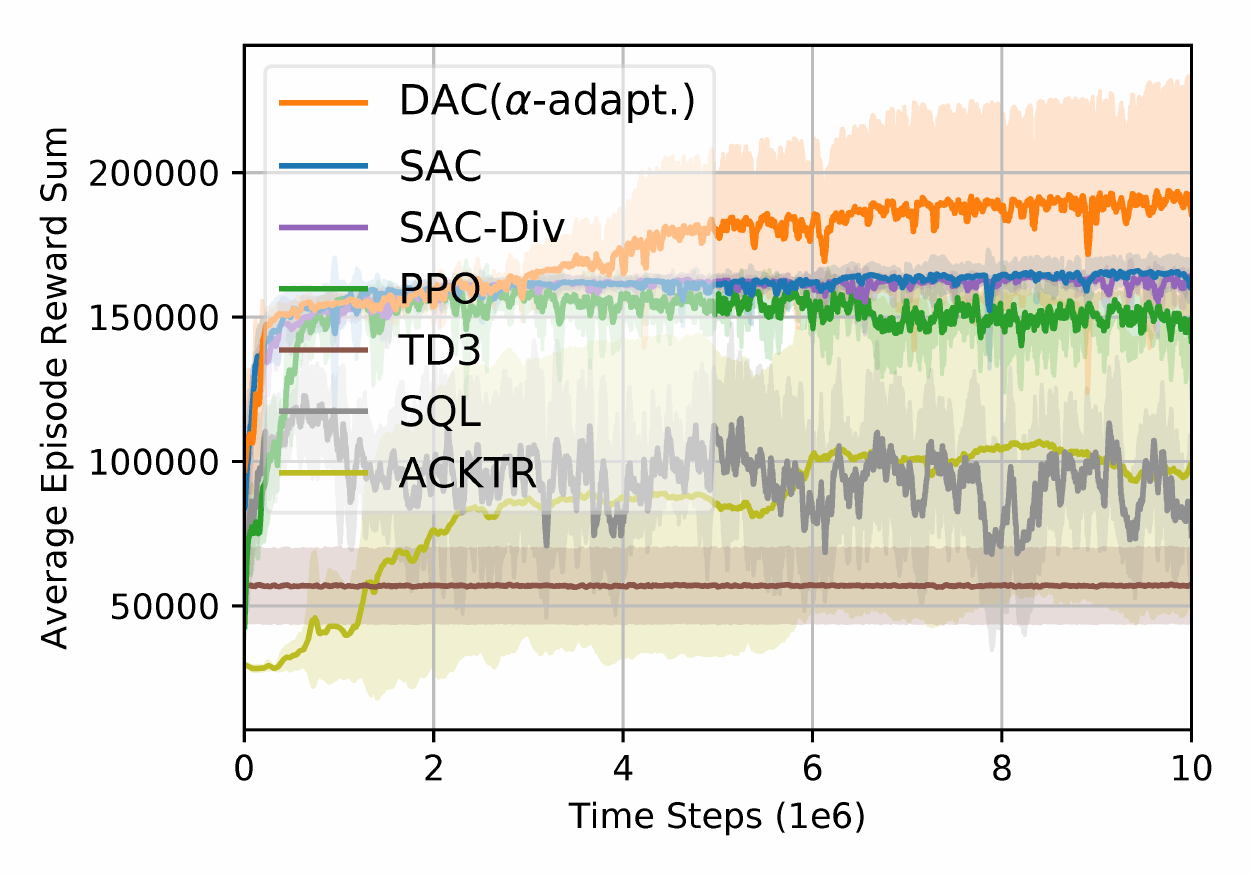}}
	\subfigure[DelayedHalfCheetah-v1]{\includegraphics[width=0.33\textwidth]{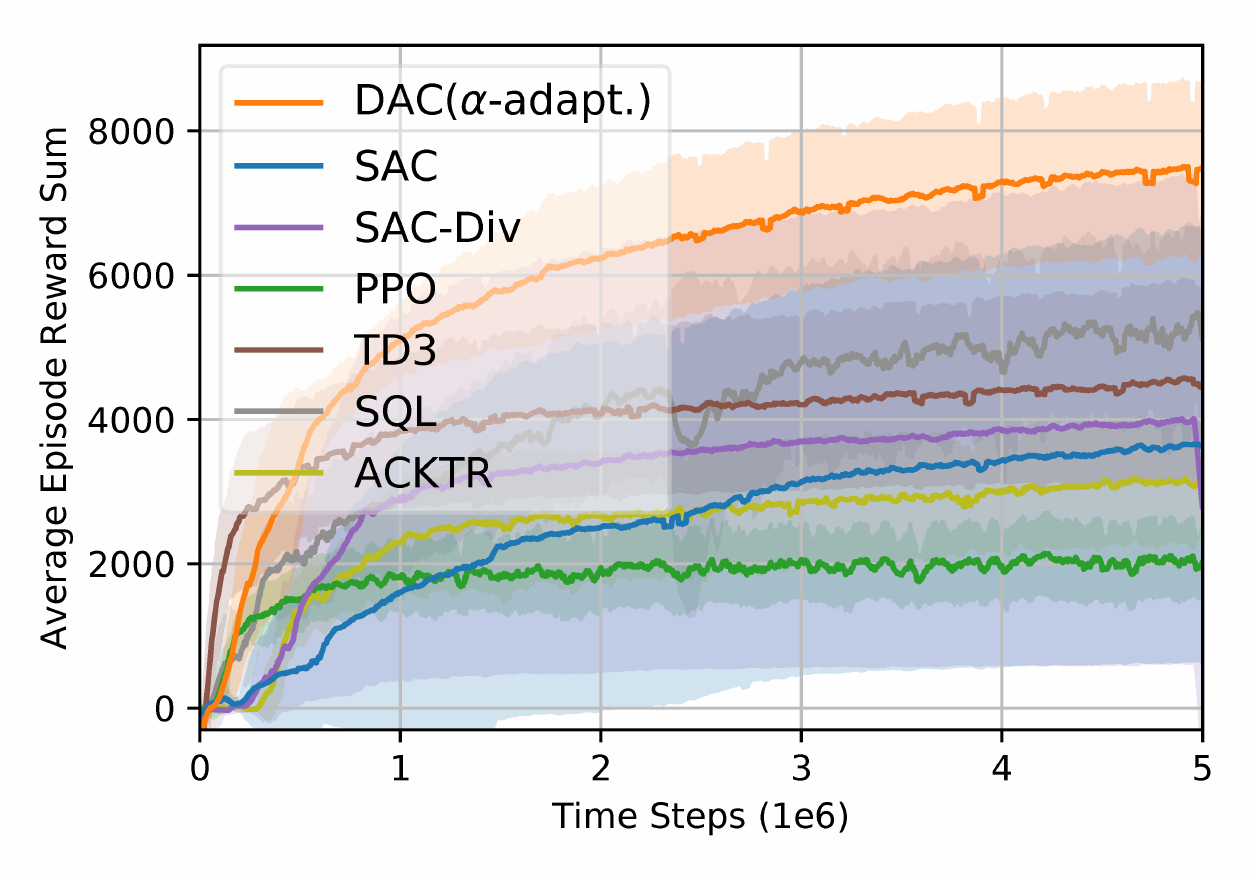}}
	\subfigure[DelayedHopper-v1]{\includegraphics[width=0.33\textwidth]{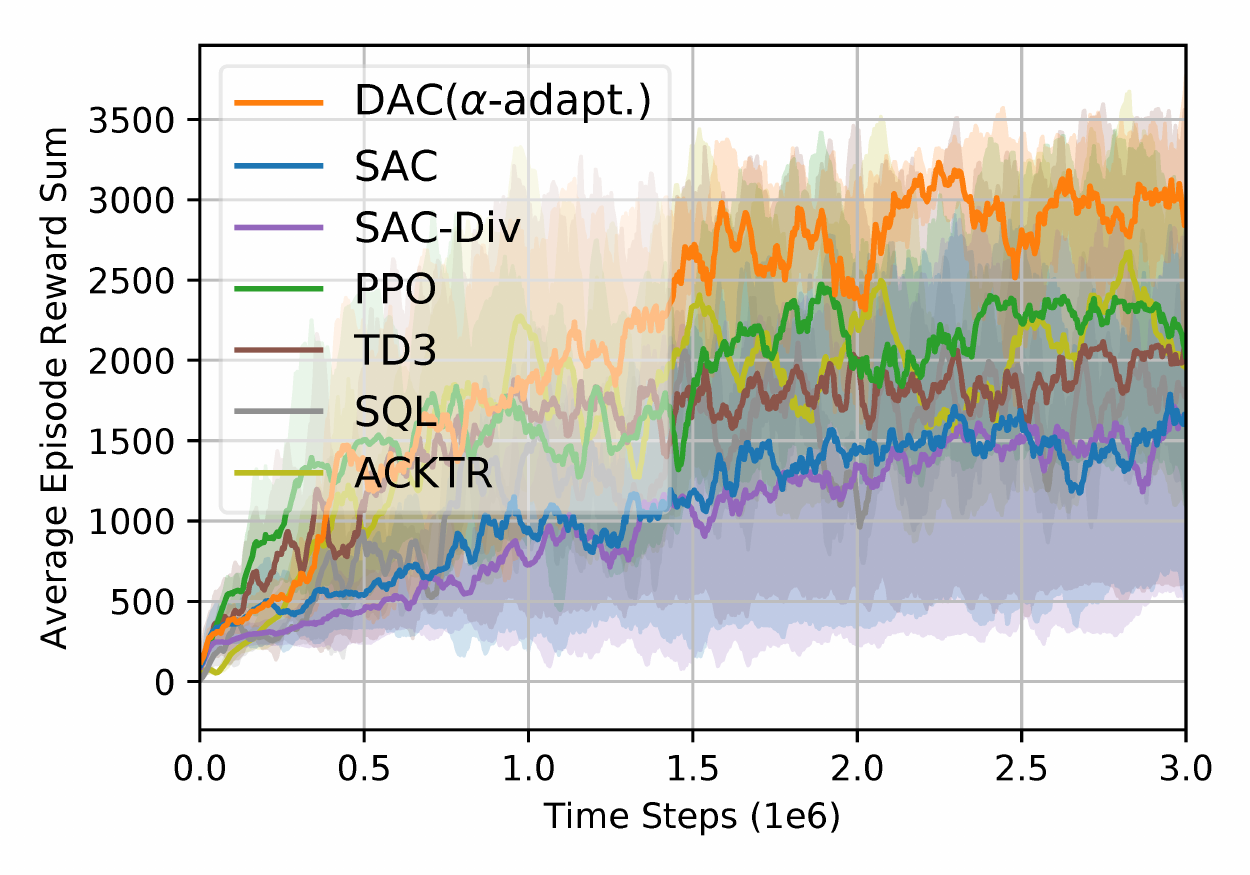}}
	\subfigure[DelayedWalker2d-v1]{\includegraphics[width=0.33\textwidth]{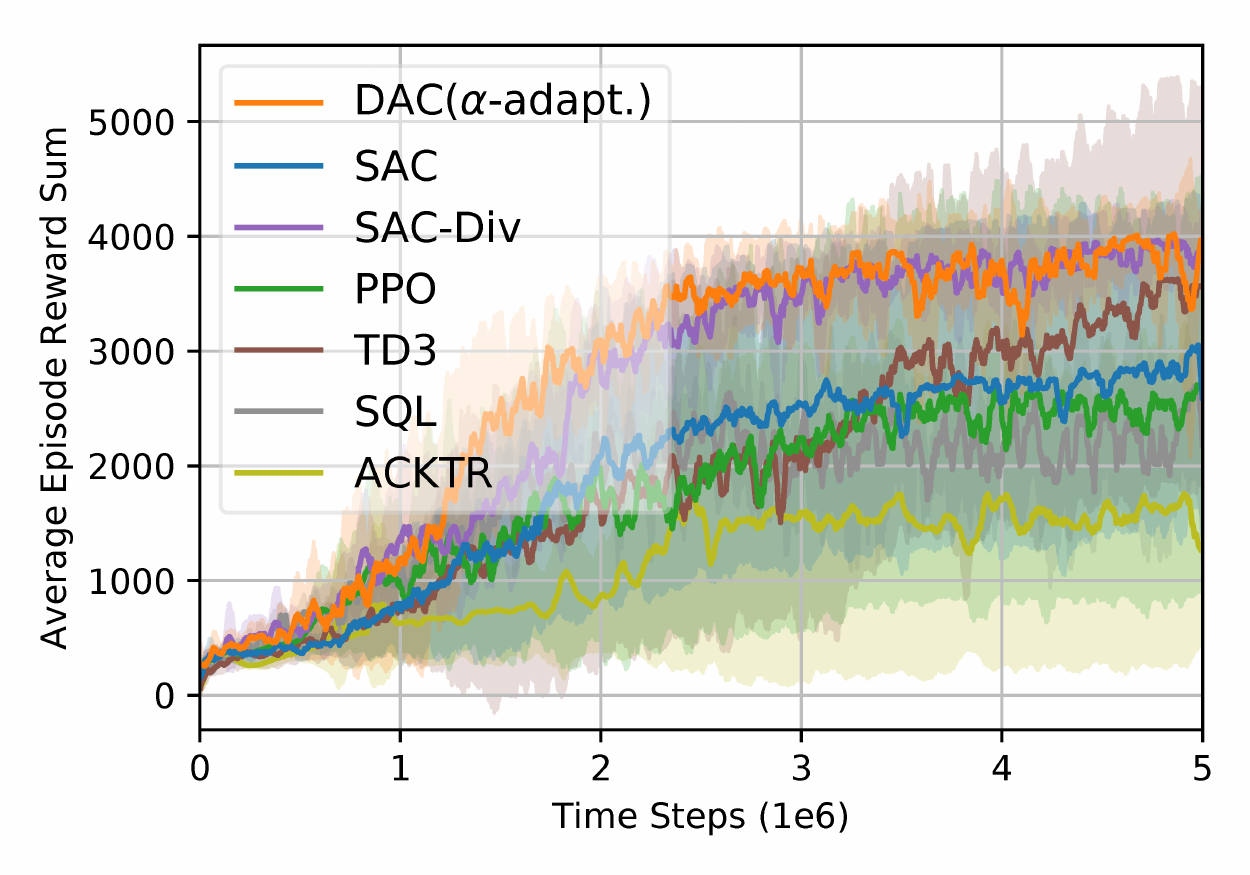}}
	\subfigure[DelayedAnt-v1]{\includegraphics[width=0.33\textwidth]{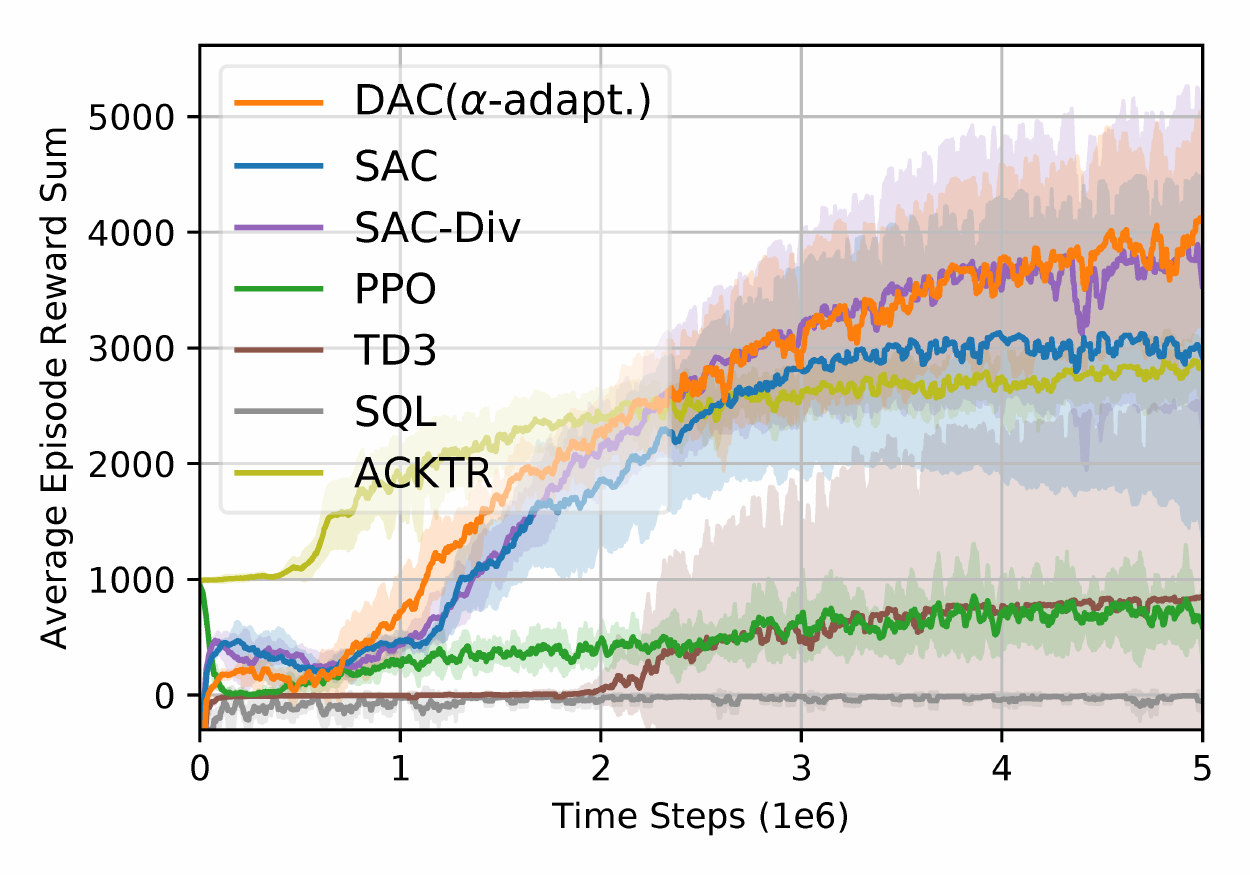}}
	\caption{Performance comparison to recent general RL algorithms}
	\label{fig:otherrl}
\end{figure}

We also compare the performance of DAC with $\alpha$-adaptation to other state-of-the-art RL algorithms. Here, we consider various on-policy RL algorithms: Proximal Policy Optimization \citep{schulman2017proximal} (PPO, a stable and popular on-policy algorithm), Actor Critic using Kronecker-factored Trust Region \citep{wu2017scalable} (ACKTR, actor-critic that approximates natural gradient by using Kronecker-factored curvature),  and off-policy RL algorithms: Twin Delayed Deep Deterministic Policy Gradient \citep{fujimoto2018addressing} (TD3, using clipped double-Q learning for reducing overestimation); and Soft Q-Learning \citep{haarnoja2017rein} (SQL, energy based policy optimization using Stein variational gradient descent). We used implementations in OpenAI baselines \citep{baselines} for PPO and ACKTR, and implementations in author's Github for other algorithms. We provide the performance results as Fig. \ref{fig:otherrl} and Table \ref{table:otherrl}, and the results show that DAC has the best performance on all considered tasks among the compared recent RL algorithms.

\vspace{2em}

\begin{table}[!h]
	\centering
	\begin{adjustbox}{width=1\textwidth}
		\begin{tabular}{L{7.1em}|C{5.1em}C{5.1em}C{5.1em}C{5.1em}C{5.1em}C{5.1em}}
			\hline
			& DAC & PPO & ACKTR & SQL & TD3 & SAC \\
			\hline
			HumanoidS & {\bf197302.37\quad$\pm$43055.31} & 160211.90\quad$\pm$3268.37 & 109655.30\quad$\pm$49166.15 & 138996.84\quad$\pm$33903.03 & 58693.87\quad$\pm$12269.93 & 167394.36\quad$\pm$7291.99 \\
			\hline
			Del. HalfCheetah & {\bf7594.70\quad$\pm$1259.23} & 2247.92\quad$\pm$640.69 & 3295.30\quad$\pm$824.05 & 5673.34\quad$\pm$1241.30 & 4639.85\quad$\pm$1393.95 & 3742.33\quad$\pm$3064.55 \\
			\hline
			Del. Hopper & {\bf3428.18\quad$\pm$69.08} & 2740.15\quad$\pm$719.63 & 2864.81\quad$\pm$1072.64 & 2720.32\quad$\pm$127.71 & 2276.58\quad$\pm$1471.66 & 2175.31\quad$\pm$1358.39 \\
			\hline
			Del. Walker2d & {\bf4067.11\quad$\pm$257.81} & 2859.27\quad$\pm$1938.50 & 1927.32\quad$\pm$1647.49 & 3323.63\quad$\pm$503.18 & 3736.72\quad$\pm$1806.37 & 3220.92\quad$\pm$1107.91 \\
			\hline
			Del. Ant & {\bf4243.19\quad$\pm$795.49} & 1224.33\quad$\pm$521.62 & 2956.51\quad$\pm$234.89 & 6.59\quad$\pm$16.42 & 904.99\quad$\pm$1811.78 & 3248.43\quad$\pm$1454.48 \\
			\hline
		\end{tabular}
	\end{adjustbox}
	\caption{Max average return of DAC and other RL algorithms}
	\label{table:otherrl}
\end{table}

\newpage
\section{More Ablation Studies}
\label{sec:mainablation}

\vspace{1em}

In this section, we provide detailed ablation studies on the DelayedMucoco tasks. First, we focus on the DelayedHalfCheetah task because the tendencies of performance changes are similar for most environments and the performance changes on the DelayedHalfCheetah task are most noticeable. Then, we provide more ablation studies for remaining DelayedMujoco tasks in Appendix \ref{subsec:ablation}.

\vspace{1em}

\begin{figure*}[!h]
    \centering
	\subfigure[Control coefficient $c$]{\includegraphics[width=0.32\textwidth]{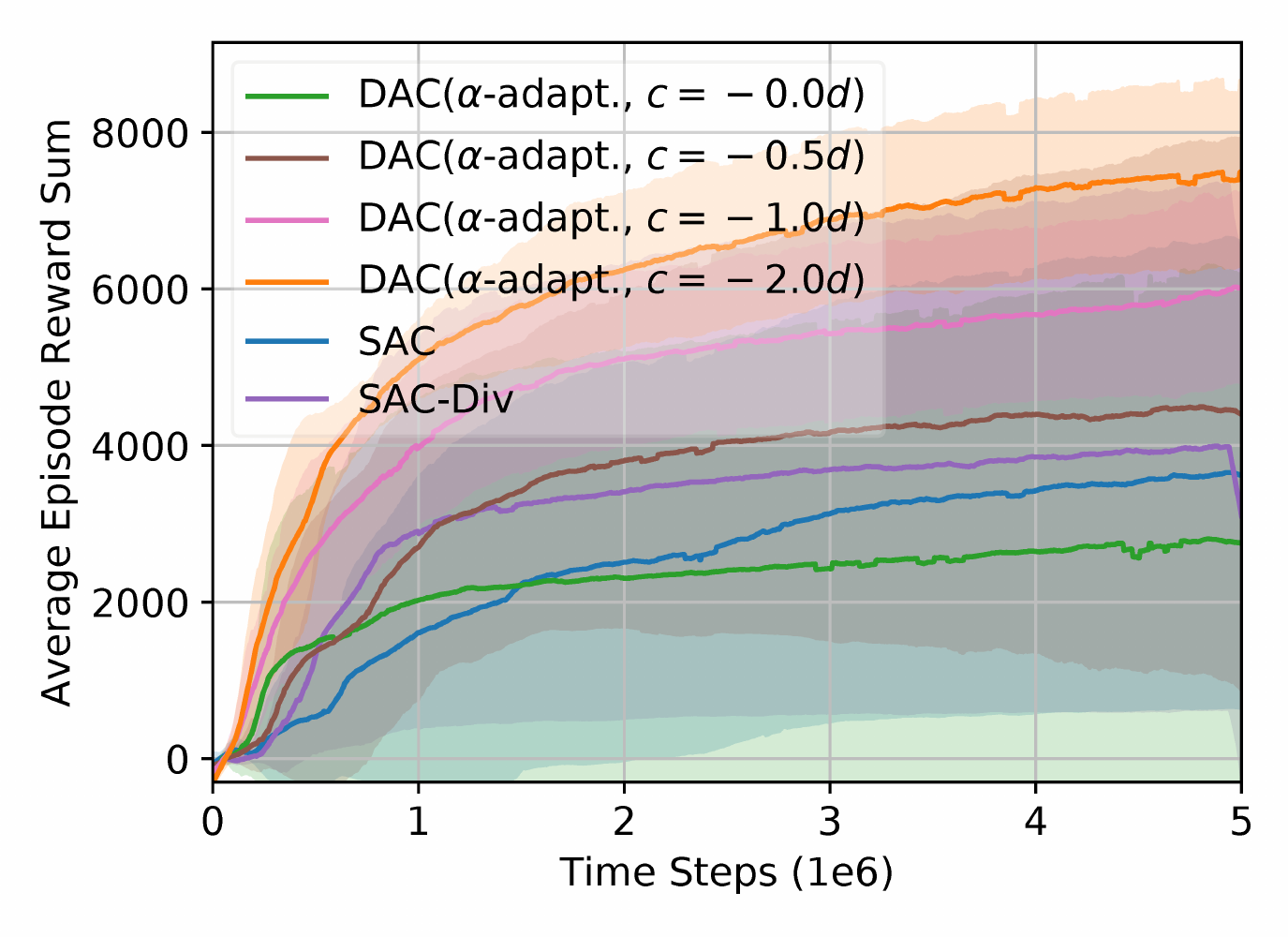}\label{fig:c}}
	\subfigure[Entropy coefficient $\beta$]{\includegraphics[width=0.32\textwidth]{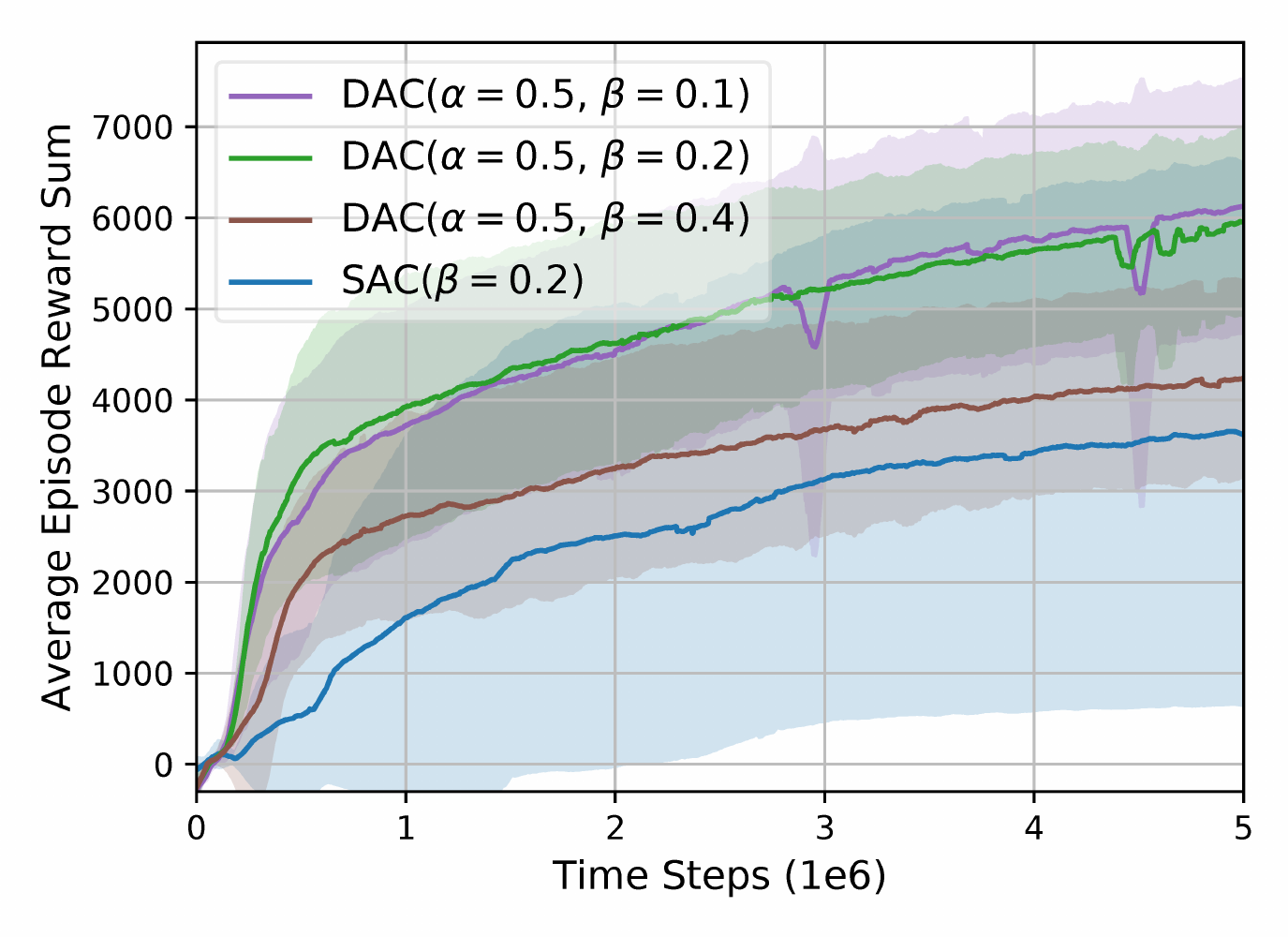}\label{fig:beta}}
	\subfigure[JS divergence]{\includegraphics[width=0.32\textwidth]{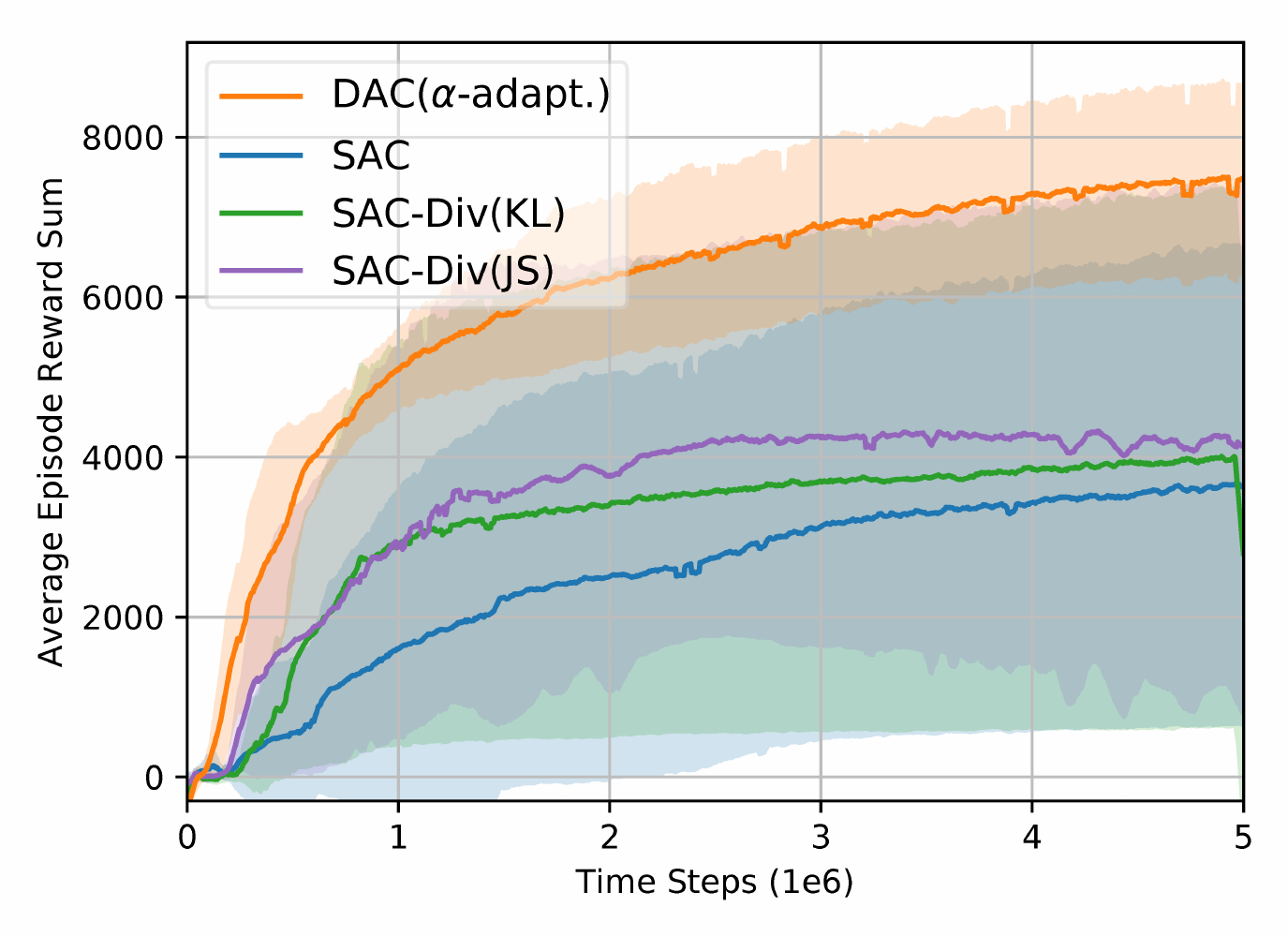}\label{fig:jsdiv}}
	\caption{Averaged learning curve for ablation study}
	\label{fig:ablation}
\end{figure*}

\vspace{1em}

{\bf Control coefficient $c$ in \eqref{eq:alphamaxmin}:} In the proposed  $\alpha$-adaptation
 \eqref{eq:alphamaxmin} in Section \ref{sec:autoalpha}, the control coefficient $c$ affects the learning behavior of $\alpha$.
Since $\Hc(\pi)$ and $D_{JS}^\alpha$ are proportional to the action dimension, we tried a few values such as $0$, $-0.5d$, $-1.0d$ and $-2.0d$, where $d=\mathrm{dim}(\mathcal{A})$.
Fig. \ref{fig:ablation}(a) shows the corresponding performance of DAC with $\alpha$-adaptation on DelayedHalfCheetah.
As seen in Fig. \ref{fig:ablation}(a), the performance depends on the change of $c$ as expected, and
$c=-2.0\cdot\mathrm{dim}(\mathcal{A})$ performs well. We observed that $-2.0d$ performed well for all considered tasks. Hence, we set $c= -2.0 d$ in \eqref{eq:lossalpha}.

\vspace{1em}

{\bf Entropy coefficient $\beta$ in \eqref{eq:objpi}:} As mentioned in \citep{haarnoja2018soft}, the  performance of SAC  depends on $\beta$. It is expected that the performance of  DAC depends on $\beta$ too.
Fig. \ref{fig:ablation}(b) shows the performance of DAC with fixed $\alpha=0.5$ for three different values of $\beta$: $\beta=0.1$, $0.2$ and $0.4$ on DelayedHalfCheetah. It is seen that the performance of DAC indeed depends on $\beta$. Although there exists performance difference for DAC depending on $\beta$, the performance of DAC is much better than SAC for a wide range of $\beta$.
One thing to note is that the coefficient of pure policy entropy regularization term for DAC is $\alpha\beta$, as seen in \eqref{eq:objpi}. Thus, DAC with $\alpha=0.5$ and $\beta=0.4$ has the same amount of pure policy entropy regularization as SAC with $\beta=0.2$. However, DAC with $\alpha=0.5$ and $\beta=0.4$ has much higher performance than SAC with $\beta=0.2$, as seen in Fig. Fig. \ref{fig:ablation}(b). So, we can see that the performance improvement of DAC comes from joint use of policy entropy $\mathcal{H}(\pi)$ and the sample action distribution from the replay buffer via $D_{JS}^\alpha(\pi||q)$.

\vspace{1em}

{\bf The effect of JS divergence:} In order to see the effect of the JS divergence on the performance, we  provide an additional ablation study in which we consider a single JS divergence for SAC-Div by using the ratio function in Section \ref{subsec:pisem}.  Fig. \ref{fig:ablation}(c) shows the performance comparison of SAC, SAC-Div(KL), SAC-Div(JS), and DAC. For SAC-Div(JS), we used $\delta_d=0.5$ for adaptive scaling in  \citep{hong2018diversity}. It is seen that there is no significant difference in performance between SAC-Div with JS divergence and SAC-Div with KL divergence. DAC still shows superiority to  both SAC-Div(KL) and SAC-Div(JS). This shows that DAC has more advantages than simply using JS divergence.

\newpage
\subsection{Ablation Studies for Remaining Tasks}
\label{subsec:ablation}

\begin{figure}[!h]
	\centering
	\subfigure[DelayedHopper-v1]{\includegraphics[width=0.3\textwidth]{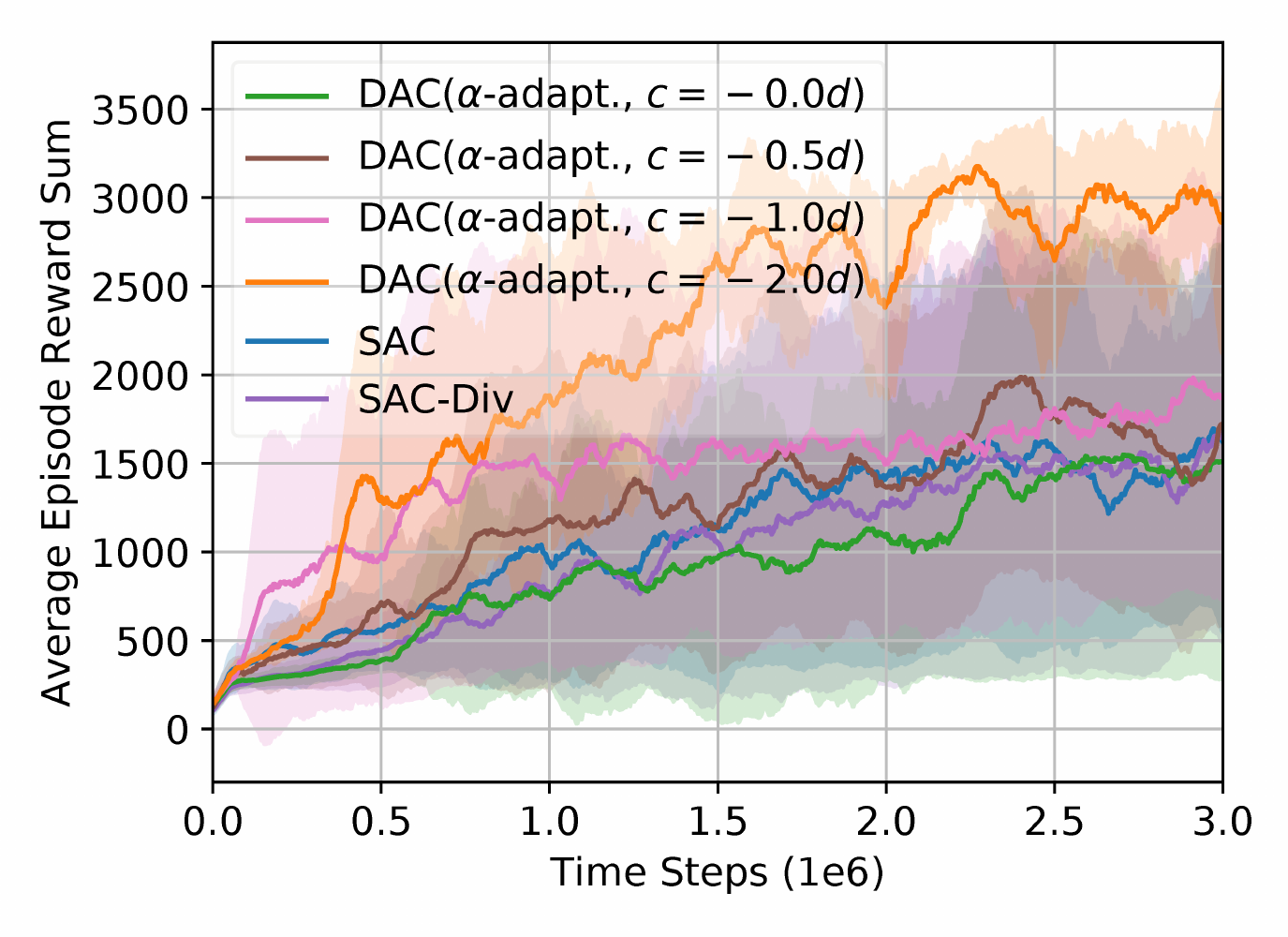}}
	\subfigure[DelayedWalker2d-v1]{\includegraphics[width=0.3\textwidth]{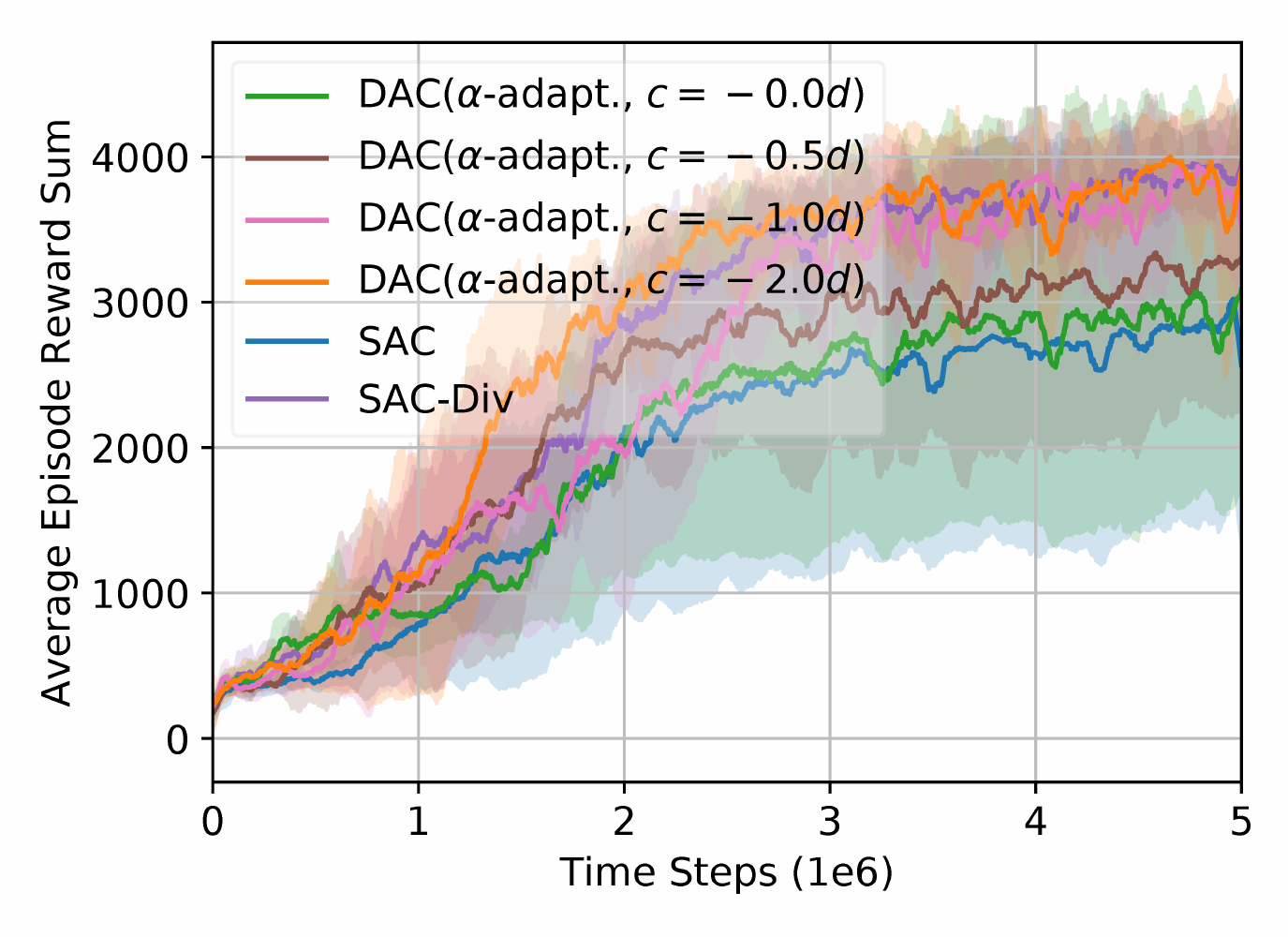}}
	\subfigure[DelayedAnt-v1]{\includegraphics[width=0.3\textwidth]{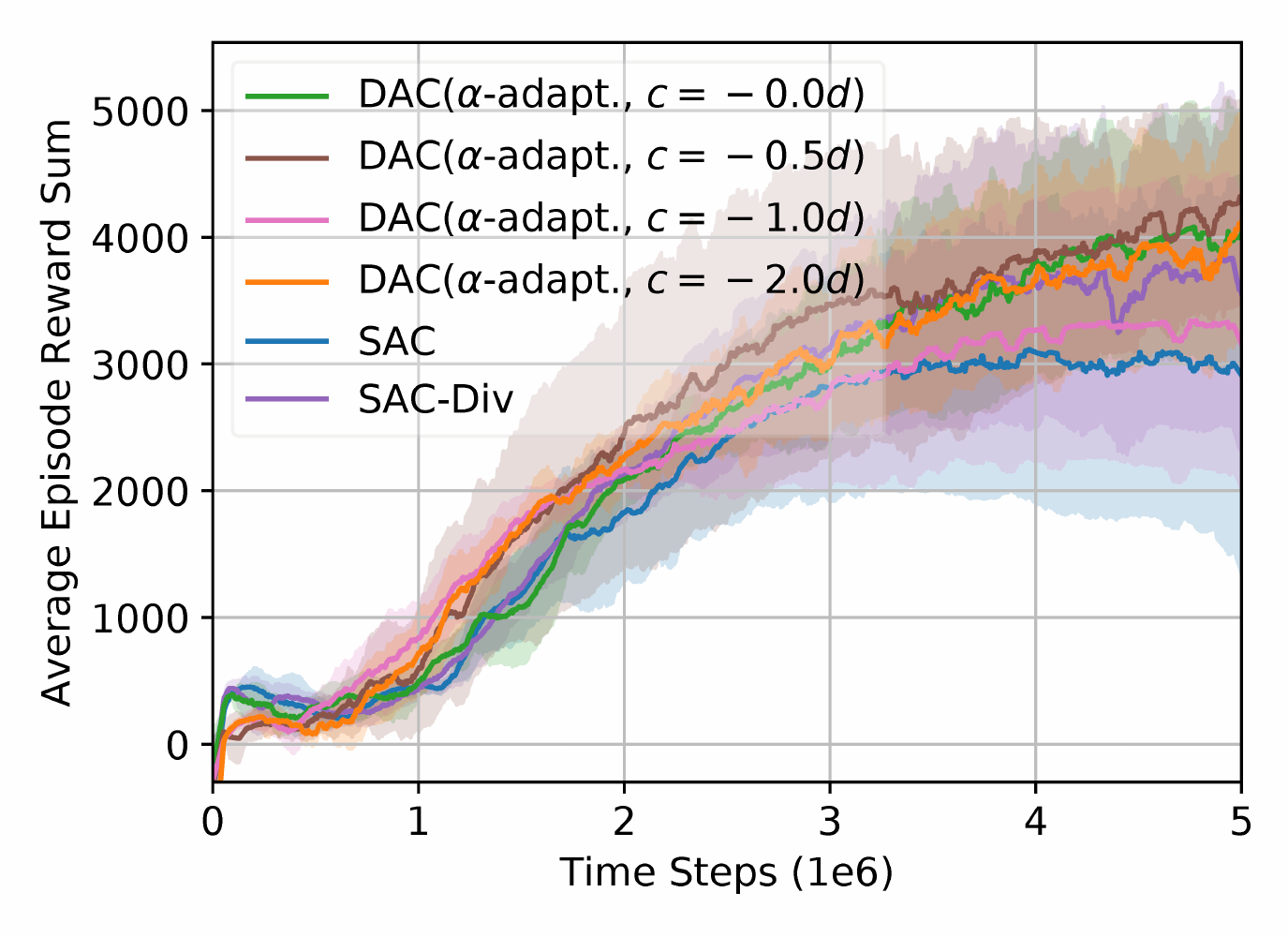}}
	\caption{Ablation study on $c$}
	\label{fig:ablation2}
\end{figure}

\begin{figure}[!h]
	\centering
	\subfigure[DelayedHopper-v1]{\includegraphics[width=0.3\textwidth]{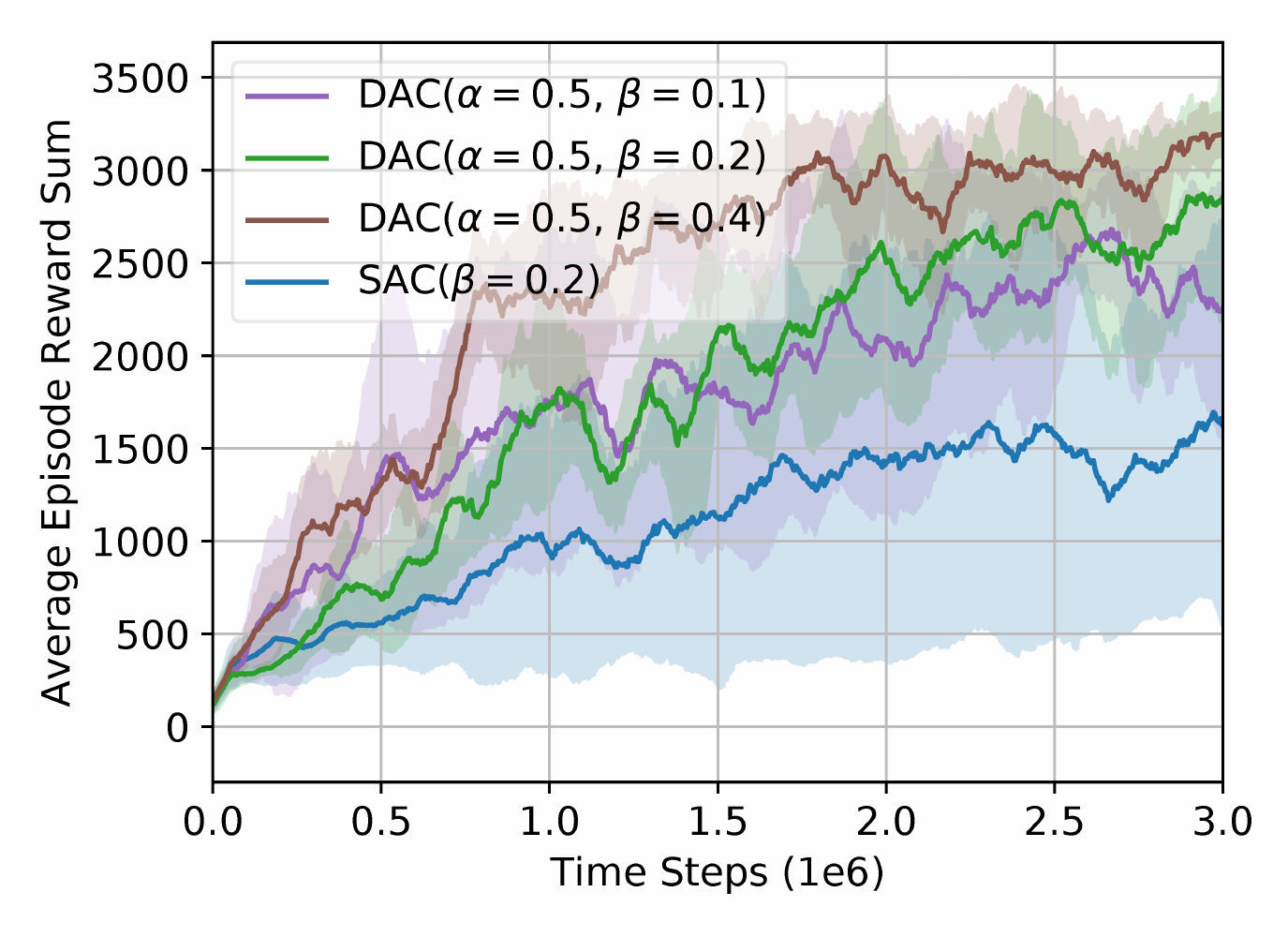}}
	\subfigure[DelayedWalker2d-v1]{\includegraphics[width=0.3\textwidth]{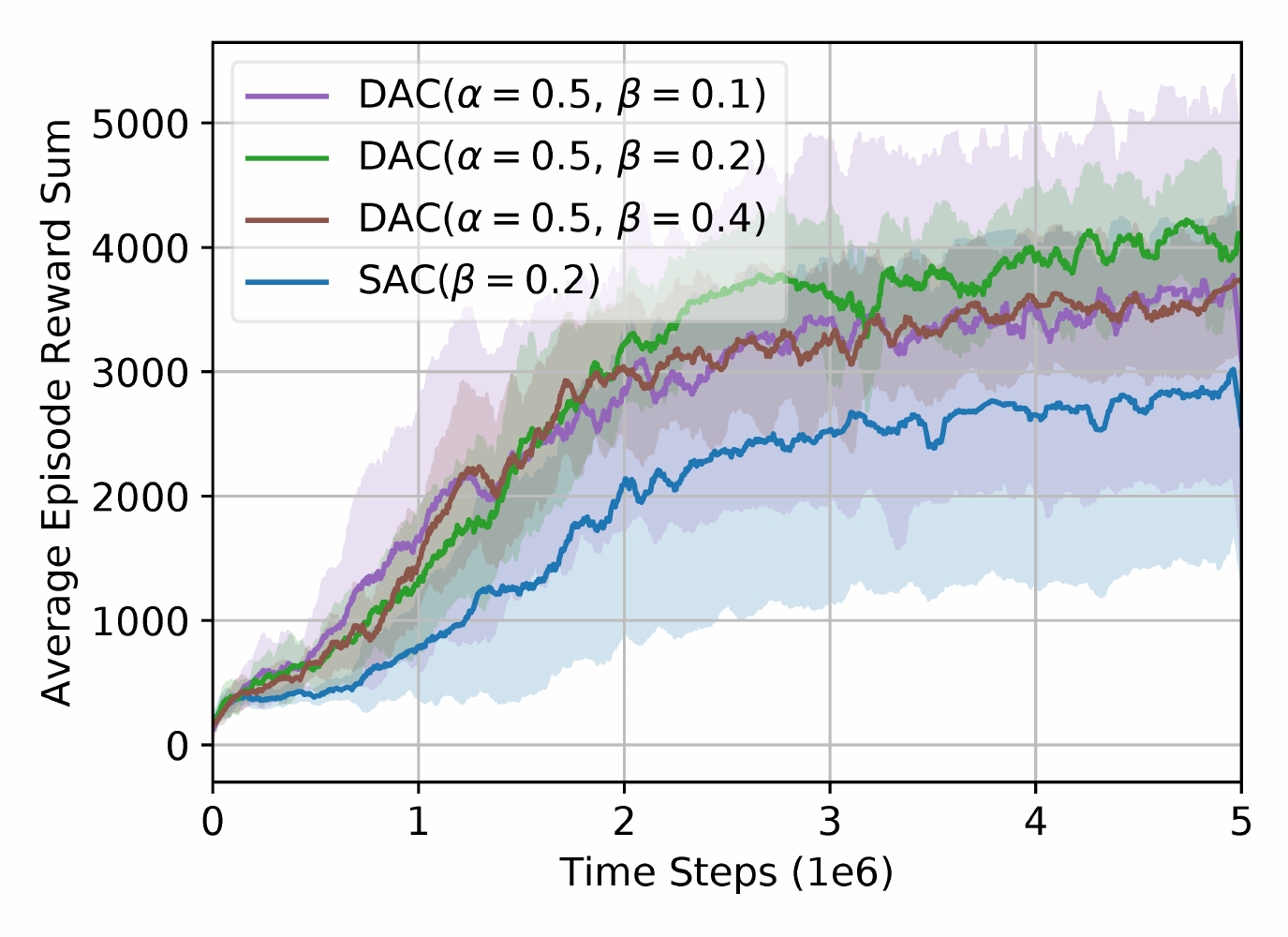}}
	\subfigure[DelayedAnt-v1]{\includegraphics[width=0.3\textwidth]{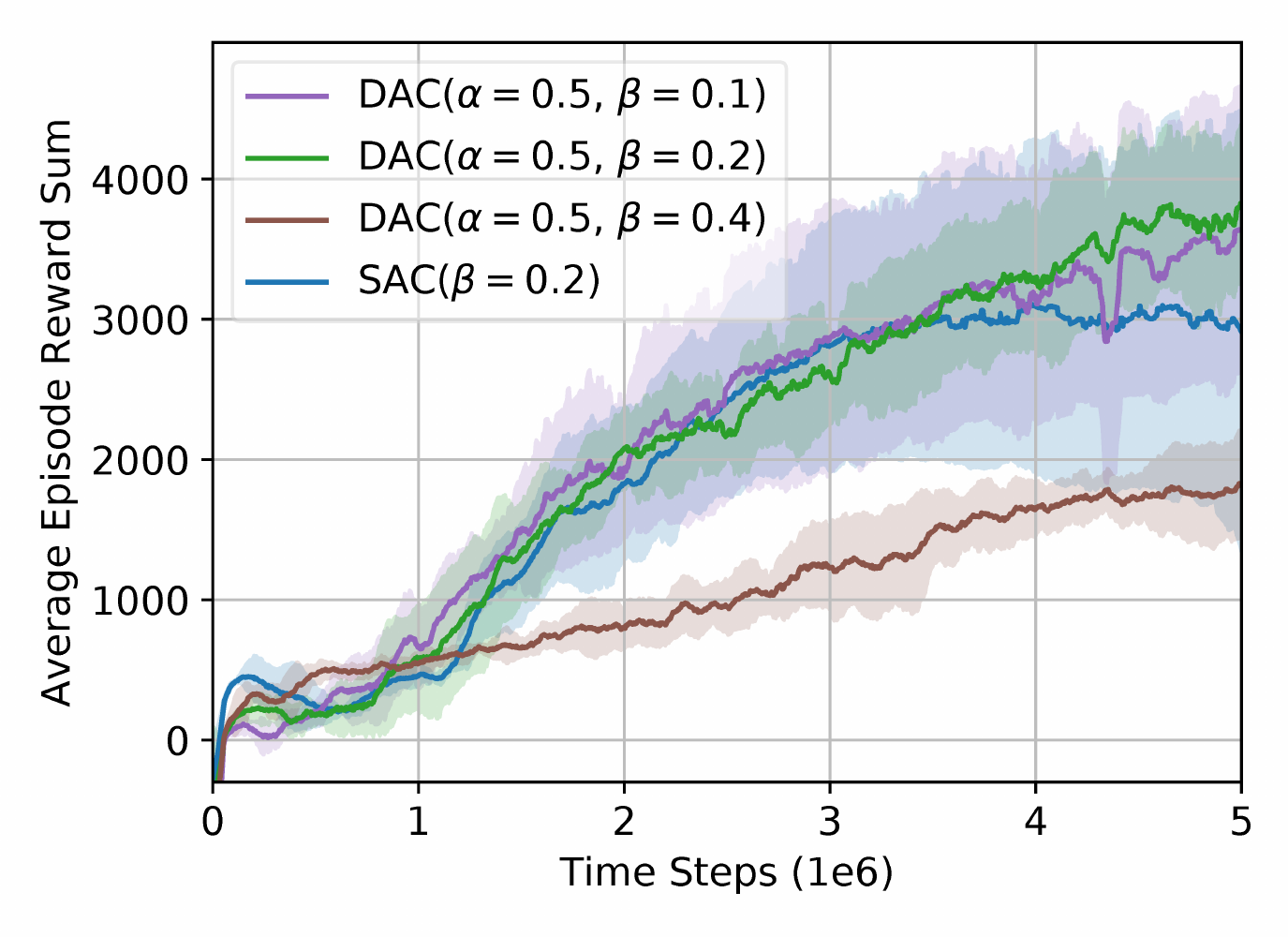}}
	\caption{Ablation study on $\beta$}
	\label{fig:ablation1}
\end{figure}

\begin{figure}[!h]
	\centering
	\subfigure[DelayedHopper-v1]{\includegraphics[width=0.3\textwidth]{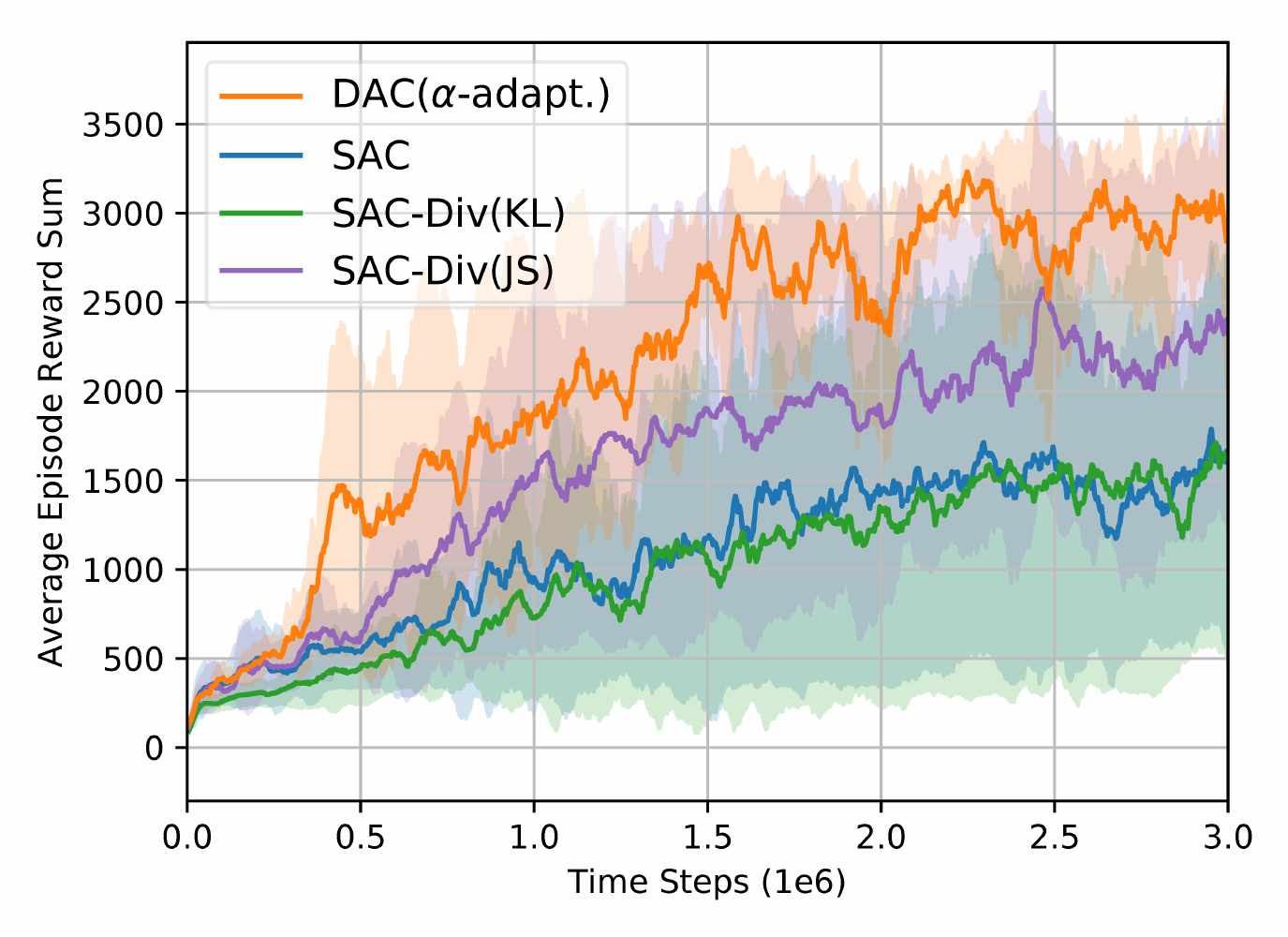}}
	\subfigure[DelayedWalker2d-v1]{\includegraphics[width=0.3\textwidth]{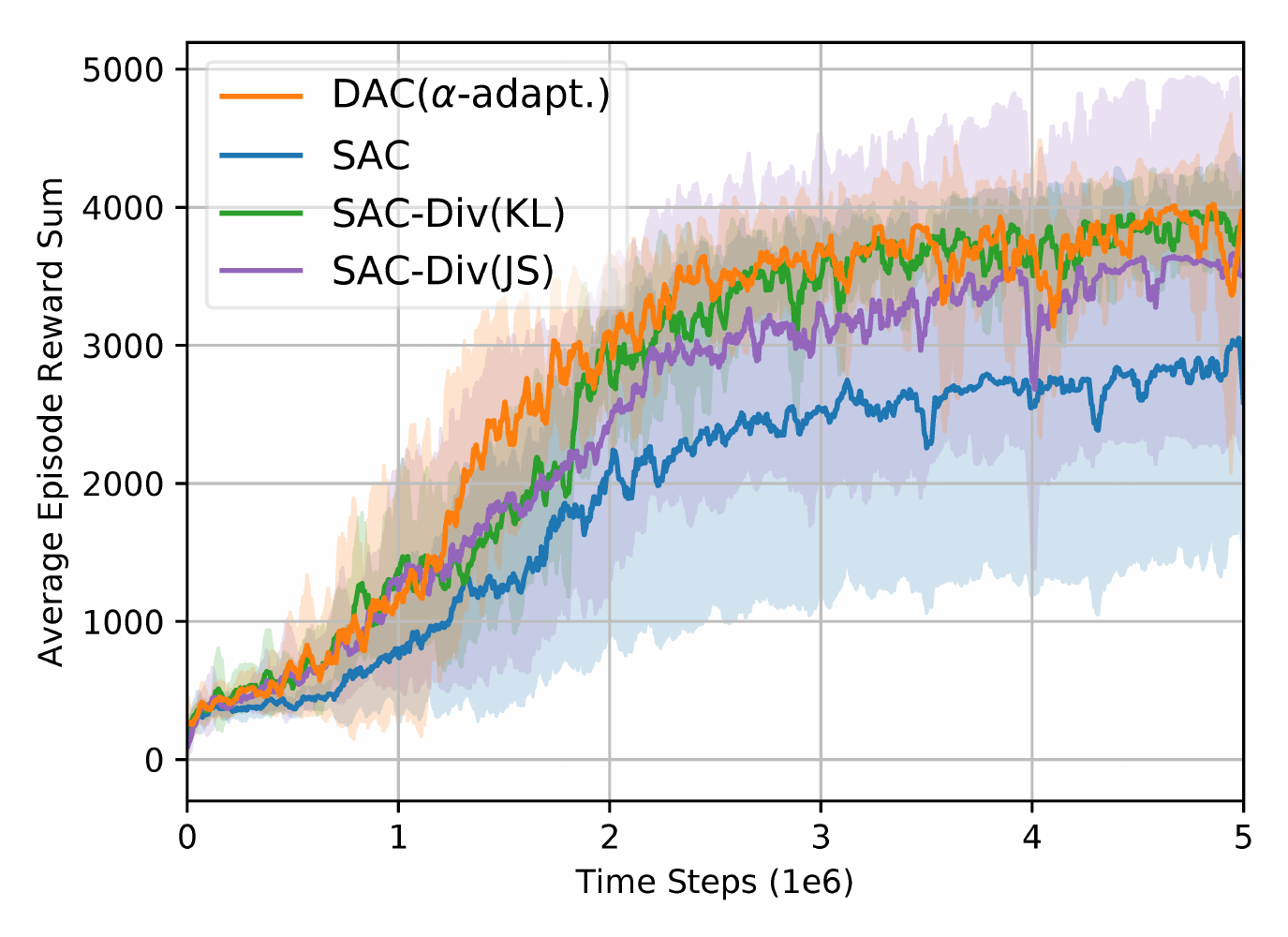}}
	\subfigure[DelayedAnt-v1]{\includegraphics[width=0.3\textwidth]{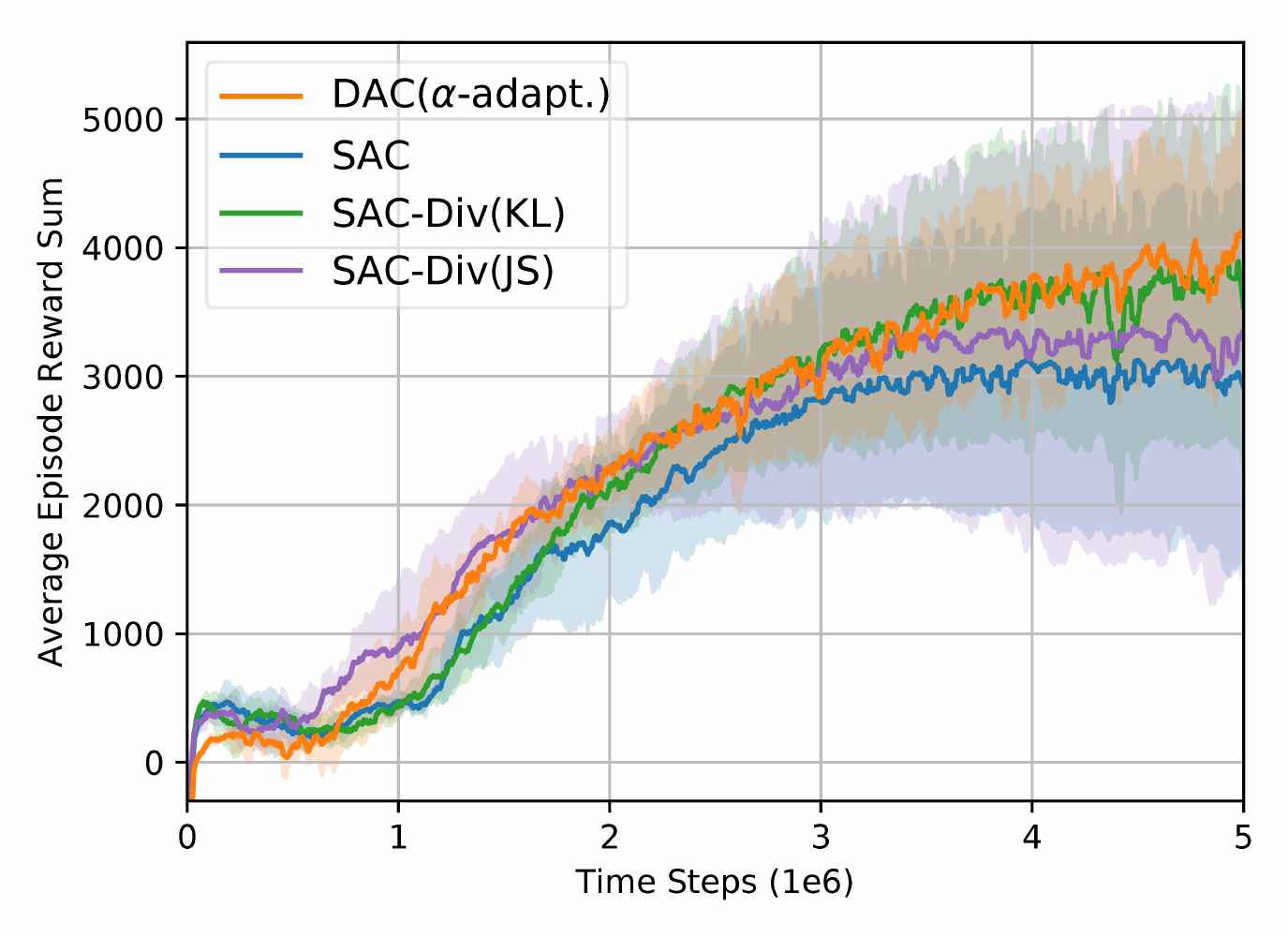}}
	\caption{Ablation study on SAC-Div with JS divergence}
	\label{fig:ablation3}
\end{figure}

Here, we provide more ablation studies for remaining delayed Mujoco tasks in Figure \ref{fig:ablation2}, Figure \ref{fig:ablation1}, and Figure \ref{fig:ablation3}. 

\vspace{1em}
{\bf Control coefficient $c$}

{\bf Entropy coefficient $\beta$}

{\bf Effect of JS divergence over SAC-Div}



\end{document}
